%% file: iclr2026_conference.tex
\PassOptionsToPackage{table}{xcolor}
\documentclass[11pt,a4paper,logo]{lumia}
\usepackage{listings}
\tcbuselibrary{listings,breakable,skins}
\usepackage[authoryear,round]{natbib}
\usepackage{etoolbox}
\usepackage{nicefrac}
\usepackage{subcaption}
\usepackage{booktabs}
\usepackage{makecell}
\usepackage{pifont}
\usepackage{multirow}
\usepackage{wrapfig}
\usepackage[table]{xcolor}

\usepackage{url}
\input{math_commands.tex}

\newtcblisting{promptbox}[1]{
  breakable, enhanced, listing only,
  listing options={basicstyle=\footnotesize\ttfamily, breaklines=true,
    breakindent=0pt, columns=fullflexible, keepspaces=true, showstringspaces=false,
    literate={–}{{\textendash}}1},
  colback=black!3, colframe=black!55, coltitle=white, fonttitle=\bfseries\small,
  title={#1}, boxrule=0.6pt, arc=2pt, left=5pt, right=5pt, top=2pt, bottom=2pt
}

\newtcolorbox{examplebox}[1]{
  breakable, enhanced, fontupper=\footnotesize,
  colback=blue!3, colframe=blue!40!black, coltitle=white, fonttitle=\bfseries\small,
  title={#1}, boxrule=0.6pt, arc=2pt, left=5pt, right=5pt, top=3pt, bottom=3pt
}
\usepackage{graphicx}
\usepackage{float}
\usepackage{placeins}
\usepackage{titletoc}

\newcommand{\cmark}{\textcolor{green!55!black}{\ding{51}}}
\newcommand{\pmark}{\textcolor{orange!90!black}{\ensuremath{\bigcirc}}}
\newcommand{\nmark}{\textcolor{red!75!black}{\ding{55}}}

\RequirePackage{xspace}
\makeatletter
\DeclareRobustCommand\onedot{\futurelet\@let@token\@onedot}
\def\@onedot{\ifx\@let@token.\else.\null\fi\xspace}
\def\eg{\emph{e.g}\onedot}

\makeatother

\graphicspath{{figures/}}
\hypersetup{colorlinks=true, linkcolor=blue!55!black, citecolor=teal!60!black, urlcolor=blue!55!black}

\definecolor{abstrabg}{HTML}{F2D1C5}
\definecolor{black50}{gray}{0.5}

\definecolor{citecolor}{RGB}{0,20,115}
\definecolor{tocnavy}{HTML}{17324D}
\definecolor{tocblue}{HTML}{2E6F95}
\definecolor{tocaccent}{HTML}{D56B55}
\definecolor{tocmuted}{HTML}{6B7785}
\definecolor{affilcolor}{HTML}{44566C}
\definecolor{tocline}{HTML}{D9E2E8}
\definecolor{appendixtocblue}{RGB}{0,20,115}

\renewcommand{\absfont}{\linespread{1.05}\fontsize{9.5}{11}\selectfont}

\fancypagestyle{firststyle}{%
  \fancyhf{}%
  \fancyhead[L]{%
  }%
  \fancyhead[R]{%
    \raisebox{2pt}{%
      \fontsize{9}{10.5}\selectfont\itshape
      Saturday 15\textsuperscript{th} August, 2026%
    }%
  }%
  \fancyfoot[C]{%
    {\sffamily\fontsize{7.4}{9}\selectfont\color{tocmuted}%
      Answer key \textrightarrow\ page 2%
    }%
  }%
  \renewcommand{\headrulewidth}{1pt}%
  \renewcommand{\footrulewidth}{1pt}%
}
\fancypagestyle{answerkeystyle}{%
  \fancyhf{}%
  \fancyfoot[C]{%
    \sffamily\fontsize{7.4}{9}\selectfont\color{tocmuted}%
    \textbf{Figure 1 answer key (from page 1; left/right):}\enspace
    I Generated/Real;\enspace II Real/Generated;\enspace
    III Real/Generated;\enspace IV Generated/Real.%
  }%
  \renewcommand{\headrulewidth}{0pt}%
  \renewcommand{\footrulewidth}{1pt}%
}

\hypersetup{
    colorlinks=true,
    linkcolor=appendixtocblue,
    citecolor=appendixtocblue,
    filecolor=appendixtocblue,
    urlcolor=appendixtocblue,
    linktoc=all,
}

\titlecontents{section}
  [15.54pt]
  {\addvspace{10.9pt}\fontsize{10}{11}\selectfont\bfseries\upshape\color{black}}
  {\contentslabel{1.5em}}
  {\hspace*{1.5em}}
  {\hfill\contentspage}
\titlecontents{subsection}
  [36.54pt]
  {\fontsize{10}{11}\selectfont\mdseries\upshape\color{black}}
  {\contentslabel{2.25em}}
  {}
  {\titlerule*[0.75em]{.}\contentspage}
\titlecontents{subsubsection}
  [65.51pt]
  {\fontsize{10}{11}\selectfont\mdseries\upshape\color{black}}
  {\contentslabel{3.15em}}
  {}
  {\titlerule*[0.75em]{.}\contentspage}

\makeatletter
\newcommand{\mainappendixgap}{\par\vspace{5.5pt}}
\newcommand{\papercontents}{%
  \begingroup
  \fontencoding{T1}\fontfamily{ptm}\selectfont
  \hypersetup{linkcolor=appendixtocblue}%
  \setlength{\parskip}{2.8pt}%
  \setcounter{tocdepth}{3}%
  \begin{adjustwidth}{0pt}{0pt}
    \vspace*{6.6pt}%
    \noindent
    {\fontsize{20.74}{24.8}\selectfont\bfseries\upshape\color{black}Contents}\par
    \vspace{4.5pt}%
    \noindent\rule{\linewidth}{0.4pt}\par
    \vspace{-6.8pt}%

    \ttl@change@i{\z@}{section}{37.5pt}
      {\addvspace{6pt}\fontsize{9}{9}\selectfont\bfseries\upshape\color{appendixtocblue}}
      {\contentslabel{1.5em}}
      {}
      {\hfill\contentspage\hspace*{23.9pt}\endgraf}%
    \ttl@change@v{section}{}{}{}%
    \ttl@change@i{\@ne}{subsection}{58.2pt}
      {\fontsize{9}{9}\selectfont\mdseries\upshape\color{appendixtocblue}}
      {\contentslabel{2.3em}}
      {}
      {\titlerule*[0.75em]{\textcolor{black}{.}}\contentspage\hspace*{23.9pt}}%
    \ttl@change@v{subsection}{}{}{}%
    \ttl@change@i{\tw@}{subsubsection}{83.1pt}
      {\fontsize{9}{9}\selectfont\mdseries\upshape\color{appendixtocblue}}
      {\contentslabel{3.15em}}
      {}
      {\titlerule*[0.75em]{\textcolor{black}{.}}\contentspage\hspace*{23.9pt}}%
    \ttl@change@v{subsubsection}{}{}{}%
    \@starttoc{toc}%
    \vspace{-3.2pt}%
    \noindent\rule{\linewidth}{0.4pt}\par
  \end{adjustwidth}
  \endgroup
}

\newcommand{\appendixcontents}{%
  \begingroup
  \fontencoding{T1}\fontfamily{ptm}\selectfont
  \hypersetup{linkcolor=appendixtocblue}%
  \setlength{\parskip}{3pt}%
  \setcounter{tocdepth}{2}%
  \begin{adjustwidth}{0pt}{0pt}
    \vspace*{6.6pt}%
    \noindent
    {\fontsize{20.74}{24.8}\selectfont\bfseries\upshape\color{black}Appendix}\par
    \vspace{19.9pt}%
    \noindent
    {\fontsize{14.4}{17.3}\selectfont\bfseries\upshape\color{black}Table of Contents}\par
    \vspace{-13.2pt}%
    \noindent\rule{\linewidth}{0.4pt}\par
    \vspace{-6.8pt}%

    \ttl@change@i{\z@}{section}{37.5pt}
      {\addvspace{8.8pt}\fontsize{9}{9}\selectfont\bfseries\upshape\color{appendixtocblue}}
      {\contentslabel{1.5em}}
      {}
      {\hfill\contentspage\hspace*{23.9pt}}%
    \ttl@change@v{section}{}{}{}%
    \ttl@change@i{\@ne}{subsection}{58.2pt}
      {\fontsize{9}{9}\selectfont\mdseries\upshape\color{appendixtocblue}}
      {\contentslabel{2.3em}}
      {}
      {\titlerule*[0.75em]{\textcolor{black}{.}}\contentspage\hspace*{23.9pt}}%
    \ttl@change@v{subsection}{}{}{}%
    \@starttoc{atoc}%
    \vspace{-3.2pt}%
    \noindent\rule{\linewidth}{0.4pt}\par
  \end{adjustwidth}
  \endgroup
}
\makeatother

\newcommand{\appendixsection}[1]{%
  \section{#1}%
  \addcontentsline{atoc}{section}{\protect\numberline{\thesection}#1}%
}
\newcommand{\appendixsubsection}[1]{%
  \subsection{#1}%
  \addcontentsline{atoc}{subsection}{\protect\numberline{\thesubsection}#1}%
}

\providecommand{\BusterXpp}{BusterX+\!+}

\title{Can We Defend Against AI-Generated Video\\[-1pt]
Attacks on Real-World Crisis Events?\\[3pt]
A Systematic Evaluation of Detectors, Generators\\[-1pt]
and Social Dissemination}

\setheadertitle{Can We Defend Against AI-Generated Video Attacks on Real-World Crisis Events?}

\newcommand{\projectwebsiteurl}{https://ra-bench-crisis-video.yxgma811120.chatgpt.site/}
\newcommand{\projectdataseturl}{https://huggingface.co/datasets/liangshuo0111/RA-Bench}
\newcommand{\projectgithuburl}{https://github.com/24029100313/RA-Bench}
\definecolor{resourceblue}{HTML}{3934A5}
\definecolor{resourcegold}{HTML}{B8862F}
\newcommand{\coverresourcelink}[4]{%
  #1\kern0.35em%
  \textcolor{black}{\textbf{#2:}}\kern0.28em%
  \href{#4}{\textcolor{resourceblue}{\textbf{#3}}}%
}
\newcommand{\coverresources}{%
  \coverresourcelink{\textcolor{tocblue}{\faGlobe}}{Website}{RA-Bench}{\projectwebsiteurl}%
  \hspace{1.45em}%
  \coverresourcelink{\raisebox{-0.18em}{\includegraphics[height=1.06em]{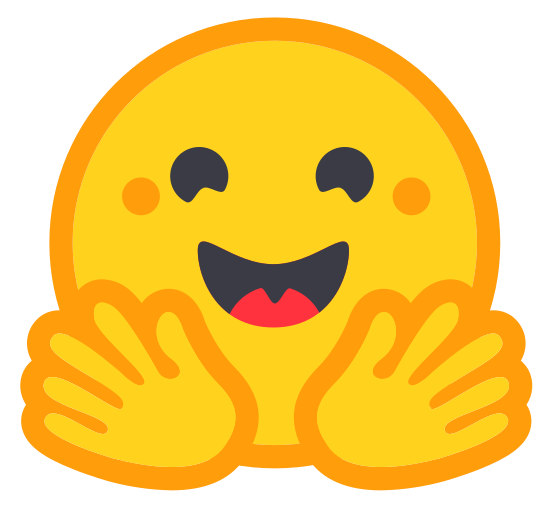}}}{Dataset}{Hugging Face}{\projectdataseturl}%
  \hspace{1.45em}%
  \coverresourcelink{\textcolor{black}{\faGithub}}{Code}{GitHub}{\projectgithuburl}%
}

\newcommand{\authorsep}{\textcolor{tocline}{\enspace\textbullet\enspace}}
\newcommand{\seniorauthorsep}{\textcolor{tocline}{\kern0.35em\textbullet\kern0.35em}}
\newcommand{\affmark}[1]{\textsuperscript{\textcolor{tocmuted}{#1}}}
\newcommand{\cofirst}[2]{%
  #1%
  \textsuperscript{\textcolor{tocaccent}{*}\textcolor{tocmuted}{,#2}}%
}

\newcommand{\cofirstlead}[2]{%
  #1%
  \textsuperscript{%
    \textcolor{tocaccent}{*}\textcolor{tocmuted}{,}%
    \textcolor{tocaccent}{\S}\textcolor{tocmuted}{,#2}%
  }%
}

\newcommand{\corrauthor}[2]{%
  #1%
  \textsuperscript{%
    \textcolor{tocaccent}{\ensuremath{\dagger}}\textcolor{tocmuted}{,#2}%
  }%
}
\newcommand{\coverauthorblock}{%
  \begin{minipage}{0.995\textwidth}
  \centering
  {\normalfont\fontsize{7.8}{9.25}\selectfont
    \cofirst{Shuo Liang}{1,2}\authorsep
    \cofirst{Yixing Ma}{1,3}\authorsep
    \cofirstlead{Pengfei Zhou}{1}\par
  }
  \vspace{0.5pt}
  {\normalfont\fontsize{7.8}{9.25}\selectfont
    Zhenglin Wan\affmark{1}\authorsep
    Xingyan Chen\affmark{3}\authorsep Zihan Mei\affmark{5}\authorsep
    Manting Li\affmark{3}\authorsep Feihan Chen\affmark{6}\authorsep
    Zhiwen Wang\affmark{7}\authorsep Bin Xu\affmark{8}\\
    Haotian Zhang\affmark{6}\authorsep Jiajun Song\affmark{6}\authorsep
    Shiya Su\affmark{9}\authorsep Run Liu\affmark{10}\authorsep
    Zhenghang Ni\affmark{2}\authorsep Yifa Yu\affmark{11}\\
    Jintao Hong\affmark{2}\authorsep Bolong Feng\affmark{10}\authorsep
    Yifei Liu\affmark{3}\authorsep Zirui Zhang\affmark{12}\authorsep
    Jingxuan Zhang\affmark{1}\authorsep Songlin Zhao\affmark{14}\\
    Yifan Bai\affmark{13}\authorsep Kang Tan\affmark{15}\authorsep
    Yizhe Liu\affmark{13}\authorsep Junhao Du\affmark{13}\authorsep
    Yongtao Ge\affmark{19}\authorsep Zhaopan Xv\affmark{16}\authorsep
    Xinyuan Zhang\affmark{16}\\[-0.2pt]
    {\fontsize{7.8}{9.25}\selectfont
    Mengru Ma\affmark{2}\seniorauthorsep
    Chunhua Shen\affmark{17}\seniorauthorsep
    \corrauthor{Wei Wang}{4}\seniorauthorsep
    \corrauthor{Yang You}{1}\seniorauthorsep
    \corrauthor{Zheng Zhu}{18}\seniorauthorsep
    \corrauthor{Kaipeng Zhang}{19}\seniorauthorsep
    \corrauthor{Wangbo Zhao}{4}\par
    }
  }
  \vspace{3pt}
  {\fontsize{6.25}{7.65}\selectfont\color{tocmuted}
    \textsuperscript{\textcolor{tocaccent}{*}}Equal contribution
    \qquad
    \textsuperscript{\textcolor{tocaccent}{\S}}\textbf{Project lead:}
    Pengfei Zhou (\href{mailto:zpf4wp@outlook.com}{zpf4wp@outlook.com})\\[-0.3pt]
    \textsuperscript{\textcolor{tocaccent}{\ensuremath{\dagger}}}%
    \textbf{Corresponding authors:} Wangbo Zhao
    (\href{mailto:wangbo.zhao96@gmail.com}{wangbo.zhao96@gmail.com})\quad
    Kaipeng Zhang
    (\href{mailto:kaipeng.zhang@shanda.com}{kaipeng.zhang@shanda.com})
    \quad
    Zheng Zhu (\href{mailto:zhengzhu@ieee.org}{zhengzhu@ieee.org})
    \\[-5.4pt]
    Yang You (\href{mailto:yangyou@nus.edu.sg}{yangyou@nus.edu.sg})\quad
    Wei Wang (\href{mailto:weiwa@cse.ust.hk}{weiwa@cse.ust.hk})
  }\par
  \vspace{3pt}
  {\normalfont\fontsize{5.65}{6.9}\selectfont\color{affilcolor}
    \setlength{\tabcolsep}{0pt}%
    \renewcommand{\arraystretch}{0.82}%
    \begin{tabular}{@{}c@{}}
      \affmark{1}National University of Singapore\authorsep
      \affmark{2}Xidian University\authorsep
      \affmark{3}University of California, Berkeley\authorsep
      \affmark{4}The Hong Kong University of Science and Technology\\
      \affmark{8}InfRec, Cardinal AI Lab\authorsep
      \affmark{7}Monash University\authorsep
      \affmark{6}Renmin University of China\authorsep
      \affmark{5}Arizona State University\authorsep
      \affmark{14}University of Illinois Urbana--Champaign\\
      \affmark{11}Stanford University\authorsep
      \affmark{17}Zhejiang University\authorsep
      \affmark{15}Carnegie Mellon University\authorsep
      \affmark{12}ETH Zurich\authorsep
      \affmark{10}Shanghai Jiao Tong University\\
      \affmark{13}Xi'an Jiaotong University\authorsep
      \affmark{18}GigaAI\authorsep
      \affmark{9}University of Wisconsin--Madison\authorsep
      \affmark{19}Alaya Lab\authorsep
      \affmark{16}Independent Researcher
    \end{tabular}
  }\par
  \end{minipage}
}
\newcommand{\institutionlogos}{%
  \noindent\makebox[\textwidth][c]{%
    \includegraphics[width=0.97\textwidth]{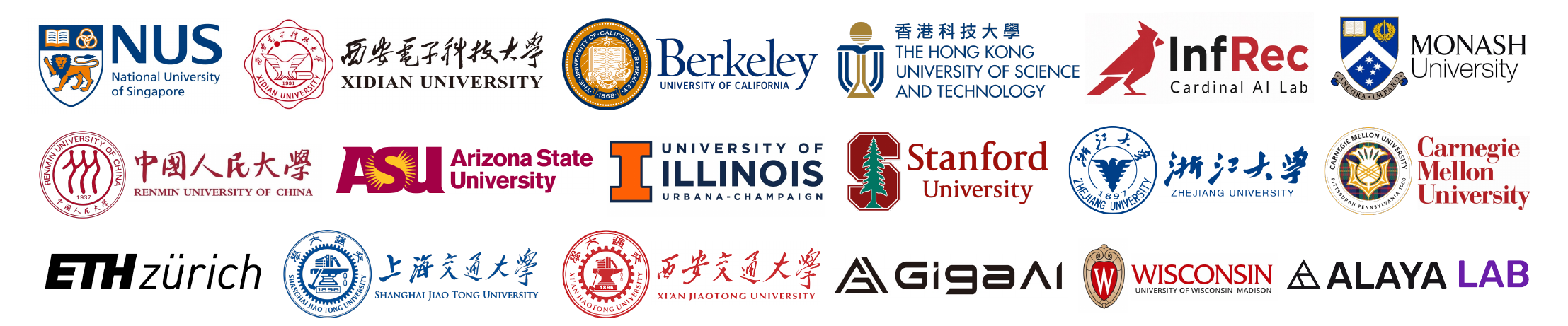}%
  }%
}
\makeatletter
\gdef\@author{\coverauthorblock}
\makeatother

\correspondingemail{}

\teaserfigure{%
    \begin{minipage}{\textwidth}
        \centering
        \includegraphics[width=0.985\linewidth]{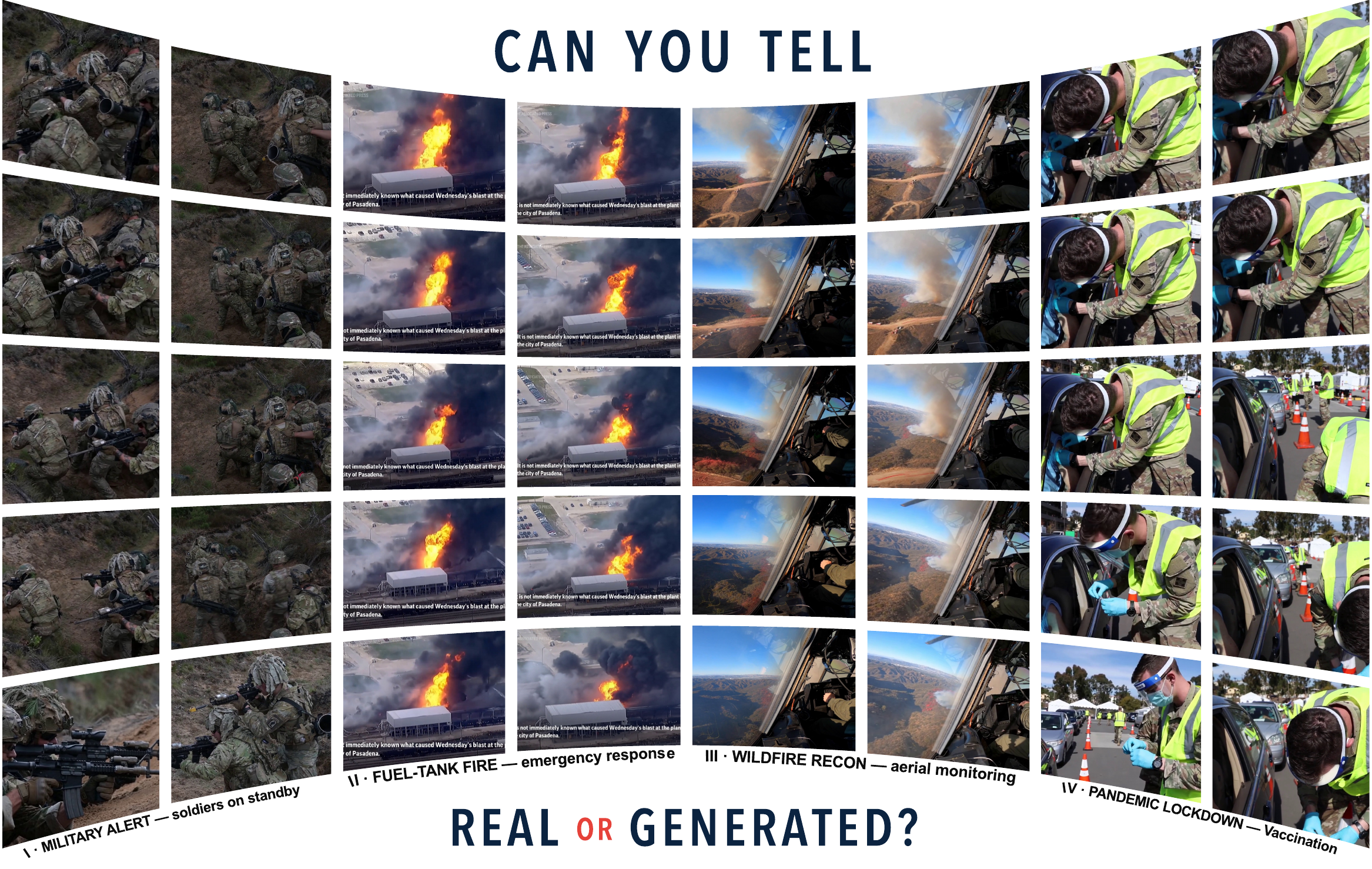}%
        \captionsetup{font=small,justification=centering,singlelinecheck=true}
        \captionof{figure}{\textbf{Which is which?} Real and AI-generated clips from four crisis scenarios are paired side by side; identify the synthetic example in each pair before consulting the answer key on the following page.}
        \label{fig:which-is-which}
    \end{minipage}
}

\makeatletter
\renewcommand{\maketitle}{%
  \thispagestyle{firststyle}
  \bgroup
  \setlength{\parindent}{0pt}%
  \begin{adjustwidth}{0pt}{0pt}
    \begin{center}
      {\fontsize{19.8}{21.5}\selectfont\bfseries \@title\par}%
      \vskip2.8pt
      {\sffamily\fontsize{8}{9}\selectfont\coverresources\par}%
      \vskip3.4pt
      {\@author\par}%
    \end{center}
  \end{adjustwidth}
  \egroup
  \vspace{0pt plus 0.10fill}
  {\institutionlogos\par}%
  \vspace{0pt plus 0.50fill}
  {\lumia@teasercontent}%
  \vspace{0pt plus 0.04fill}
}
\makeatother

\begin{document}


\maketitle

\clearpage
\thispagestyle{answerkeystyle}
\papercontents
\clearpage

\begin{abstract}

Recent video generators can fabricate realistic depictions of wars, disasters, public emergencies, and other real-world crises, creating substantial risks of misinformation. 
Existing benchmarks, however, provide limited evidence on detector and generator behavior in such settings, including how detectability varies with generation conditions, how people perceive generated videos, and whether detectors remain reliable during social dissemination.
To address this gap, we introduce \textbf{RA-Bench}, a benchmark for AI-generated video detection that uses \textbf{R}eal videos as \textbf{A}nchors. RA-Bench contains 17{,}886 videos, comprising 1{,}830 real-video anchors across 10 social-risk categories and 16{,}056 generated clips from four open-source and five closed-source generators. 
Based on RA-Bench, we organize our evaluation along three dimensions. We first assess detector generalization across seven traditional detectors, ten zero-shot multimodal models under three review settings, and two MLLMs specifically fine-tuned on AI-generated video detection. Across these methods, none of the three detector families generalizes consistently across RA-Bench instances. We then examine how detectability varies with generation quality, conditioning information, and sampling seeds. These analyses show that generation properties affect detector families differently, while source-level detection patterns remain stable across seeds. Finally, we study human authenticity judgments and detector reliability during social dissemination. We find that videos that mislead people are also difficult for current detectors, and that social dissemination makes detection harder. Together, these findings show that current methods struggle to detect realistic AI-generated videos, highlighting the need for detectors robust to evolving video generators.

\end{abstract}
\abscontent
\section{Introduction}
\label{sec:intro}
Recent video generation models~\citep{openai2024sora,google2024veo2,seedance2026seedance} can produce increasingly realistic clips with coherent appearance and motion. By reducing the time and effort required to create high-quality videos, these models can substantially improve video-production efficiency. However, these advances can also create new societal risks. For instance, generated videos depicting real-world crises, such as wars, disasters, and public-health emergencies, may deceive viewers and provoke public panic. The examples in Figure~\ref{fig:which-is-which} highlight the challenge of distinguishing generated content from real videos.


In response to these risks, numerous benchmarks~\citep{ma2024decof,chen2024demamba,he2024exposing} have been developed to evaluate the ability of existing detectors~\citep{wang2020cnngenerated,ojha2023universal,ma2024decof,chen2024demamba,interno2025restrav} to identify realistic AI-generated videos. Recent benchmarks have broadened generator coverage, increased dataset scale, and introduced more comprehensive evaluation
protocols~\citep{ma2024decof,chen2024demamba,ni2026genvidbench,ma2026aigvdbench}. Meanwhile, other studies have extended the task beyond binary classification by examining forensic explanations, world-simulation settings, and source backtracking~\citep{wen2025busterx,chen2025genworld,liao2026chameleon}. 

However, as shown in Table~\ref{tab:benchmark_comparison}, existing studies primarily focus on general video content, with limited systematic analysis of detector performance and generator behavior in the context of real-world crises and other important socially consequential events. Therefore, \emph{it remains unclear whether current detectors are sufficiently reliable to protect society from the threats posed by increasingly realistic video generation models and potential misuses.}



To tackle this problem, we introduce \textbf{RA-Bench}, a benchmark for AI-generated video detection that uses \textbf{R}eal videos as \textbf{A}nchors. RA-Bench comprises 1{,}830 real-video anchors drawn from 675 publicly available source videos and spans 10 social-risk categories and 44 subcategories. To simulate a plausible misinformation scenario in which real image is used to generate subsequent events, we condition image-to-video (I2V) generators on the first frame of each anchor. Given the same first frame and text prompt, four open-source and five closed-source generators produce 16{,}056 generated clips, resulting in a total of 17{,}886 videos.


Based on RA-Bench, we perform a systematic evaluation along three dimensions:
\emph{1. Detector Generalization.} We evaluate three detector families: seven traditional detectors, ten zero-shot multimodal model under three review settings, and two MLLMs specifically fine-tuned for AI-generated video detection.
\emph{2. Generation Properties.} We investigate how video detectability varies with generation quality, conditioning information, and sampling seeds. Specifically, we examine whether higher perceptual quality makes generated videos harder to detect, how the amount of real-image conditioning affects detector behavior, and whether source-level detection patterns remain stable across generation seeds.
\emph{3. Human Behavior.} We examine perceptual judgments of authenticity and detector behavior during social dissemination. We conduct a human study to identify generated videos that viewers find difficult to distinguish from real videos and specifically evaluate detectors on these human-deceptive cases, which constitute \textbf{RA-Bench-HumanProof}. We further introduce \textbf{RA-Bench-LastMile}, a social dissemination simulation for evaluating detector robustness.


Our experiments reveal three main findings:
\begin{itemize}
    \item \textbf{None of the three detector families generalizes consistently across RA-Bench sources.}
    Traditional detectors fall from public-reference AUCs of 67.6--98.6\% to source-level means of 43.9--57.3\% on RA-Bench, and their rankings change across generators. Scaling zero-shot multimodal models does not remove their sensitivity to prompts and generation sources. Replacing timestamps with frame indices reduces Skyra to 54.4--54.9\% mean BAcc, while 
    %
    %
    \BusterXpp\ 
    achieves only 4.1--9.1\% FakeR (Section~\ref{sec:detector-side}).

    \item \textbf{Generation properties affect detector families differently, with limited variation across seeds.}
    A 50-point Condition Fidelity increase is associated with a 14.4-point decrease in Gemini Diagnostic FakeR, whereas dynamic content increases Gemini's evidence for the generated class but leaves the traditional-detector mean nearly unchanged. Across T2V, first-frame I2V, and first+last-frame I2V, the seven-detector mean AUC changes from 33.4\% to 50.9\% and 44.6\%, whereas the average FakeR across fine-tuned MLLMs decreases from 70.5\% to 42.7\% and 28.3\%. (Section~\ref{sec:generation-side}).

    \item \textbf{Videos that mislead people are also difficult for current detectors, and social dissemination weakens detection further.}
    Reviewers identify 68.6\% of open-source videos as generated, but only 52.9\% of closed-source videos, with Seedance2.0 and Kling falling to 40.7\% and 45.1\%. We use 633 generated videos that are labeled \emph{Real} by all five reviewers to form \textbf{RA-Bench-HumanProof}. On this subset, Gemini Binary and Diagnostic reach only 54.7\% and 54.5\% BAcc, while the seven traditional detectors average 47.5\% AUC. Separately, the Full condition of the social dissemination simulation reduces mean FakeR across the five fine-tuned configurations from 46.0\% to 1.4\% (Section~\ref{sec:human-reliability}).
\end{itemize}

In conclusion, our findings demonstrate that current methods still cannot reliably detect realistic AI-generated videos. This limitation is especially consequential when generated content depicts real-world crises, highlighting the need for detection systems that remain effective as generation models evolve and videos circulate in the real world.
\section{Related Work}

\paragraph{Benchmarks for AI-Generated Video Detection.}
As video generation models become more realistic, benchmarking whether existing detectors can accurately identify AI-generated videos has become increasingly important. Early efforts, such as GVF~\citep{ma2024decof} and GenVideo~\citep{chen2024demamba}, pair real videos with generated counterparts and evaluate cross-generator transfer. Subsequent benchmarks further expand the scale, generator coverage, and evaluation protocols, including GenVidBench~\citep{ni2026genvidbench} and AIGVDBench~\citep{ma2026aigvdbench}. Recent benchmarks further extend evaluation beyond binary classification to MLLM-based explanation~\citep{wen2025busterx}, world-simulation settings~\citep{chen2025genworld}, and source backtracking~\citep{liao2026chameleon}. Despite this progress, existing benchmarks remain limited in their coverage of misuse scenarios involving real-world crises and other socially consequential events.
To address this gap, we introduce RA-Bench and use it to systematically assess the robustness and limitations of existing detectors in these settings.

\paragraph{Realistic video generation models.}
In recent years, video generation models have rapidly evolved from an emerging research topic into a technology with significant real-world and societal impact. Improving video realism has long been a central objective in this field, from early approaches extended from image generation models, such as Stable Video Diffusion~\citep{blattmann2023stable} and VideoCrafter~\citep{chen2023videocrafter1}, to recent advanced models~\citep{bartal2024lumiere,  polyak2024moviegen, kong2024hunyuanvideo, openai2024sora, google2024veo2, wan2025wan, hacohen2025ltxvideo, seedance2026seedance, cloudflare2026happyhorse10i2v}, which have achieved substantial improvements in narrative coherence and visual consistency. However, these advances in generation quality have raised concerns about potential misuse and negative impacts on society. This motivates us to investigate whether existing detection methods can reliably identify increasingly realistic AI-generated videos, which can further push forward responsible video generation methods.


\paragraph{Detectors for AI-generated videos.}
Inspired by the detector taxonomy used in AIGVDBench~\citep{ma2026aigvdbench}, we broadly divide existing AI-generated video detectors into traditional discriminative approaches and MLLM-based approaches. Traditional approaches include video classification models, generated-image detection models, and generated-video detection models. Video classification models~\citep{feichtenhofer2019slowfast, bertasius2021space, liu2022video, tong2022videomae} directly treat detection as a binary video classification problem. Generated-image detectors~\citep{wang2020cnngenerated, ojha2023universal, tan2024npr, chen2025forgelens} mainly focus on image-level artifacts, while generated-video detectors~\citep{ma2024decof, he2024exposing, chen2024demamba, interno2025restrav} further capture temporal evidence across frames. Beyond these traditional approaches, MLLM-based detectors~\citep{wen2025busterx, li2025skyra} have gained increasing attention because they can provide detailed explanations beyond binary classification results. In this work, we evaluate representative detector families on real-world crises and other socially consequential events, and derive insights to guide their future design.

\begin{table}[t]
\centering
\footnotesize
\setlength{\tabcolsep}{4.4pt}
\renewcommand{\arraystretch}{1.08}

\resizebox{\linewidth}{!}{
\begin{tabular}{@{}lccccc@{}}
\toprule

\textbf{Benchmark}
& \makecell{\textbf{Real-event}\\\textbf{grounding}}
& \makecell{\textbf{Social-risk}\\\textbf{taxonomy}}
& \makecell{\textbf{Real-event-}\\\textbf{conditioned I2V}}
& \makecell{\textbf{Human-deceptive}\\\textbf{challenge set}}
& \makecell{\textbf{Social dissemination}\\\textbf{simulation}}
\\

\midrule

GVF~\citep{ma2024decof}
    & \nmark
    & \nmark
    & \nmark
    & \nmark
    & \nmark
\\

GenVideo~\citep{chen2024demamba}
    & \nmark
    & \nmark
    & \nmark
    & \nmark
    & \pmark
\\

GenVidBench~\citep{ni2026genvidbench}
    & \nmark
    & \nmark
    & \pmark
    & \nmark
    & \nmark
\\

GenBuster-Bench~\citep{wen2025busterx}
    & \nmark
    & \nmark
    & \nmark
    & \nmark
    & \pmark
\\

GenWorld~\citep{chen2025genworld}
    & \nmark
    & \nmark
    & \pmark
    & \nmark
    & \nmark
\\

AIGVDBench~\citep{ma2026aigvdbench}
    & \nmark
    & \nmark
    & \nmark
    & \nmark
    & \nmark
\\

Chameleon~\citep{liao2026chameleon}
    & \pmark
    & \nmark
    & \pmark
    & \nmark
    & \pmark
\\

\rowcolor{gray!12}
\textbf{RA-Bench}
    & \textbf{\cmark}
    & \textbf{\cmark}
    & \textbf{\cmark}
    & \textbf{\cmark}
    & \textbf{\cmark}
\\

\bottomrule
\end{tabular}
}

\vspace{-2pt}

\caption{
\textbf{Comparison of benchmark designs for AI-generated video detection.}
We compare how existing benchmarks handle real-event sources,
social-risk categories, real-event-conditioned generation, human deception, and
social dissemination simulations during dataset construction and evaluation.
RA-Bench also includes recent open- and closed-source video generators
and representative detector families.
}

\label{tab:benchmark_comparison}

\vspace{-1pt}

{\footnotesize
\raggedright
\textit{Criteria.}
A checkmark, circle, and cross denote full, partial, and absent coverage,
respectively. Full coverage requires traceable footage from identifiable
socially consequential events, a hierarchical social-risk taxonomy, I2V
conditioned on a matched event-source frame, a challenge set selected
through blind human authenticity judgments, and a social dissemination
simulation. For the final column, partial coverage denotes isolated social
dissemination operations rather than a complete social dissemination simulation. Dataset scale,
generator coverage, explanations, and source attribution are outside scope.
\par}

\vspace{-10pt}
\end{table}
\section{RA-Bench}
\label{sec:benchmark-construction}
\begin{figure*}[t]
    \centering
    \includegraphics[width=\textwidth]{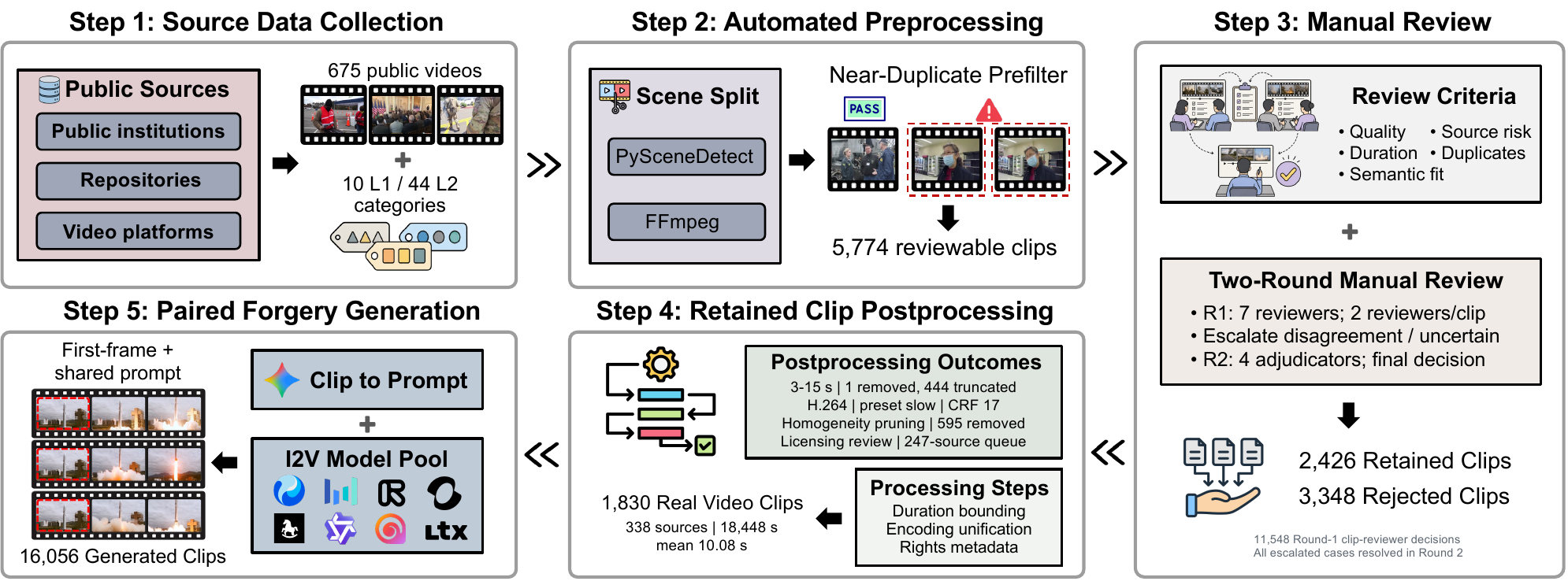}
    \caption{\textbf{Overview of the five-stage RA-Bench construction pipeline.} We collect public videos from real-world crises and other socially consequential events, segment them into scene-level clips and screen near-duplicates, conduct two-round manual review, and apply release-oriented postprocessing to form the real anchor set. Each retained anchor is then paired with image-to-video (I2V) generated counterparts using first-frame conditioning and a shared prompting pipeline.}
    \label{fig:RA-Bench-pipeline}
\end{figure*}

Unlike prior benchmarks~\citep{ma2026aigvdbench, wen2025busterx}, which collect seed video data from general-purpose web video corpora such as OpenVid-1M~\citep{nan2024openvid}, RA-Bench focuses on real videos from real-world crises and other socially consequential events, where forged videos can pose public risk. We first collect public source videos and organize them into social-risk categories (Section~\ref{sec:data-sources}). Then, each source video is segmented into scene-level clips and screened for near-duplicates (Section~\ref{sec:preprocessing}). The resulting clips are filtered through two rounds of human review (Section~\ref{sec:review}), after which postprocessing standardizes the retained clips into the final anchor set (Section~\ref{sec:postprocess}). Each real clip is then paired with image-to-video generated clips that form the generated side of the benchmark (Section~\ref{sec:generation}). The resulting RA-Bench consists of 1{,}830 real-video clips and 16{,}056 generated clips. An overview of the construction pipeline is shown in Figure~\ref{fig:RA-Bench-pipeline}, and the benchmark composition is summarized in Figure~\ref{fig:RA-Bench-overview}.


\subsection{Source Data Collection} \label{sec:data-sources}

To prepare source data for constructing the benchmark, we first collect 675 videos depicting real-world crises and other socially consequential events from public platforms. We then group these videos into 10 social-risk categories (L1) and 44 subcategories (L2). The L1 categories are weather and natural disasters, war and armed conflict, politics and governance, public safety, accidents and infrastructure failures, economic and social panic, public health, technology, space and exploration, and large public events. Details of the L2 categories are provided in Appendix~\ref{app:l2-taxonomy}. We do not impose a uniform per-category quota, since equal-size quotas would distort the natural prevalence of social-risk scenes and move the sample away from realistic deployment conditions. The L1 taxonomy and the released category distribution are shown in Figure~\ref{fig:RA-Bench-overview}(a,b).


Among the collected video sources, government and public-institution accounts form the largest group, while open-license repositories (\eg Wikimedia Commons) and public video platforms (\eg YouTube) account for most of the remaining sources. Appendix~\ref{app:source-inventory} lists the sources used for data collection. Since licensing conditions vary across sources, we record the redistribution rights for each source and use this information in the rights check described in Section~\ref{sec:postprocess}.




\subsection{Automated Preprocessing} \label{sec:preprocessing}
We segment each source video into scene-level clips for later review and image-to-video generation. Specifically, we detect scene cuts with PySceneDetect\footnote{\url{https://www.scenedetect.com/docs/latest/}} and split the videos at these cuts using FFmpeg\footnote{\url{https://ffmpeg.org/ffmpeg.html}}, producing 5{,}774 reviewable clips from 675 source videos. Since adjacent scenes in a source video tend to share similar framing and content, we run a near-duplicate prefilter on the clip pool. We adopt a ResNet-18~\citep{he2016deep} pretrained on ImageNet-1K~\citep{russakovsky2015imagenet} to encode each clip and compare cosine similarities between neighboring clips from the same source video. A pair is flagged when both frame- and clip-level similarities exceed the specified thresholds, and flagged pairs are shown to reviewers as duplicate warnings during manual review (Section~\ref{sec:review}). More details of the preprocessing are provided in Appendix~\ref{app:preprocessing-details}.





\subsection{Manual Review} \label{sec:review}
We conduct a two-round manual review of the 5{,}774 clips obtained after preprocessing. In Round~1, seven volunteers review the clips using predefined evaluation criteria for visual quality, semantic fit to the assigned subcategory, duration suitability, duplicate content, and source-related risks. Each clip is independently assessed by two reviewers, who assign one of three actions: \emph{retain}, \emph{reject}, or \emph{uncertain}. Clips with reviewer disagreement and those marked as uncertain are routed to Round~2. In Round~2, another four volunteers serve as adjudicators and make the final decision for each routed clip. After this process, 2{,}426 clips are retained as the standard sample pool and 3{,}348 are rejected. Further details of the manual review process are provided in Appendix~\ref{app:review-details}.



\subsection{Retained Clip Postprocessing} \label{sec:postprocess}
We postprocess the 2{,}426 retained clips to obtain the final real video set, using duration bounding, encoding unification, homogeneity pruning, and licensing review. Specifically, we first bound each clip to a 3--15 seconds window, discarding clips shorter than 3 seconds and truncating longer ones, thereby maintaining a moderate clip duration that balances sufficient information and acceptable cost for detection models. We then re-encode every clip with H.264 and later apply the same encoding to the generated clips. This reduces the risk that detectors rely on codec configurations as label cues. Subsequently, we remove redundant clips extracted from each source video to reduce near-duplicates from the same event. Finally, we review source-level licensing conditions and record the resulting redistribution rights as metadata for each clip. After these operations, the final set contains 1{,}830 clips from 338 unique source videos, totaling 18{,}448\,s (about 5.1 hours) with a mean duration of 10.08\,s. The category composition of the released set is shown in Figure~\ref{fig:RA-Bench-overview}(b). Further postprocessing details are provided in Appendix~\ref{app:postprocessing-details}.

\subsection{Paired Video Generation} \label{sec:generation}

\begin{figure}[t]
\centering
\includegraphics[width=\linewidth]{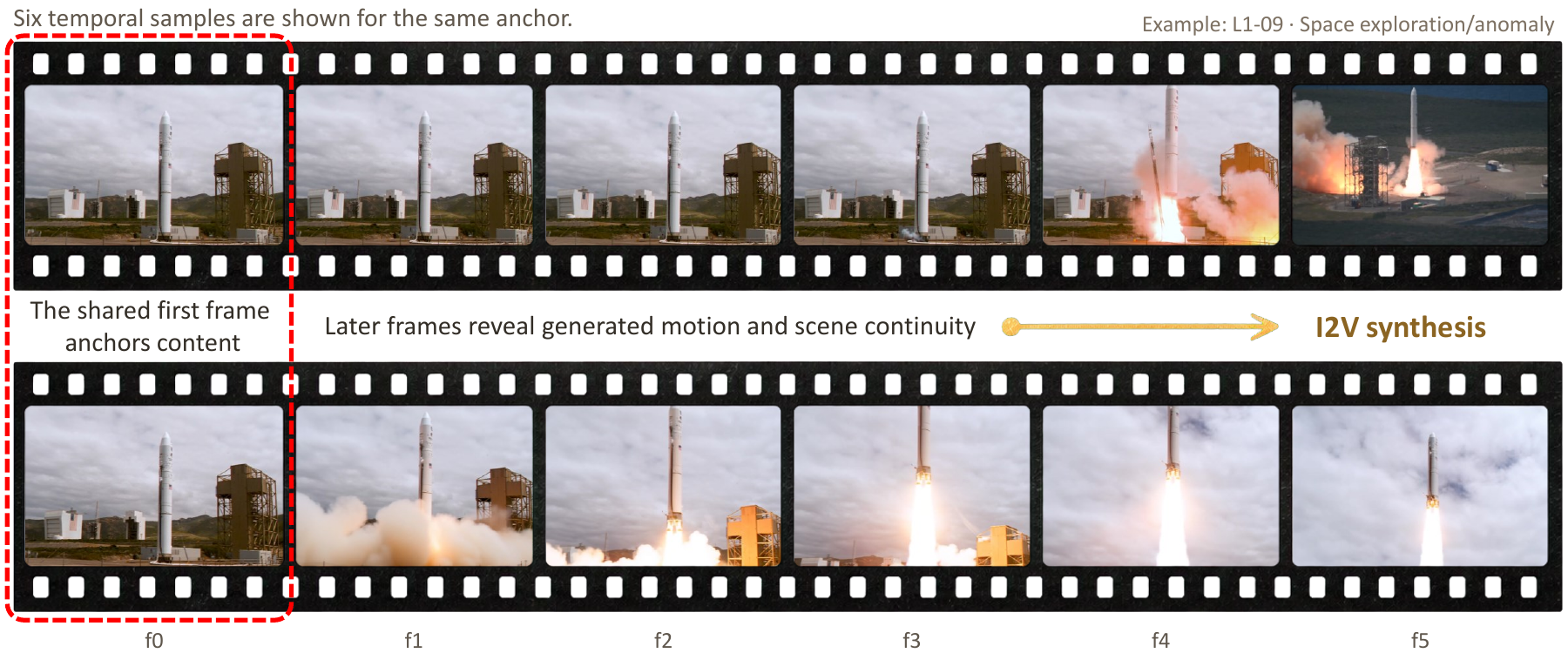}
\caption{\textbf{A real anchor and its generated counterpart for a representative
RA-Bench crisis event.} The top filmstrip shows the real space-launch anchor,
and the bottom shows the generated video, each represented by six uniformly
sampled frames (\texttt{f0}--\texttt{f5}). The red dashed box marks the shared
first frame used to condition the generator. The generated video continues from
this frame, preserving much of the scene layout, capture style, and lighting
while synthesizing new motion as the rocket leaves the pad. This example uses
the dynamic-duration Wan2.2 setting. Additional examples across crisis
categories and generators are provided in Figures~\ref{fig:gallery}
and~\ref{fig:cross_generator}.}
\label{fig:real_vs_gen_main}
\end{figure}

We pair each real clip with counterparts generated by image-to-video (I2V) models using a unified generation pipeline, as shown in Figure~\ref{fig:real_vs_gen_main}. Under this design, each real clip and its generated counterparts share the same scene semantics and are grounded in real-world crises and other socially consequential events.

We first caption each real clip with Gemini-3.1-Pro-Preview~\citep{google2026gemini31pro}, following structured video-captioning practice~\citep{ju2024miradata, luo2025any2caption}, and convert the caption into a prompt shared across all generators (Appendix~\ref{app:generation-details}). We then condition each generator on the obtained caption and the first frame of the real clip. This design yields visually plausible generated clips, increases benchmark difficulty, and reflects a common social media manipulation scenario in which a still image from a real-world crisis or another socially consequential event is used to fabricate a video that may pose public risk. Finally, we set the generated video length proportional to the duration of the corresponding real clip while constraining it to a 2--8\,s range. Each generator maps this target to a supported duration setting so that every generated clip remains within the range supported by most existing video generation models, as detailed in Appendix~\ref{app:generation-details}. This dynamic duration setting helps assess whether detection models learn duration-based shortcuts.

\begin{figure*}[t]
\centering
\begin{minipage}[t]{0.305\textwidth}
\centering
\includegraphics[width=\linewidth]{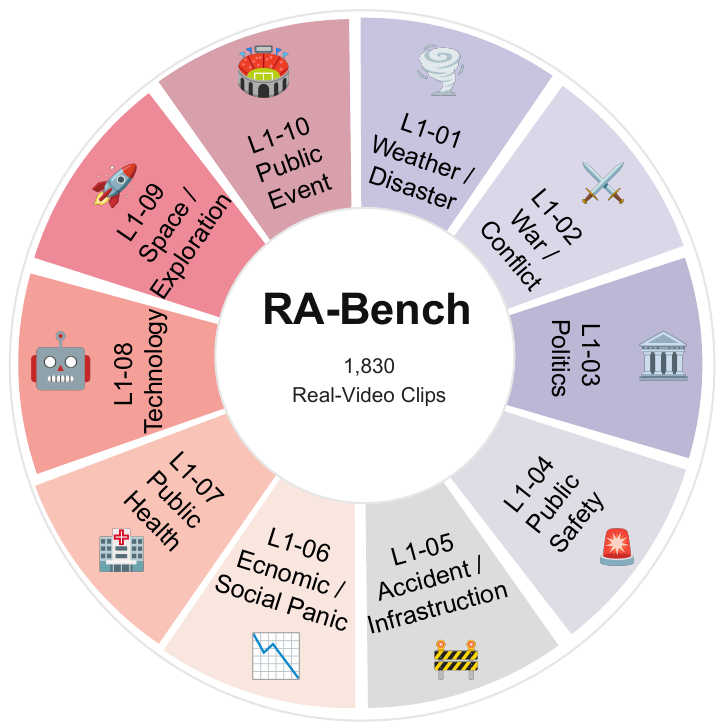}\\[-1mm]
{\small\textbf{(a) Real-event taxonomy}}
\end{minipage}
\hspace{0.006\textwidth}
\begin{minipage}[t]{0.365\textwidth}
\centering
\includegraphics[width=\linewidth]{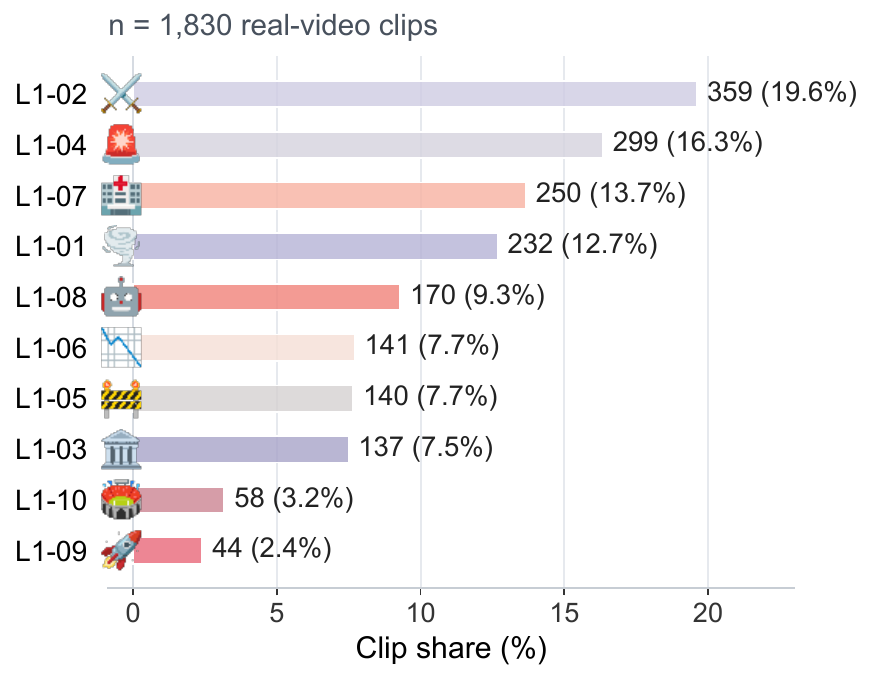}\\[-1mm]
{\small\textbf{(b) Category distribution}}
\end{minipage}
\hspace{-0.02\textwidth}
\begin{minipage}[t]{0.305\textwidth}
\centering
\includegraphics[width=\linewidth]{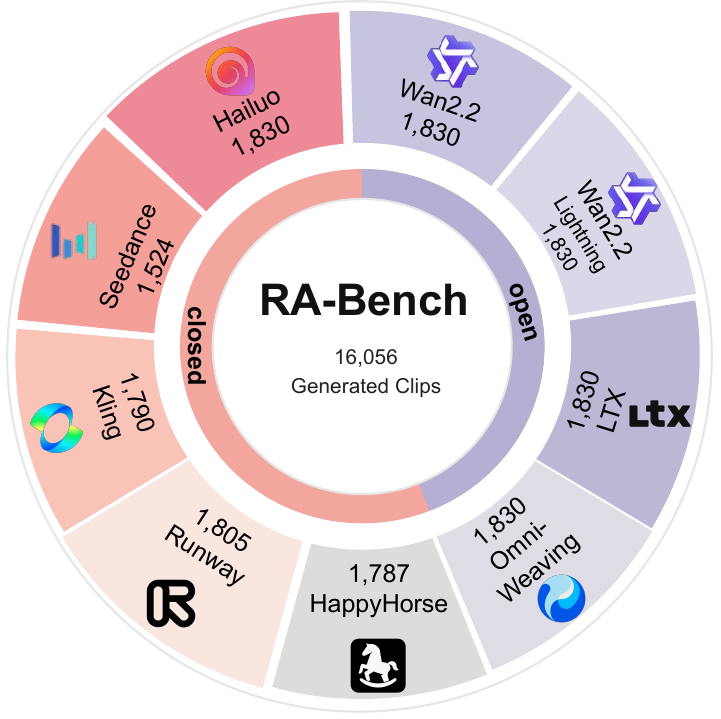}\\[-1mm]
{\small\textbf{(c) Generation sources}}
\end{minipage}
\caption{\textbf{Overview of RA-Bench.}
(a) The L1 real-event taxonomy used to organize the 1{,}830 real-video clips into 10 broad social-risk domains. The finer-grained L2 taxonomy is provided in Appendix~\ref{app:l2-taxonomy}.
(b) The clip distribution across the L1 domains.
(c) The image-to-video generation sources used to construct the generated-video set, grouped into open-source models and closed-source API providers; numbers indicate the generated clips included from each source.}
\label{fig:RA-Bench-overview}
\end{figure*}

After generation, we re-encode every generated clip using the same H.264 codec as the real clips. We preserve each generator's native resolution and frame rate, and remove audio from all clips. We generate clips with four open-source and five closed-source generators, as shown in Figure~\ref{fig:RA-Bench-overview}(c). The four open-source generators cover all 1{,}830 anchors. We submit the same anchors to each closed-source provider, but provider-specific content-safety filters reject some requests, resulting in smaller paired subsets. In total, RA-Bench contains 16{,}056 generated clips.

\section{Experiments and Analysis}
\label{sec:analysis}

We evaluate current methods for detecting AI-generated video on RA-Bench and analyze the factors associated with detection difficulty. Our experiments cover all four open-source and five closed-source generators in RA-Bench; the generators and paired clip counts are summarized in Figure~\ref{fig:RA-Bench-overview}(c). Each open-source generator is evaluated on 1{,}830 generated clips and their matched real anchors. For closed-source generators, provider-side moderation can reduce coverage, so each is evaluated on the returned generated clips and their corresponding real anchors. To examine sensitivity to generation duration, the detector-side experiments in Section~\ref{sec:detector-side} additionally include a fixed-duration Wan2.2 variant as a control for dynamic-duration generation. We mark this auxiliary control with an asterisk and exclude it from all benchmark-level averages. For continuous fake scores, we report paired AUC and TPR@5\%FPR; for discrete decisions, we report balanced accuracy (BAcc), macro-F1, and fake recall (FakeR). Our analysis has three parts. We first evaluate whether traditional detectors, zero-shot multimodal models, and MLLMs fine-tuned for AI-generated video detection generalize across generation sources (Section~\ref{sec:detector-side}). We then examine how generation quality, conditioning information, and sampling seeds relate to detectability (Section~\ref{sec:generation-side}). Finally, we compare detector behavior with human authenticity judgments and evaluate detector robustness during social dissemination (Section~\ref{sec:human-reliability}). Evaluation implementation details are summarized in Appendix~\ref{app:evaluation-implementation}; prompt templates and complete source-wise results are provided in the corresponding appendices.

\subsection{How Well Do Current Detectors Generalize?}
\label{sec:detector-side}

\subsubsection{Traditional Detectors Do Not Transfer Reliably to RA-Bench}
\label{sec:traditional}

Table~\ref{tab:traditional_main} reports results for seven traditional detectors across the nine RA-Bench generation sources and the fixed-duration Wan2.2 control. For context, the public-reference column reports AIGVDBench LTX-I2V results for six detectors~\citep{ma2026aigvdbench} and the VidProM result for ReStraV~\citep{interno2025restrav}; these AUCs range from 67.6\% to 98.6\%. On RA-Bench, the seven-detector mean falls to 50.9--57.3\% across the open-source generators and 43.9--54.0\% across the closed-source generators. For 26 of the 63 detector--source pairs, AUC is below 50\%, indicating that generated videos receive lower fake scores than real anchors in the same paired evaluation subset more often than the reverse.

\providecommand{\genlogo}[1]{%
  \raisebox{-0.25em}{%
    \includegraphics[height=1.15em]{figures/logos/#1}%
  }%
}
\providecommand{\genhead}[3]{%
  \makecell{%
    \genlogo{#1}\\[-0.2mm]
    \textbf{#2}\\
    \textbf{#3}%
  }%
}

\begin{table}[t]
\caption{\textbf{Traditional detector performance across the nine RA-Bench generation
sources} We report paired AUC and TPR at
5\% FPR (T@5\%) in \%. T@5\% is generated-video recall at an operating point
with 5\% FPR on the matched real videos. The public-reference column reports
AIGVDBench LTX-I2V AUC for six detectors and VidProM AUC for ReStraV; these are
contextual references rather than matched-domain baselines. Wan2.2 fixed$^{*}$
is an auxiliary control and is excluded from the
Open average. Spearman correlations use only the six detectors that share the
AIGVDBench reference.}
\label{tab:traditional_main}

\centering
\scriptsize
\setlength{\tabcolsep}{2.0pt}
\renewcommand{\arraystretch}{1.02}

\resizebox{\linewidth}{!}{%
\begin{tabular}{l l c *{5}{c} c *{5}{c} c}
\toprule
 & &
 \textbf{Public} &
 \multicolumn{5}{c}{\textbf{Open-source settings}} &
 \textbf{Open} &
 \multicolumn{5}{c}{\textbf{Closed-source generators}} &
 \textbf{Closed} \\

\cmidrule(lr){3-3}
\cmidrule(lr){4-8}
\cmidrule(lr){10-14}

\textbf{Detector} &
\textbf{Metric} &
\makecell{\textbf{High}\\\textbf{ref.}} &
\genhead{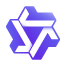}{Wan2.2}{dyn.} &
\genhead{wan.png}{Wan2.2}{fixed$^{*}$} &
\genhead{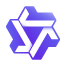}{Wan2.2}{Light.} &
\genhead{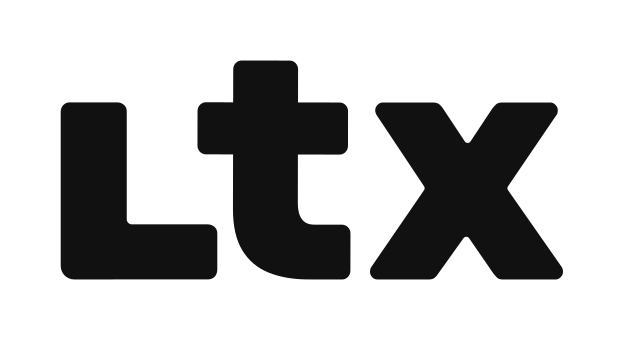}{LTX}{} &
\genhead{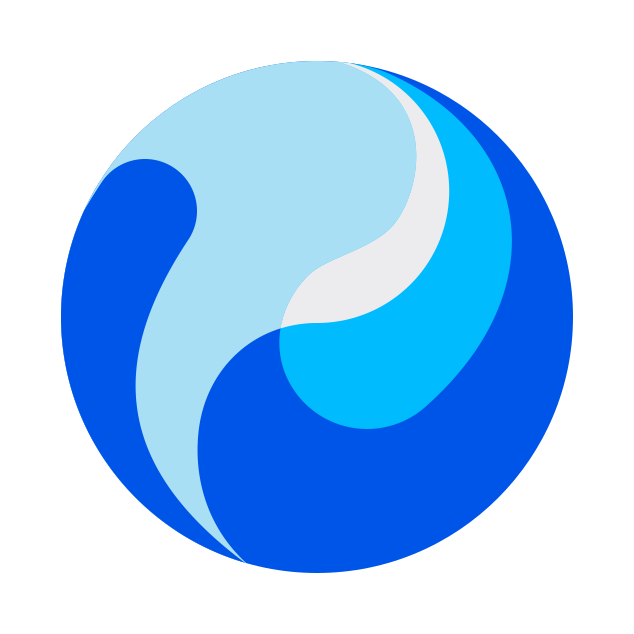}{Omni}{Weav.} &
\textbf{avg.} &
\genhead{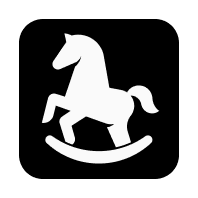}{Happy}{Horse} &
\genhead{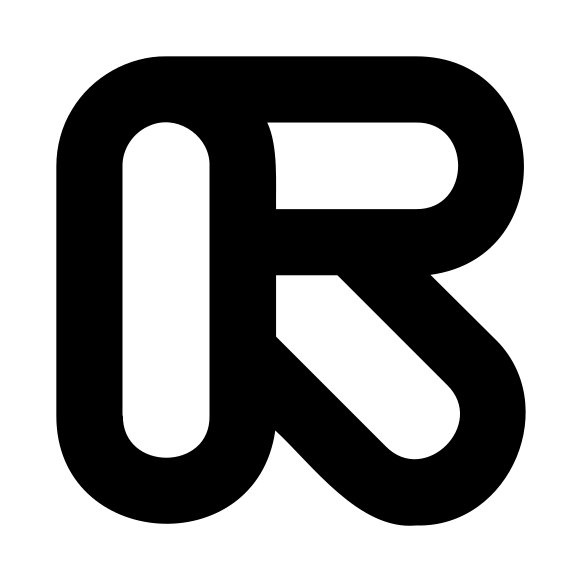}{Runway}{} &
\genhead{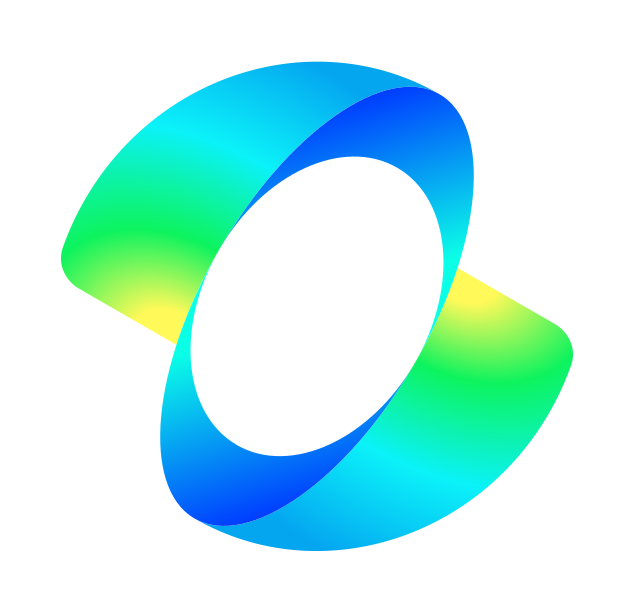}{Kling}{} &
\genhead{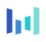}{Seedance}{2.0} &
\genhead{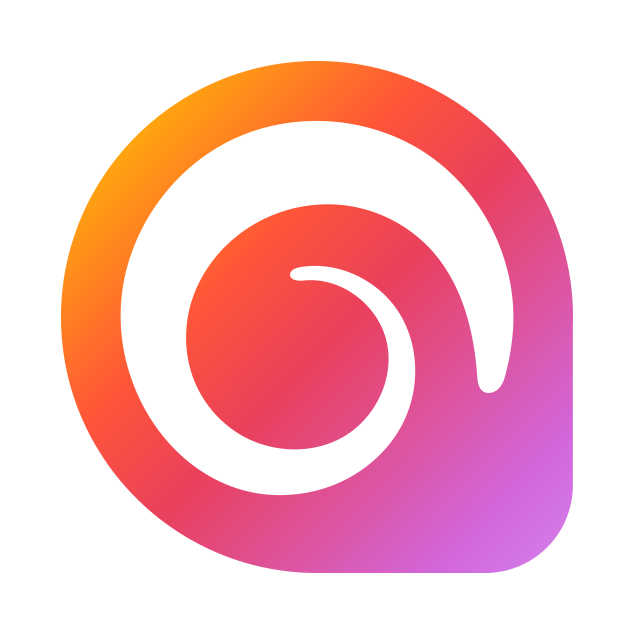}{Hailuo}{} &
\textbf{avg.} \\
\midrule

\rowcolor{blue!8}
\multicolumn{15}{l}{\textit{Image / frame-level detectors}} \\

CNNSpot & AUC
& 81.4 & 35.5 & 34.6 & 43.8 & 64.0 & 45.8 & 47.3
& 39.2 & 31.9 & 45.4 & 40.4 & 53.9 & 42.2 \\

& T@5\%
& \textcolor{black!45}{--}
& \textcolor{black!62}{2.3}
& \textcolor{black!62}{1.7}
& \textcolor{black!62}{3.0}
& \textcolor{black!62}{11.7}
& \textcolor{black!62}{3.5}
& \textcolor{black!62}{5.1}
& \textcolor{black!62}{2.8}
& \textcolor{black!62}{1.6}
& \textcolor{black!62}{3.9}
& \textcolor{black!62}{2.8}
& \textcolor{black!62}{5.8}
& \textcolor{black!62}{3.4} \\

NPR & AUC
& 67.6 & 50.6 & 49.9 & 55.2 & 58.6 & 50.9 & 53.8
& 54.6 & 52.3 & 44.3 & 48.7 & 54.6 & 50.9 \\

& T@5\%
& \textcolor{black!45}{--}
& \textcolor{black!62}{3.8}
& \textcolor{black!62}{4.0}
& \textcolor{black!62}{4.6}
& \textcolor{black!62}{4.5}
& \textcolor{black!62}{5.6}
& \textcolor{black!62}{4.6}
& \textcolor{black!62}{4.5}
& \textcolor{black!62}{3.4}
& \textcolor{black!62}{3.2}
& \textcolor{black!62}{3.9}
& \textcolor{black!62}{5.7}
& \textcolor{black!62}{4.2} \\

UnivFD & AUC
& 89.9 & 32.7 & 33.4 & 43.0 & 51.2 & 39.9 & 41.7
& 43.0 & 38.4 & 39.1 & 35.0 & 44.0 & 39.9 \\

& T@5\%
& \textcolor{black!45}{--}
& \textcolor{black!62}{0.4}
& \textcolor{black!62}{0.5}
& \textcolor{black!62}{2.1}
& \textcolor{black!62}{0.9}
& \textcolor{black!62}{0.5}
& \textcolor{black!62}{1.0}
& \textcolor{black!62}{1.7}
& \textcolor{black!62}{0.5}
& \textcolor{black!62}{1.1}
& \textcolor{black!62}{0.6}
& \textcolor{black!62}{1.3}
& \textcolor{black!62}{1.1} \\

ForgeLens & AUC
& 92.9
& 60.0
& 61.0
& \textbf{69.6}
& \textbf{64.1}
& 63.0
& \textbf{64.2}
& \textbf{64.9}
& \textbf{59.5}
& 59.2
& 48.1
& 61.7
& 58.7 \\

& T@5\%
& \textcolor{black!45}{--}
& \textcolor{black!62}{19.0}
& \textcolor{black!62}{18.7}
& \textcolor{black!62}{25.2}
& \textcolor{black!62}{11.5}
& \textcolor{black!62}{17.6}
& \textcolor{black!62}{18.3}
& \textcolor{black!62}{18.7}
& \textcolor{black!62}{8.0}
& \textcolor{black!62}{9.6}
& \textcolor{black!62}{3.1}
& \textcolor{black!62}{8.9}
& \textcolor{black!62}{9.7} \\

\midrule
\rowcolor{cyan!8}
\multicolumn{15}{l}{\textit{Video / temporal-level detectors}} \\

DeCoF & AUC
& 81.5
& \textbf{62.5}
& \textbf{63.2}
& 56.7
& 62.5
& \textbf{72.1}
& 63.4
& 58.5
& 55.2
& 57.2
& \textbf{60.6}
& \textbf{63.2}
& \textbf{58.9} \\

& T@5\%
& \textcolor{black!45}{--}
& \textcolor{black!62}{3.9}
& \textcolor{black!62}{4.7}
& \textcolor{black!62}{2.8}
& \textcolor{black!62}{2.3}
& \textcolor{black!62}{6.7}
& \textcolor{black!62}{3.9}
& \textcolor{black!62}{1.8}
& \textcolor{black!62}{1.8}
& \textcolor{black!62}{0.5}
& \textcolor{black!62}{1.0}
& \textcolor{black!62}{4.8}
& \textcolor{black!62}{2.0} \\

D3 & AUC
& 77.7 & 55.4 & 55.7 & 49.1 & 58.0 & 52.9 & 53.8
& 28.0 & 24.4 & 28.9 & 50.4 & 44.0 & 35.1 \\

& T@5\%
& \textcolor{black!45}{--}
& \textcolor{black!62}{6.4}
& \textcolor{black!62}{5.5}
& \textcolor{black!62}{2.1}
& \textcolor{black!62}{6.2}
& \textcolor{black!62}{3.3}
& \textcolor{black!62}{4.5}
& \textcolor{black!62}{6.3}
& \textcolor{black!62}{1.8}
& \textcolor{black!62}{6.2}
& \textcolor{black!62}{10.6}
& \textcolor{black!62}{1.9}
& \textcolor{black!62}{5.3} \\

ReStraV & AUC
& 98.6 & 59.3 & 51.2 & 62.4 & 42.5 & 68.3 & 58.1
& 55.7 & 45.8 & \textbf{67.4} & 49.3 & 56.4 & 54.9 \\

& T@5\%
& \textcolor{black!45}{--}
& \textcolor{black!62}{8.5}
& \textcolor{black!62}{6.7}
& \textcolor{black!62}{10.1}
& \textcolor{black!62}{4.6}
& \textcolor{black!62}{15.0}
& \textcolor{black!62}{9.6}
& \textcolor{black!62}{8.6}
& \textcolor{black!62}{2.7}
& \textcolor{black!62}{0.0}
& \textcolor{black!62}{6.0}
& \textcolor{black!62}{8.0}
& \textcolor{black!62}{5.0} \\

\midrule
\rowcolor{gray!12}
\multicolumn{2}{l}{\textbf{7-det. mean AUC}}
& 84.2 & 50.9 & 49.9 & 54.2 & 57.3 & 56.1 & 54.6
& 49.1 & 43.9 & 48.8 & 47.5 & 54.0 & 48.7 \\

\rowcolor{gray!8}
\multicolumn{2}{l}{\textbf{7-det. mean T@5\%}}
& -- & 6.3 & 6.0 & 7.1 & 6.0 & 7.5 & 6.7
& 6.4 & 2.8 & 3.5 & 4.0 & 5.2 & 4.4 \\

\rowcolor{gray!6}
\multicolumn{2}{l}{\textbf{Spearman vs ref.}}
& -- & 0.14 & 0.14 & 0.20 & 0.31 & 0.14 & 0.20
& 0.54 & 0.54 & 0.54 & $-$0.37 & 0.31 & 0.31 \\

\bottomrule
\end{tabular}%
}
\end{table}

\begin{figure}[t]
    \centering
    \includegraphics[width=0.90\linewidth]{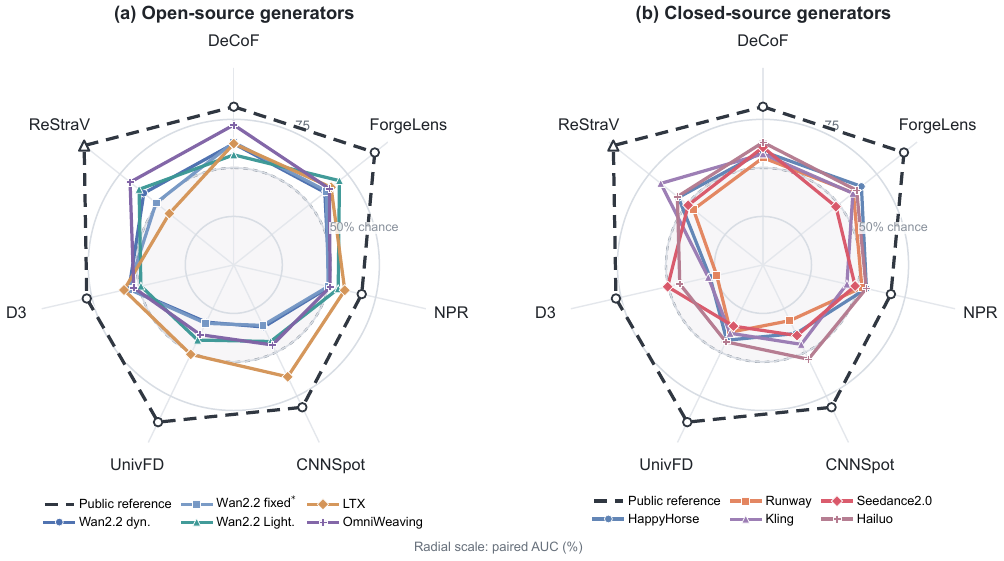}
    \caption{\textbf{Source-specific AUC profiles for the seven traditional detectors.}
    Each spoke represents one detector, and each polygon represents one
    generation setting. The outer dashed contour shows the public-reference
    result for each detector (AIGVDBench LTX-I2V for six detectors and VidProM
    for ReStraV), while the faint inner ring marks 50\% AUC. (a) The four
    open-source generators in RA-Bench and the fixed-duration Wan2.2 control.
    (b) The five closed-source generators in RA-Bench.}
    \label{fig:radar_main}
\end{figure}

Public-reference rankings also fail to transfer to RA-Bench. Among the six detectors evaluated on the same AIGVDBench LTX-I2V reference, the public ranking has a Spearman correlation of only \(0.26\) with the ranking by mean AUC across the nine RA-Bench sources. UnivFD moves from second to sixth, whereas NPR moves from sixth to third. The leading detector also changes across generators: DeCoF and ForgeLens each rank first on four sources, while ReStraV ranks first on Kling. The loss is therefore not a uniform decrease from the public references; it changes which detector appears strongest.

The failure pattern also differs across detectors and sources. D3 obtains 53.8\% mean AUC across the open-source generators but only 35.1\% across the closed-source generators. CNNSpot and UnivFD fall below 50\% AUC on seven and eight of the nine RA-Bench sources, respectively, despite public-reference AUCs of 81.4\% and 89.9\%. Performance remains weak when false positives are constrained: at 5\% FPR, the seven-detector mean identifies only 6.0--7.5\% of generated videos from the open-source generators and 2.8--6.4\% from the closed-source generators. Figure~\ref{fig:radar_main} visualizes the gaps from the public references and the crossings among source profiles. Reference alignment, rank transfer, and additional operating points at 1\% FPR and 95\% TPR are reported in Appendix~\ref{app:traditional-extended}.

\paragraph{Temporal reallocation yields only modest gains.}
One possible explanation for the weak results is that sparse uniform sampling
misses brief local inconsistencies. We test this explanation for five
sparse-frame detectors by comparing Uniform-8 with Global--Local-8 under the
same eight-frame budget. Global--Local-8 combines four uniformly spaced frames
with four consecutive frames centered on the strongest temporal-change
response; detector weights, visual preprocessing, and score aggregation remain
fixed. The routing procedure is described in
Appendix~\ref{app:evaluation-implementation}.

\begin{figure}[H]
    \centering
    \includegraphics[width=\linewidth]
    {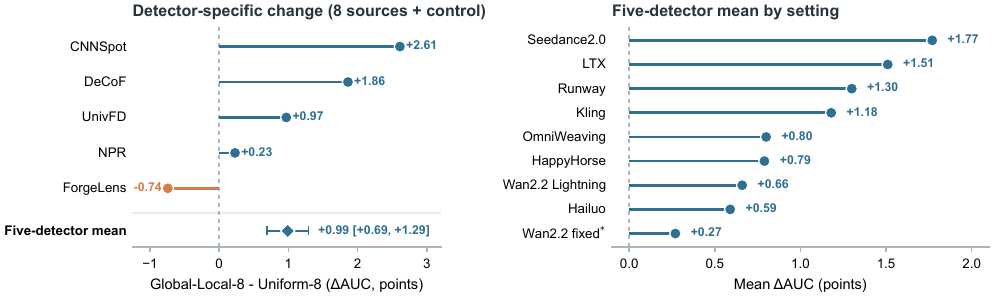}
    \caption{\textbf{Effect of temporal allocation under a fixed eight-frame
    budget.} (a) AUC change from Uniform-8 to Global--Local-8 for each of five
    sparse-frame detectors, averaged over eight fixed-length RA-Bench sources and the
    fixed-duration Wan2.2 control; the diamond and interval show the
    five-detector mean and source-block bootstrap 95\% confidence interval.
    (b) The corresponding five-detector mean for each evaluated setting. The
    ablation excludes dynamic-duration Wan2.2. Positive values favor
    Global--Local-8.}
    \label{fig:sampling-performance}
\end{figure}

Across the eight fixed-length RA-Bench sources and the fixed-duration Wan2.2
control included in this ablation, Global--Local-8 increases mean AUC by only
\(0.99\) points over Uniform-8 (source-block bootstrap 95\% CI,
\([0.69,1.29]\); Figure~\ref{fig:sampling-performance}). Four of the five
detectors improve, with the largest gains for CNNSpot (\(+2.61\) points) and
DeCoF (\(+1.86\) points), while ForgeLens decreases by \(0.74\) points. The
detector-averaged change is positive for all nine evaluated settings but ranges
from only \(+0.27\) to \(+1.77\) points. Reallocating the same frames recovers
some local temporal evidence, but it does not change the central result: high
public-reference AUC neither translates into reliable detection nor identifies
a consistently strong detector across RA-Bench sources.

\FloatBarrier

\paragraph{Takeaway.}
Traditional detector AUC falls from 67.6--98.6\% in public references to source-level means of 43.9--57.3\% on RA-Bench, with different detectors leading on different generators.

\subsubsection{Zero-Shot Multimodal Models Remain Unreliable Across Prompts and Sources}
\label{sec:zeroshot}

Table~\ref{tab:zeroshot_main} evaluates representative zero-shot multimodal
models under Binary, Diagnostic, and Rating prompts; their output formats and
complete templates are provided in Appendix~\ref{app:zeroshot-prompts}.
Overall detection performance remains limited. Qwen3.5-27B obtains 53.2/46.7
BAcc/macro-F1 under Binary and 51.3/37.2 under Diagnostic, while
Qwen3.5-122B-A10B reaches 53.0/45.7 and 54.8/50.8, respectively.
Gemini-3.1-Pro-Preview is the strongest evaluated model, reaching 63.4/62.9
under Binary, 63.1/62.5 under Diagnostic, and 63.6/63.0 AUC/verdict macro-F1
under Rating. Its Binary BAcc nevertheless ranges from 54.3 on Seedance2.0 to
74.5 on LTX. No evaluated model therefore combines strong overall performance
with stable behavior across prompt formats and RA-Bench sources.

\providecommand{\genlogo}[1]{\raisebox{-0.25em}{\includegraphics[height=1.15em]{figures/logos/#1}}}
\providecommand{\genhead}[3]{\makecell{\genlogo{#1}\\[-0.2mm]\textbf{#2}\\\textbf{#3}}}

\begin{table}[H]
\caption{\textbf{Source-specific zero-shot multimodal model results on the nine RA-Bench generation sources.} Binary and Diagnostic cells report BAcc/macro-F1, while Rating cells report paired AUC/verdict macro-F1 (top/bottom). Values are in \%. Open and Closed are source-equal averages over the four open-source and five closed-source RA-Bench generators, respectively; Wan2.2 fixed$^{*}$ is excluded from the Open average.}
\label{tab:zeroshot_main}
\centering
\scriptsize
\setlength{\tabcolsep}{2.4pt}
\renewcommand{\arraystretch}{1.00}
\resizebox{\textwidth}{!}{
\begin{tabular}{l l *{5}{c} c *{5}{c} c}
\toprule
 & & \multicolumn{5}{c}{\textbf{Open-source settings}} & \textbf{Open} & \multicolumn{5}{c}{\textbf{Closed-source generators}} & \textbf{Closed} \\
\cmidrule(lr){3-7}\cmidrule(lr){9-13}
\textbf{Model} & \textbf{Prompt} & \genhead{wan.png}{Wan2.2}{dyn.} & \genhead{wan.png}{Wan2.2}{fixed$^{*}$} & \genhead{wan_lightning.png}{Wan2.2}{Light.} & \genhead{ltx.png}{LTX}{} & \genhead{omni.png}{Omni}{Weav.} & \textbf{avg.} & \genhead{happyhorse.png}{Happy}{Horse} & \genhead{runway.png}{Runway}{} & \genhead{kling.png}{Kling}{} & \genhead{seedance.png}{Seedance}{2.0} & \genhead{hailuo.png}{Hailuo}{} & \textbf{avg.} \\
\midrule
\multicolumn{14}{l}{\textit{Discrete classification prompts (BAcc/macro-F1)}} \\
Qwen3.5-27B & Binary & \makecell{51.3\\[-0.45mm]\textcolor{black!62}{43.8}} & \makecell{51.2\\[-0.45mm]\textcolor{black!62}{43.6}} & \makecell{55.3\\[-0.45mm]\textcolor{black!62}{49.9}} & \makecell{59.7\\[-0.45mm]\textcolor{black!62}{56.2}} & \makecell{54.8\\[-0.45mm]\textcolor{black!62}{49.2}} & \makecell{55.3\\[-0.45mm]\textcolor{black!62}{49.8}} & \makecell{51.6\\[-0.45mm]\textcolor{black!62}{44.2}} & \makecell{55.3\\[-0.45mm]\textcolor{black!62}{50.0}} & \makecell{50.0\\[-0.45mm]\textcolor{black!62}{41.4}} & \makecell{49.8\\[-0.45mm]\textcolor{black!62}{41.7}} & \makecell{51.4\\[-0.45mm]\textcolor{black!62}{43.9}} & \makecell{51.6\\[-0.45mm]\textcolor{black!62}{44.2}} \\
 & Diagnostic & \makecell{50.9\\[-0.45mm]\textcolor{black!62}{36.5}} & \makecell{50.7\\[-0.45mm]\textcolor{black!62}{36.1}} & \makecell{51.8\\[-0.45mm]\textcolor{black!62}{38.4}} & \makecell{53.3\\[-0.45mm]\textcolor{black!62}{41.2}} & \makecell{51.7\\[-0.45mm]\textcolor{black!62}{38.2}} & \makecell{51.9\\[-0.45mm]\textcolor{black!62}{38.6}} & \makecell{50.8\\[-0.45mm]\textcolor{black!62}{36.4}} & \makecell{50.9\\[-0.45mm]\textcolor{black!62}{36.5}} & \makecell{50.5\\[-0.45mm]\textcolor{black!62}{35.6}} & \makecell{50.2\\[-0.45mm]\textcolor{black!62}{35.0}} & \makecell{51.1\\[-0.45mm]\textcolor{black!62}{36.9}} & \makecell{50.7\\[-0.45mm]\textcolor{black!62}{36.1}} \\
\addlinespace[1pt]
\makecell[l]{Qwen3.5-122B\\[-0.5mm]-A10B} & Binary & \makecell{51.6\\[-0.45mm]\textcolor{black!62}{43.5}} & \makecell{50.7\\[-0.45mm]\textcolor{black!62}{41.9}} & \makecell{52.7\\[-0.45mm]\textcolor{black!62}{45.3}} & \makecell{58.3\\[-0.45mm]\textcolor{black!62}{53.8}} & \makecell{54.1\\[-0.45mm]\textcolor{black!62}{47.5}} & \makecell{54.2\\[-0.45mm]\textcolor{black!62}{47.5}} & \makecell{51.8\\[-0.45mm]\textcolor{black!62}{43.8}} & \makecell{50.2\\[-0.45mm]\textcolor{black!62}{41.1}} & \makecell{53.4\\[-0.45mm]\textcolor{black!62}{46.4}} & \makecell{52.4\\[-0.45mm]\textcolor{black!62}{44.9}} & \makecell{52.3\\[-0.45mm]\textcolor{black!62}{44.6}} & \makecell{52.0\\[-0.45mm]\textcolor{black!62}{44.2}} \\
 & Diagnostic & \makecell{54.1\\[-0.45mm]\textcolor{black!62}{50.2}} & \makecell{53.9\\[-0.45mm]\textcolor{black!62}{49.8}} & \makecell{55.7\\[-0.45mm]\textcolor{black!62}{52.3}} & \makecell{65.6\\[-0.45mm]\textcolor{black!62}{64.7}} & \makecell{60.2\\[-0.45mm]\textcolor{black!62}{58.2}} & \makecell{58.9\\[-0.45mm]\textcolor{black!62}{56.3}} & \makecell{51.6\\[-0.45mm]\textcolor{black!62}{46.5}} & \makecell{51.7\\[-0.45mm]\textcolor{black!62}{46.6}} & \makecell{52.1\\[-0.45mm]\textcolor{black!62}{47.1}} & \makecell{49.7\\[-0.45mm]\textcolor{black!62}{44.6}} & \makecell{52.2\\[-0.45mm]\textcolor{black!62}{47.5}} & \makecell{51.5\\[-0.45mm]\textcolor{black!62}{46.4}} \\
\addlinespace[1pt]
\makecell[l]{Qwen3.7-Plus\\[-0.5mm](thinking)} & Binary & \makecell{53.7\\[-0.45mm]\textcolor{black!62}{45.8}} & \makecell{53.6\\[-0.45mm]\textcolor{black!62}{45.7}} & \makecell{55.5\\[-0.45mm]\textcolor{black!62}{48.8}} & \makecell{64.6\\[-0.45mm]\textcolor{black!62}{61.8}} & \makecell{54.8\\[-0.45mm]\textcolor{black!62}{47.7}} & \makecell{57.2\\[-0.45mm]\textcolor{black!62}{51.0}} & \makecell{53.7\\[-0.45mm]\textcolor{black!62}{45.9}} & \makecell{54.9\\[-0.45mm]\textcolor{black!62}{47.8}} & \makecell{53.9\\[-0.45mm]\textcolor{black!62}{46.2}} & \makecell{51.2\\[-0.45mm]\textcolor{black!62}{41.5}} & \makecell{52.9\\[-0.45mm]\textcolor{black!62}{44.5}} & \makecell{53.3\\[-0.45mm]\textcolor{black!62}{45.2}} \\
 & Diagnostic & \makecell{54.2\\[-0.45mm]\textcolor{black!62}{47.7}} & \makecell{53.6\\[-0.45mm]\textcolor{black!62}{46.7}} & \makecell{55.7\\[-0.45mm]\textcolor{black!62}{50.0}} & \makecell{64.5\\[-0.45mm]\textcolor{black!62}{62.1}} & \makecell{57.2\\[-0.45mm]\textcolor{black!62}{52.2}} & \makecell{57.9\\[-0.45mm]\textcolor{black!62}{53.0}} & \makecell{53.4\\[-0.45mm]\textcolor{black!62}{46.4}} & \makecell{54.0\\[-0.45mm]\textcolor{black!62}{47.5}} & \makecell{54.2\\[-0.45mm]\textcolor{black!62}{47.6}} & \makecell{51.9\\[-0.45mm]\textcolor{black!62}{43.9}} & \makecell{52.9\\[-0.45mm]\textcolor{black!62}{45.6}} & \makecell{53.3\\[-0.45mm]\textcolor{black!62}{46.2}} \\
\addlinespace[1pt]
\makecell[l]{Gemini-3.1-Pro\\[-0.5mm]-Preview} & Binary & \makecell{63.4\\[-0.45mm]\textcolor{black!62}{63.1}} & \makecell{65.8\\[-0.45mm]\textcolor{black!62}{65.7}} & \makecell{61.6\\[-0.45mm]\textcolor{black!62}{61.1}} & \makecell{74.5\\[-0.45mm]\textcolor{black!62}{74.5}} & \makecell{66.6\\[-0.45mm]\textcolor{black!62}{66.5}} & \makecell{66.5\\[-0.45mm]\textcolor{black!62}{66.3}} & \makecell{62.5\\[-0.45mm]\textcolor{black!62}{62.1}} & \makecell{67.8\\[-0.45mm]\textcolor{black!62}{67.7}} & \makecell{60.1\\[-0.45mm]\textcolor{black!62}{59.5}} & \makecell{54.3\\[-0.45mm]\textcolor{black!62}{52.6}} & \makecell{59.9\\[-0.45mm]\textcolor{black!62}{59.2}} & \makecell{60.9\\[-0.45mm]\textcolor{black!62}{60.2}} \\
 & Diagnostic & \makecell{63.8\\[-0.45mm]\textcolor{black!62}{63.4}} & \makecell{63.7\\[-0.45mm]\textcolor{black!62}{63.3}} & \makecell{61.8\\[-0.45mm]\textcolor{black!62}{61.2}} & \makecell{74.5\\[-0.45mm]\textcolor{black!62}{74.5}} & \makecell{65.1\\[-0.45mm]\textcolor{black!62}{64.8}} & \makecell{66.3\\[-0.45mm]\textcolor{black!62}{65.9}} & \makecell{61.5\\[-0.45mm]\textcolor{black!62}{60.9}} & \makecell{67.5\\[-0.45mm]\textcolor{black!62}{67.3}} & \makecell{60.2\\[-0.45mm]\textcolor{black!62}{59.4}} & \makecell{53.8\\[-0.45mm]\textcolor{black!62}{51.6}} & \makecell{60.2\\[-0.45mm]\textcolor{black!62}{59.4}} & \makecell{60.6\\[-0.45mm]\textcolor{black!62}{59.7}} \\
\addlinespace[1pt]
GPT-5.5 & Binary & \makecell{50.8\\[-0.45mm]\textcolor{black!62}{35.7}} & \makecell{51.4\\[-0.45mm]\textcolor{black!62}{37.0}} & \makecell{51.5\\[-0.45mm]\textcolor{black!62}{37.1}} & \makecell{56.9\\[-0.45mm]\textcolor{black!62}{47.6}} & \makecell{50.9\\[-0.45mm]\textcolor{black!62}{35.9}} & \makecell{52.5\\[-0.45mm]\textcolor{black!62}{39.1}} & \makecell{51.7\\[-0.45mm]\textcolor{black!62}{37.5}} & \makecell{51.6\\[-0.45mm]\textcolor{black!62}{37.5}} & \makecell{51.2\\[-0.45mm]\textcolor{black!62}{36.5}} & \makecell{50.6\\[-0.45mm]\textcolor{black!62}{35.2}} & \makecell{51.1\\[-0.45mm]\textcolor{black!62}{36.3}} & \makecell{51.2\\[-0.45mm]\textcolor{black!62}{36.6}} \\
 & Diagnostic & \makecell{50.7\\[-0.45mm]\textcolor{black!62}{35.1}} & \makecell{51.0\\[-0.45mm]\textcolor{black!62}{35.9}} & \makecell{51.1\\[-0.45mm]\textcolor{black!62}{36.0}} & \makecell{54.7\\[-0.45mm]\textcolor{black!62}{43.2}} & \makecell{50.7\\[-0.45mm]\textcolor{black!62}{35.2}} & \makecell{51.8\\[-0.45mm]\textcolor{black!62}{37.4}} & \makecell{51.0\\[-0.45mm]\textcolor{black!62}{35.9}} & \makecell{50.7\\[-0.45mm]\textcolor{black!62}{35.3}} & \makecell{51.0\\[-0.45mm]\textcolor{black!62}{35.8}} & \makecell{50.4\\[-0.45mm]\textcolor{black!62}{34.7}} & \makecell{50.8\\[-0.45mm]\textcolor{black!62}{35.5}} & \makecell{50.8\\[-0.45mm]\textcolor{black!62}{35.4}} \\
\midrule
\multicolumn{14}{l}{\textit{Continuous rating prompt with explicit verdict (AUC/macro-F1)}} \\
Qwen3.5-27B & Rating & \makecell{52.1\\[-0.45mm]\textcolor{black!62}{34.5}} & \makecell{52.0\\[-0.45mm]\textcolor{black!62}{34.3}} & \makecell{53.4\\[-0.45mm]\textcolor{black!62}{35.4}} & \makecell{53.4\\[-0.45mm]\textcolor{black!62}{37.6}} & \makecell{50.1\\[-0.45mm]\textcolor{black!62}{34.6}} & \makecell{52.3\\[-0.45mm]\textcolor{black!62}{35.5}} & \makecell{50.9\\[-0.45mm]\textcolor{black!62}{34.8}} & \makecell{53.0\\[-0.45mm]\textcolor{black!62}{35.0}} & \makecell{49.6\\[-0.45mm]\textcolor{black!62}{34.4}} & \makecell{47.7\\[-0.45mm]\textcolor{black!62}{33.7}} & \makecell{50.4\\[-0.45mm]\textcolor{black!62}{35.0}} & \makecell{50.3\\[-0.45mm]\textcolor{black!62}{34.6}} \\
\addlinespace[1pt]
\makecell[l]{Qwen3.5-122B\\[-0.5mm]-A10B} & Rating & \makecell{56.3\\[-0.45mm]\textcolor{black!62}{55.8}} & \makecell{56.0\\[-0.45mm]\textcolor{black!62}{55.6}} & \makecell{57.2\\[-0.45mm]\textcolor{black!62}{57.2}} & \makecell{65.8\\[-0.45mm]\textcolor{black!62}{64.3}} & \makecell{60.4\\[-0.45mm]\textcolor{black!62}{59.5}} & \makecell{59.9\\[-0.45mm]\textcolor{black!62}{59.2}} & \makecell{53.2\\[-0.45mm]\textcolor{black!62}{53.4}} & \makecell{57.1\\[-0.45mm]\textcolor{black!62}{56.4}} & \makecell{54.2\\[-0.45mm]\textcolor{black!62}{54.2}} & \makecell{45.4\\[-0.45mm]\textcolor{black!62}{46.8}} & \makecell{52.7\\[-0.45mm]\textcolor{black!62}{52.8}} & \makecell{52.5\\[-0.45mm]\textcolor{black!62}{52.7}} \\
\addlinespace[1pt]
\makecell[l]{Qwen3.7-Plus\\[-0.5mm](thinking)} & Rating & \makecell{57.8\\[-0.45mm]\textcolor{black!62}{48.5}} & \makecell{57.3\\[-0.45mm]\textcolor{black!62}{47.7}} & \makecell{59.6\\[-0.45mm]\textcolor{black!62}{51.6}} & \makecell{69.3\\[-0.45mm]\textcolor{black!62}{62.7}} & \makecell{63.8\\[-0.45mm]\textcolor{black!62}{50.8}} & \makecell{62.6\\[-0.45mm]\textcolor{black!62}{53.4}} & \makecell{55.1\\[-0.45mm]\textcolor{black!62}{46.9}} & \makecell{56.6\\[-0.45mm]\textcolor{black!62}{50.7}} & \makecell{58.6\\[-0.45mm]\textcolor{black!62}{49.5}} & \makecell{51.0\\[-0.45mm]\textcolor{black!62}{43.3}} & \makecell{55.6\\[-0.45mm]\textcolor{black!62}{46.0}} & \makecell{55.4\\[-0.45mm]\textcolor{black!62}{47.3}} \\
\addlinespace[1pt]
\makecell[l]{Gemini-3.1-Pro\\[-0.5mm]-Preview} & Rating & \makecell{62.5\\[-0.45mm]\textcolor{black!62}{64.8}} & \makecell{63.6\\[-0.45mm]\textcolor{black!62}{65.7}} & \makecell{62.2\\[-0.45mm]\textcolor{black!62}{61.2}} & \makecell{77.2\\[-0.45mm]\textcolor{black!62}{75.8}} & \makecell{64.5\\[-0.45mm]\textcolor{black!62}{65.2}} & \makecell{66.6\\[-0.45mm]\textcolor{black!62}{66.7}} & \makecell{63.0\\[-0.45mm]\textcolor{black!62}{60.8}} & \makecell{67.7\\[-0.45mm]\textcolor{black!62}{67.7}} & \makecell{59.8\\[-0.45mm]\textcolor{black!62}{60.1}} & \makecell{54.4\\[-0.45mm]\textcolor{black!62}{51.2}} & \makecell{61.4\\[-0.45mm]\textcolor{black!62}{60.3}} & \makecell{61.2\\[-0.45mm]\textcolor{black!62}{60.0}} \\
\addlinespace[1pt]
GPT-5.5 & Rating & \makecell{59.3\\[-0.45mm]\textcolor{black!62}{35.5}} & \makecell{60.2\\[-0.45mm]\textcolor{black!62}{36.0}} & \makecell{61.6\\[-0.45mm]\textcolor{black!62}{36.2}} & \makecell{69.7\\[-0.45mm]\textcolor{black!62}{45.1}} & \makecell{66.6\\[-0.45mm]\textcolor{black!62}{35.2}} & \makecell{64.3\\[-0.45mm]\textcolor{black!62}{38.0}} & \makecell{60.8\\[-0.45mm]\textcolor{black!62}{36.8}} & \makecell{64.5\\[-0.45mm]\textcolor{black!62}{36.6}} & \makecell{61.4\\[-0.45mm]\textcolor{black!62}{36.3}} & \makecell{53.8\\[-0.45mm]\textcolor{black!62}{34.5}} & \makecell{61.2\\[-0.45mm]\textcolor{black!62}{35.7}} & \makecell{60.3\\[-0.45mm]\textcolor{black!62}{36.0}} \\
\bottomrule
\end{tabular}
}
\end{table}

Changing the prompt format can reverse a model's class preference even when
the visual input and decision task remain fixed. Qwen3.5-0.8B increases from
19.7\% FakeR under Binary to 97.8\% under Diagnostic and 100.0\% under Rating.
Qwen3.5-4B moves from 85.2\% to 5.5\% and 21.9\%. Gemini's FakeR varies by only
2.7 points and remains at least 51.8\% across the three prompts. GPT-5.5 also
changes little across prompts, but its FakeR remains 2.9--4.4\% while RealR
exceeds 99.2\%. Prompt stability can therefore reflect a persistent class
preference rather than reliable detection
(Figure~\ref{fig:app_zeroshot_prompt_bias}).

Requiring structured outputs does not consistently improve detection. Under
Rating, Gemini-3.1-Pro-Preview obtains 63.6/63.0 paired AUC/verdict macro-F1,
and Qwen3.5-122B-A10B obtains 55.8/55.6. GPT-5.5 instead reaches 62.1/36.9.
Its continuous score ranks generated videos above their matched real anchors
better than chance, but its explicit verdict predicts almost every input as
\emph{Real}. The Diagnostic aspect scores can also become uninformative.
Qwen3.5-0.8B and Qwen3.5-9B repeat one value across all five aspects in 99.7\%
and 99.8\% of complete outputs, respectively. Gemini repeats one value across
all five aspects in 55.2\%. Even for Qwen3.5-122B-A10B, the five aspect AUCs
remain 54.5--54.8\%, while its Diagnostic verdict reaches 54.8\% BAcc.
Structured prompting therefore provides neither consistently stronger class
separation nor reliable aspect-level evidence. Full class-conditional, rating,
and diagnostic analyses are provided in Appendix~\ref{app:zeroshot-full}.

\begin{figure}[H]
    \centering
    \includegraphics[width=\textwidth]{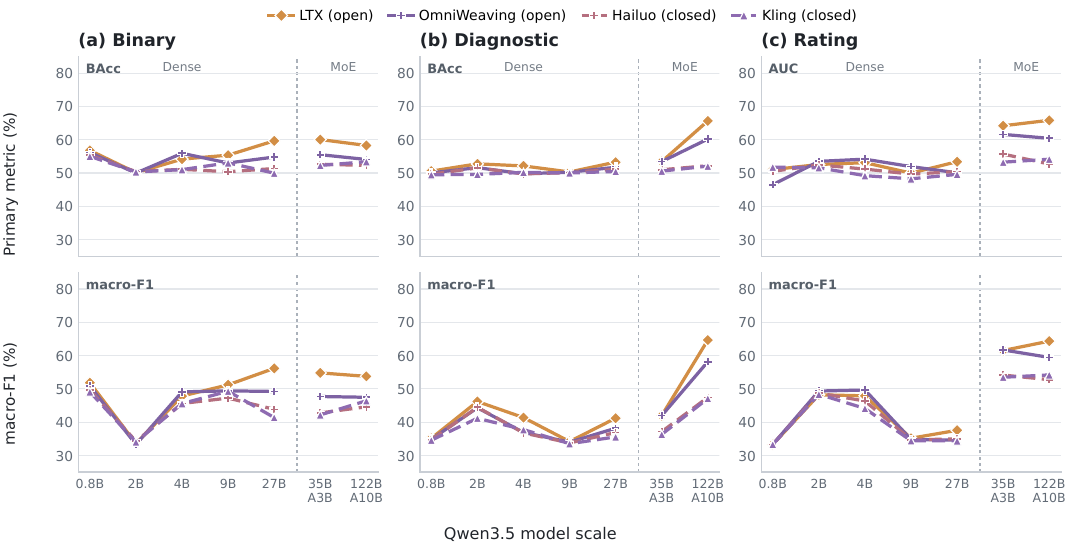}
    \caption{\textbf{Qwen3.5 scaling across prompts and generation sources.}
    The four columns correspond to representative RA-Bench generators. The top
    row reports BAcc for Binary and Diagnostic and paired AUC for Rating; the
    bottom row reports macro-F1 from each prompt's explicit Overall Verdict.
    Horizontal dashed lines in the top row mark the 50\% reference level. The
    vertical dotted line separates dense and MoE variants, whose trajectories
    are drawn separately. Solid and dashed lines denote open-source and
    closed-source generators, respectively.}
    \label{fig:zeroshot_source_trajectories}
\end{figure}

Increasing Qwen3.5 scale does not produce consistent gains across prompts and
generators. From 0.8B to 122B-A10B, Rating AUC rises from 51.0 to 65.8 on LTX
but only from 50.5 to 52.7 on Hailuo. Over the same endpoints, macro-F1 changes
from 49.7 to 44.6 for Binary on Hailuo, 35.2 to 64.7 for Diagnostic on LTX, and
33.3 to 64.3 for Rating on LTX. At 122B-A10B, the Binary BAcc, Diagnostic BAcc,
and Rating AUC values for Hailuo and Kling remain within 4.2 points of the 50\%
reference, whereas LTX reaches 65.6\% BAcc under Diagnostic and 65.8\% AUC
under Rating. Scaling therefore improves selected source--prompt pairs but does
not produce a reliably accurate detector across RA-Bench.

\paragraph{Takeaway.}
Scaling zero-shot multimodal models does not remove their sensitivity to prompt format or generation source.

\subsubsection{Fine-Tuned MLLMs Exhibit Protocol Dependence and Class Bias}
\label{sec:finetuned-shortcut}

\providecommand{\genlogo}[1]{\raisebox{-0.25em}{\includegraphics[height=1.15em]{figures/logos/#1}}}
\providecommand{\genhead}[3]{\makecell{\genlogo{#1}\\[-0.2mm]\textbf{#2}\\\textbf{#3}}}

\begin{table}[H]
\caption{\textbf{Fine-tuned MLLMs across the nine RA-Bench generation sources.} Skyra prompts receive the same 16 frames; \BusterXpp~\citep{wen2025busterxpp} uses its released pipeline. Each source is evaluated on matched real--generated pairs, with BAcc, FakeR, and macro-F1 reported in \%. The RealR values beside each model name are measured on the full set of 1{,}830 real anchors and are listed in official-timestamp/frame-index order for Skyra. Open and Closed are source-equal averages over the four open-source and five closed-source RA-Bench generators, respectively; Wan2.2 fixed$^{*}$ is excluded from Open.}
\label{tab:finetuned_main}
\centering
\scriptsize
\setlength{\tabcolsep}{2.25pt}
\renewcommand{\arraystretch}{0.94}
\resizebox{\textwidth}{!}{
\begin{tabular}{ll *{5}{c} c *{5}{c} c}
\toprule
 & & \multicolumn{5}{c}{\textbf{Open-source settings}} & \textbf{Open}
 & \multicolumn{5}{c}{\textbf{Closed-source generators}} & \textbf{Closed} \\
\cmidrule(lr){3-7}\cmidrule(lr){9-13}
\textbf{Prompt} & \textbf{Metric}
 & \genhead{wan.png}{Wan2.2}{dyn.} & \genhead{wan.png}{Wan2.2}{fixed$^{*}$}
 & \genhead{wan_lightning.png}{Wan2.2}{Light.} & \genhead{ltx.png}{LTX}{}
 & \genhead{omni.png}{Omni}{Weav.} & \textbf{avg.}
 & \genhead{happyhorse.png}{Happy}{Horse} & \genhead{runway.png}{Runway}{}
 & \genhead{kling.png}{Kling}{} & \genhead{seedance.png}{Seed}{ance2.0}
 & \genhead{hailuo.png}{Hailuo}{} & \textbf{avg.} \\
\midrule
\rowcolor{blue!6}
\multicolumn{14}{l}{\textbf{Skyra-SFT} \; (RealR: 87.4 / 55.4)} \\
\multirow{3}{*}{\makecell[l]{Official\\timestamp}} & \textbf{BAcc} & 60.6 & 91.9 & 59.9 & 74.7 & 51.9 & 61.8 & 80.2 & 80.3 & 80.1 & 73.5 & 54.9 & 73.8 \\
 & \textbf{FakeR} & 33.8 & 96.4 & 32.5 & 61.9 & 16.4 & 36.1 & 73.3 & 73.2 & 72.8 & 59.6 & 22.5 & 60.3 \\
 & \textbf{macro-F1} & 57.5 & 91.9 & 56.7 & 74.2 & 45.0 & 58.4 & 80.1 & 80.2 & 80.0 & 73.0 & 49.6 & 72.6 \\
\multirow{3}{*}{\makecell[l]{Frame\\index}} & \textbf{BAcc} & 60.2 & 59.1 & 58.7 & 31.1 & 50.9 & 50.2 & 59.4 & 61.0 & 59.1 & 52.8 & 56.0 & 57.7 \\
 & \textbf{FakeR} & 65.1 & 62.8 & 62.0 & 6.7 & 46.3 & 45.0 & 63.9 & 66.6 & 62.5 & 52.4 & 56.6 & 60.4 \\
 & \textbf{macro-F1} & 60.1 & 59.0 & 58.7 & 26.7 & 50.8 & 49.1 & 59.3 & 60.9 & 59.1 & 52.8 & 56.0 & 57.6 \\
\midrule
\rowcolor{cyan!6}
\multicolumn{14}{l}{\textbf{Skyra-RL} \; (RealR: 84.0 / 49.5)} \\
\multirow{3}{*}{\makecell[l]{Official\\timestamp}} & \textbf{BAcc} & 61.1 & 90.8 & 60.7 & 76.9 & 52.1 & 62.7 & 81.2 & 81.1 & 81.4 & 74.8 & 56.1 & 74.9 \\
 & \textbf{FakeR} & 38.2 & 97.7 & 37.3 & 69.8 & 20.2 & 41.4 & 78.7 & 78.3 & 78.9 & 65.9 & 28.3 & 66.0 \\
 & \textbf{macro-F1} & 58.9 & 90.8 & 58.4 & 76.8 & 46.7 & 60.2 & 81.2 & 81.1 & 81.4 & 74.6 & 52.4 & 74.1 \\
\multirow{3}{*}{\makecell[l]{Frame\\index}} & \textbf{BAcc} & 60.0 & 59.0 & 58.6 & 29.7 & 53.0 & 50.3 & 59.7 & 61.3 & 60.6 & 54.5 & 57.1 & 58.6 \\
 & \textbf{FakeR} & 70.5 & 68.4 & 67.7 & 9.9 & 56.4 & 51.1 & 70.5 & 73.2 & 71.5 & 61.5 & 64.6 & 68.3 \\
 & \textbf{macro-F1} & 59.6 & 58.6 & 58.3 & 26.8 & 52.9 & 49.4 & 59.2 & 60.7 & 60.1 & 54.3 & 56.9 & 58.2 \\
\midrule
\rowcolor{orange!7}
\multicolumn{14}{l}{\textbf{\BusterXpp} \; (RealR: 93.7)} \\
\multirow{3}{*}{\makecell[l]{Released\\pipeline}} & \textbf{BAcc} & 49.8 & 50.2 & 51.4 & 48.9 & 49.9 & 50.0 & 50.1 & 50.2 & 49.7 & 49.7 & 49.9 & 49.9 \\
 & \textbf{FakeR} & 6.0 & 6.7 & 9.1 & 4.1 & 6.1 & 6.3 & 6.5 & 6.8 & 5.6 & 6.4 & 6.1 & 6.3 \\
 & \textbf{macro-F1} & 37.9 & 38.6 & 40.8 & 36.1 & 38.0 & 38.2 & 38.4 & 38.6 & 37.6 & 38.1 & 38.0 & 38.1 \\
\bottomrule
\end{tabular}
}
\end{table}

Under the official-timestamp prompt, Skyra-SFT and Skyra-RL reach source-equal mean BAcc values of 68.5\% and 69.5\%, respectively, on RA-Bench (Table~\ref{tab:finetuned_main}). These means do not indicate consistent cross-source detection: FakeR ranges from 16.4\% to 73.3\% for Skyra-SFT and from 20.2\% to 78.9\% for Skyra-RL. \BusterXpp\ exhibits a different failure mode. Its FakeR remains between 4.1\% and 9.1\%, while its macro-F1 remains between 36.1\% and 40.8\%. Together with its 93.0--93.8\% real recall, this pattern indicates a strong preference for the \emph{Real} class rather than reliable recognition of generated videos. The evaluated fine-tuned systems therefore exhibit distinct weaknesses: source-sensitive detection for Skyra and a strong class bias for \BusterXpp.

\begin{figure*}[!t]
    \centering
    \includegraphics[width=0.82\textwidth]{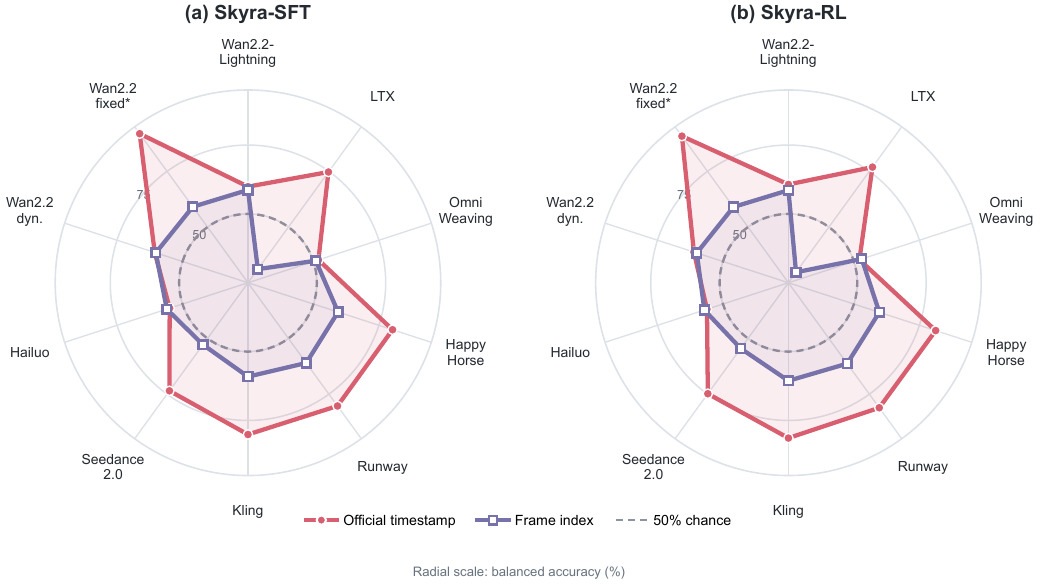}
    \caption{\textbf{Skyra performance changes with temporal-tag representation.} The official-timestamp and frame-index prompts receive the same 16 frames. Each radar reports per-source BAcc across the nine RA-Bench generators and Wan2.2 fixed$^{*}$, an auxiliary control excluded from benchmark-level averages; the dashed ring marks 50\%.}
    \label{fig:skyra_shortcut}
\end{figure*}

The fixed-duration control reveals an additional anomaly in Skyra's official-timestamp results. Wan2.2 dynamic and its fixed-duration control use the same generator, anchors, prompts, and first-frame conditioning, with duration policy as the controlled difference. Fixing the duration at approximately 5 seconds increases FakeR from 33.8\% to 96.4\% for Skyra-SFT and from 38.2\% to 97.7\% for Skyra-RL. A related pattern appears within RA-Bench. Across the eight RA-Bench generation sources that include clips ending at exactly 5.00 seconds and clips with other final timestamps, the former receive substantially higher FakeR; the source-equal gaps are 48.6 and 43.4 points, respectively. These comparisons reveal a strong association between the displayed temporal grid and Skyra's predictions, but they do not determine whether the difference arises from duration-dependent visual content or from the temporal labels themselves.

To distinguish these explanations, we replace absolute timestamps with frame indices while preserving the same 16 visual frames and their order. This intervention reduces the exact-5-second FakeR gap from 48.6 to 1.5 points for Skyra-SFT and from 43.4 to 1.4 points for Skyra-RL. Mean BAcc also falls from 68.5\% to 54.4\% and from 69.5\% to 54.9\%, respectively. Because the visual input is unchanged, the collapse of the temporal-grid gap identifies a strong sensitivity to temporal-label representation. An audit of the released ViF metadata~\citep{li2025skyra} further shows that an exact-5-second final timestamp is correlated with the class label, providing a plausible source of this protocol prior (Appendix~\ref{app:finetuned-full}).

Removing timestamps does not recover stable content-based detection. Under frame indices, per-source BAcc still spans 31.1--61.0\% for Skyra-SFT and 29.7--61.3\% for Skyra-RL (Figure~\ref{fig:skyra_shortcut}). Temporal labels therefore explain an important component of Skyra's behavior, but not all variation across generation sources. RL fine-tuning removes neither the protocol dependence nor the remaining source-specific failures.

\FloatBarrier

\paragraph{Takeaway.}
For the fine-tuned MLLMs, replacing timestamps with frame indices lowers Skyra to 54.4--54.9\% mean BAcc, while \BusterXpp\ reaches only 4.1--9.1\% FakeR.

\subsubsection{Summary of Detector Generalization}
\label{sec:detector-side-summary}

Across all three detector families, performance fails to transfer consistently across RA-Bench sources. Traditional detector AUC falls from 67.6--98.6\% in public references to source-level means of 43.9--57.3\% on RA-Bench, with different detectors leading on different generators. Scaling parameters of zero-shot multimodal models does not remove their sensitivity to prompt format or generation source. For the fine-tuned MLLMs, replacing timestamps with frame indices lowers Skyra to 54.4--54.9\% mean BAcc, while \BusterXpp\ reaches only 4.1--9.1\% FakeR. Public benchmark performance, larger model scale, and task-specific fine-tuning therefore do not by themselves ensure generalization across generation sources.

\subsection{What Makes Generated Videos Hard to Detect?}
\label{sec:generation-side}

\subsubsection{Generation Quality Does Not Uniformly Determine Detection Difficulty}
\label{sec:generation-quality}

We evaluate all 16{,}056 generated clips in RA-Bench using the released VBench++ I2V evaluators~\citep{huang2025vbenchpp}. We report three aggregate scores: the official VBench-I2V Quality Score, which combines six dimensions of temporal consistency, motion, and frame-wise quality; Condition Fidelity, defined as the mean of the normalized Video--Image Subject and Background Consistency scores; and Combined Quality, their equal-weight mean. We omit Camera Motion because RA-Bench prompts do not specify the controlled camera-motion labels required by that evaluator. To separate clip-level quality variation from differences among generation sources, we estimate associations within each source. We further stratify clips by Dynamic Degree because low-motion clips can receive high temporal-consistency scores simply by changing little. For each continuous quality score, we report the adjusted change in fake-side detector output associated with an interquartile increase, giving each source equal weight. These estimates describe associations within RA-Bench rather than causal effects. Figure~\ref{fig:generation-quality} presents the results, with full scoring and statistical details provided in Appendix~\ref{app:generation-quality}.

\begin{figure*}[t]
    \centering
    \includegraphics[width=\textwidth]{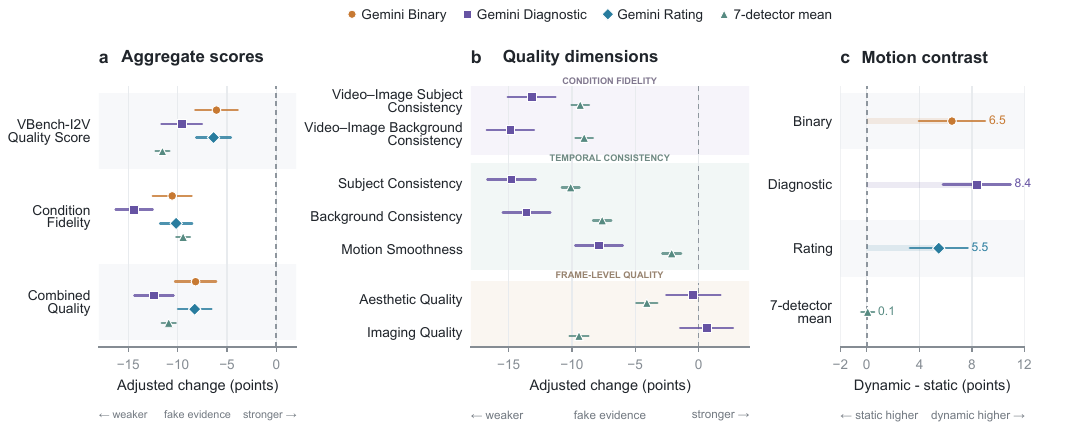}
    \caption{\textbf{Generation quality and fake-side detectability on RA-Bench.}
    (a) Associations of the VBench-I2V Quality Score, Condition Fidelity, and Combined Quality with three Gemini-3.1-Pro-Preview outputs and the mean fake-score percentile of seven traditional detectors.
    (b) Dimension-level associations for Gemini Diagnostic and the traditional-detector mean; complete prompt-wise results are reported in Appendix~\ref{app:generation-quality}.
    (c) Dynamic-minus-static contrasts after adjustment for generation source and matched real-video anchor.
    Panels (a) and (b) report adjusted changes associated with an interquartile increase within each source and Dynamic Degree group. Negative values indicate weaker fake evidence at higher quality, while positive values in panel (c) indicate stronger fake evidence for dynamic clips. Traditional scores are percentile-normalized within detector and source before averaging. Error bars denote 95\% confidence intervals based on standard errors clustered by real-video anchor.}
    \label{fig:generation-quality}
\end{figure*}

Across the three Gemini outputs and the traditional-detector mean in Figure~\ref{fig:generation-quality}(a), higher values of all three aggregate scores are associated with weaker fake evidence. For Gemini Diagnostic, an interquartile increase in the VBench-I2V Quality Score, Condition Fidelity, and Combined Quality is associated with decreases of 9.6, 14.4, and 12.4 percentage points in fake recall, respectively. Condition Fidelity also has the largest negative association for Gemini Binary and Rating. The traditional-detector mean decreases by 9.5--11.5 percentile points across the three scores, with its largest change associated with the VBench-I2V Quality Score. No single aggregate score therefore characterizes detection difficulty consistently across detector families.

The dimension-level results explain this difference. For Gemini Diagnostic, higher Video--Image Subject and Background Consistency and higher within-video Subject and Background Consistency are each associated with 13.2--14.9 percentage-point decreases in fake recall. Motion Smoothness shows a smaller decrease of 7.9 percentage points, while Aesthetic Quality and Imaging Quality show no clear association with the Diagnostic verdict. The traditional-detector mean follows a different pattern: Subject Consistency and Imaging Quality show the largest decreases, at 10.1 and 9.4 percentile points, respectively, while Motion Smoothness changes it by only $-2.1$ points. Gemini Diagnostic is thus most closely associated with Condition Fidelity and temporal consistency, whereas the traditional-detector mean also varies strongly with frame-level Imaging Quality.

Dynamic Degree further shows why generation quality cannot be treated as a single axis. After adjustment for generation source and matched real-video anchor, dynamic clips receive 5.5--8.4 points more fake evidence than static clips across the three Gemini outputs. The traditional-detector mean changes by only 0.1 percentile points, and its 95\% confidence interval spans zero. Because Dynamic Degree contributes positively to the VBench-I2V Quality Score, this association partly offsets the negative associations of its consistency dimensions and helps explain why the Quality Score has a weaker negative association with Gemini than Condition Fidelity. Taken together, Condition Fidelity and temporal consistency are associated with weaker fake evidence, whereas dynamic clips are associated with stronger Gemini fake evidence and no clear change in the traditional-detector mean. Detection difficulty therefore depends on both the quality dimension and the detector family being evaluated.

\paragraph{Takeaway.}
Within sources, stronger Condition Fidelity and temporal consistency are associated with weaker evidence for the generated class, whereas dynamic content strengthens Gemini's fake evidence but leaves the traditional-detector mean nearly unchanged.

\subsubsection{Real-Image Conditioning Affects Detector Families Differently}
\label{sec:conditioning}

\begin{figure*}[t]
    \centering
    \includegraphics[width=\textwidth]{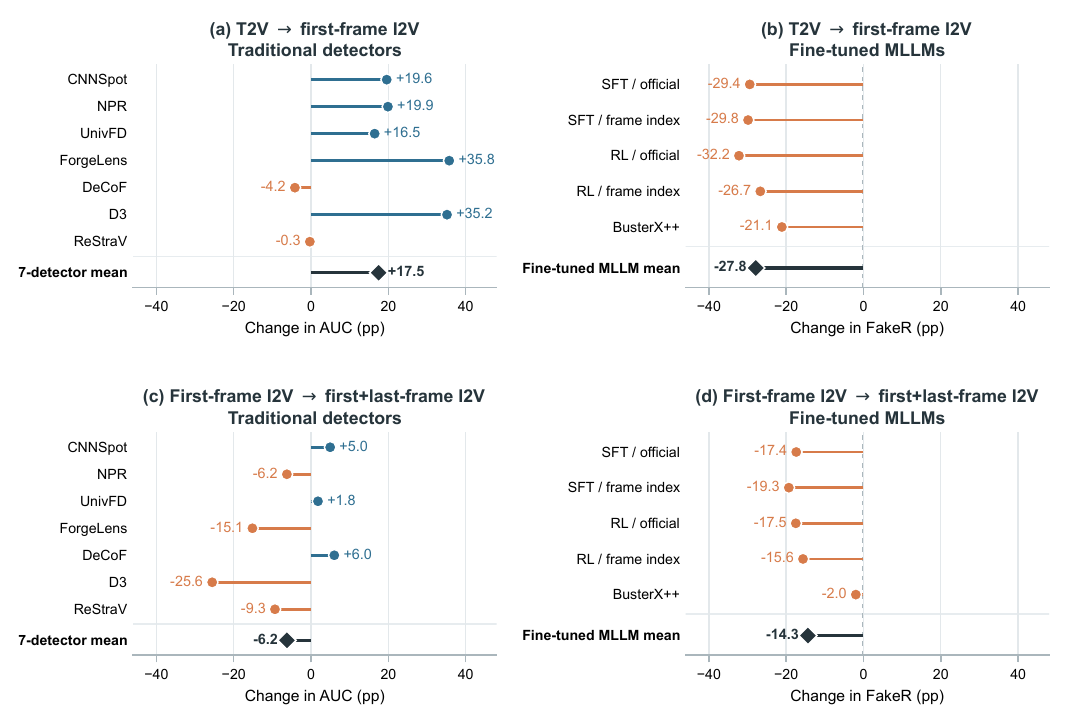}
    \caption{\textbf{Changes in detectability across Wan2.2 generation settings.}
    Each panel reports the second setting minus the first setting named in its title. Panels (a) and (c) show paired AUC for the seven traditional detectors, while panels (b) and (d) show FakeR for the five fine-tuned MLLM settings. Positive values indicate higher AUC or FakeR in the second setting; negative values indicate lower values. Diamonds denote detector-family means.}
    \label{fig:conditioning-effects}
\end{figure*}

RA-Bench conditions each generated video on the first frame of its matched real-video anchor. To examine detector behavior under different amounts of real-image conditioning, we use Wan2.2 to generate the same 1{,}830 anchor-derived prompts under three settings: T2V, first-frame I2V, and first+last-frame I2V. T2V receives only the prompt, whereas the two I2V settings additionally receive the matched first frame or the matched first and last frames. We report seed-0 results for the seven traditional detectors and five settings of MLLMs fine-tuned for AI-generated video detection: official-timestamp and frame-index prompts for Skyra-SFT and Skyra-RL, and the released \BusterXpp\ pipeline. Table~\ref{tab:conditioning_main} reports the absolute results, while Figure~\ref{fig:conditioning-effects} shows the changes between adjacent generation settings. Complete classification metrics and cross-seed results are provided in Appendix~\ref{app:conditioning}.

\begin{table}[t]
\caption{\textbf{Detection under three Wan2.2 generation settings.}
We compare T2V, first-frame I2V (the RA-Bench setting), and
first+last-frame I2V on the same 1{,}830 anchor-derived prompts at seed 0.
Traditional detectors report AUC with T@5\% (TPR at 5\% FPR) in gray below;
Skyra-SFT, Skyra-RL, and \BusterXpp\ report FakeR. All values are percentages.
\BusterXpp\ abstentions are counted as incorrect.}
\label{tab:conditioning_main}
\centering
\scriptsize
\setlength{\tabcolsep}{4.0pt}
\renewcommand{\arraystretch}{1.02}
\begin{tabular}{@{}llccc@{}}
\toprule
\textbf{Detector / configuration} & \textbf{Metric}
& \textbf{T2V}
& \makecell{\textbf{First frame}\\[-0.2ex]\textbf{(I2V; RA-Bench)}}
& \makecell{\textbf{First+last frames}\\[-0.2ex]\textbf{(I2V)}} \\
\midrule
\rowcolor{gray!8}
\multicolumn{5}{l}{\textit{Traditional detectors}} \\
CNNSpot   & \makecell{AUC\\[-0.35mm]\textcolor{black!62}{T@5\%}}
  & \makecell{15.9\\[-0.35mm]\textcolor{black!62}{0.2}} & \makecell{35.5\\[-0.35mm]\textcolor{black!62}{2.4}} & \makecell{40.5\\[-0.35mm]\textcolor{black!62}{3.1}} \\
NPR       & \makecell{AUC\\[-0.35mm]\textcolor{black!62}{T@5\%}}
  & \makecell{30.7\\[-0.35mm]\textcolor{black!62}{0.9}} & \makecell{50.6\\[-0.35mm]\textcolor{black!62}{3.8}} & \makecell{44.4\\[-0.35mm]\textcolor{black!62}{3.0}} \\
UnivFD    & \makecell{AUC\\[-0.35mm]\textcolor{black!62}{T@5\%}}
  & \makecell{16.3\\[-0.35mm]\textcolor{black!62}{0.1}} & \makecell{32.7\\[-0.35mm]\textcolor{black!62}{0.4}} & \makecell{34.5\\[-0.35mm]\textcolor{black!62}{0.6}} \\
ForgeLens & \makecell{AUC\\[-0.35mm]\textcolor{black!62}{T@5\%}}
  & \makecell{24.2\\[-0.35mm]\textcolor{black!62}{1.0}} & \makecell{60.0\\[-0.35mm]\textcolor{black!62}{19.0}} & \makecell{44.8\\[-0.35mm]\textcolor{black!62}{6.8}} \\
DeCoF     & \makecell{AUC\\[-0.35mm]\textcolor{black!62}{T@5\%}}
  & \makecell{66.6\\[-0.35mm]\textcolor{black!62}{1.5}} & \makecell{62.5\\[-0.35mm]\textcolor{black!62}{3.9}} & \makecell{68.5\\[-0.35mm]\textcolor{black!62}{6.8}} \\
D3        & \makecell{AUC\\[-0.35mm]\textcolor{black!62}{T@5\%}}
  & \makecell{20.2\\[-0.35mm]\textcolor{black!62}{2.0}} & \makecell{55.4\\[-0.35mm]\textcolor{black!62}{6.4}} & \makecell{29.8\\[-0.35mm]\textcolor{black!62}{3.5}} \\
ReStraV   & \makecell{AUC\\[-0.35mm]\textcolor{black!62}{T@5\%}}
  & \makecell{59.6\\[-0.35mm]\textcolor{black!62}{9.8}} & \makecell{59.3\\[-0.35mm]\textcolor{black!62}{8.5}} & \makecell{50.0\\[-0.35mm]\textcolor{black!62}{4.0}} \\
\textbf{7-detector mean} & \makecell{AUC\\[-0.35mm]\textcolor{black!62}{T@5\%}}
  & \makecell{\textbf{33.4}\\[-0.35mm]\textcolor{black!62}{\textbf{2.2}}} & \makecell{\textbf{50.9}\\[-0.35mm]\textcolor{black!62}{\textbf{6.3}}} & \makecell{\textbf{44.6}\\[-0.35mm]\textcolor{black!62}{\textbf{4.0}}} \\
\midrule
\rowcolor{blue!6}
\multicolumn{5}{l}{\textbf{Skyra-SFT}} \\
Official timestamp & FakeR & 63.2 & 33.8 & 16.4 \\
Frame index        & FakeR & 94.9 & 65.1 & 45.8 \\
\midrule
\rowcolor{cyan!6}
\multicolumn{5}{l}{\textbf{Skyra-RL}} \\
Official timestamp & FakeR & 70.4 & 38.2 & 20.7 \\
Frame index        & FakeR & 97.2 & 70.5 & 54.9 \\
\midrule
\rowcolor{orange!7}
\multicolumn{5}{l}{\textbf{\BusterXpp}} \\
Released pipeline & FakeR & 27.0 & 6.0 & 4.0 \\
\addlinespace[1pt]
\textbf{Mean across fine-tuned MLLM settings} & \textbf{FakeR} & \textbf{70.5} & \textbf{42.7} & \textbf{28.3} \\
\bottomrule
\end{tabular}
\end{table}

From T2V to first-frame I2V, the seven-detector mean AUC increases from 33.4\% to 50.9\%, with higher AUC for five of the seven traditional detectors. Over the same comparison, the mean FakeR across the five fine-tuned MLLM settings decreases from 70.5\% to 42.7\%, and every setting shows a decrease of 21.0--32.2 points. The two detector families therefore respond in opposite directions to first-frame conditioning.

From first-frame to first+last-frame I2V, FakeR decreases by 15.6--19.3 points across the four Skyra settings, and the mean across all five fine-tuned MLLM settings falls from 42.7\% to 28.3\%. \BusterXpp\ FakeR decreases from 6.0\% to 4.0\%; its smaller change reflects the already low FakeR rather than stable detection. Both Skyra prompt formats follow the same direction, showing that the decline is not specific to the temporal-label format.

Traditional detectors do not show the same monotonic pattern. Moving from first-frame to first+last-frame I2V increases AUC for CNNSpot, UnivFD, and DeCoF, but decreases it for NPR, ForgeLens, D3, and ReStraV. Their mean AUC falls from 50.9\% to 44.6\%, and mean T@5\% falls from 6.3\% to 4.0\%, while detector-specific AUC changes range from $-25.6$ to $+6.0$ points. For Skyra under the official-timestamp prompts and for \BusterXpp, the FakeR decrease from first-frame to first+last-frame I2V persists across all three seeds (Appendix~\ref{app:conditioning}). Generation settings therefore do not impose a shared difficulty ordering across detector families: more real-image conditioning is associated with progressively lower FakeR for the fine-tuned MLLMs, whereas traditional-detector AUC changes in detector-specific directions.

\paragraph{Takeaway.}
Across T2V, first-frame I2V, and first+last-frame I2V, the mean FakeR of the evaluated MLLMs fine-tuned for AI-generated video detection falls from 70.5\% to 42.7\% and 28.3\%, while traditional-detector AUCs move in both directions.

\FloatBarrier

\subsubsection{Detection Results Vary Little Across Sampling Seeds}
\label{sec:seed}

Sampling can change a video's appearance and motion even when its prompt and conditioning image are fixed. All primary results for the four open-source RA-Bench generators use seed 0. To determine whether the detector--source patterns depend on this choice, we generate two additional realizations, with seeds 42 and 123, from the same 1{,}830 anchors per generator. We hold the prompts, conditioning images, and all other generation settings fixed. Closed-source providers are omitted because their APIs do not expose a controllable seed, and the fixed-duration Wan2.2 control is omitted because this analysis concerns the four open-source RA-Bench generators. Table~\ref{tab:seed_stability_main} summarizes the comparison.

\begin{minipage}{\textwidth}
\captionof{table}{\textbf{Detection results remain stable across generation seeds.}
Results for seeds 0, 42, and 123 on the same 1{,}830 anchors per generator:
(a) source-level mean AUC over seven traditional detectors; (b) source-wise
FakeR and BAcc ranges for the fine-tuned MLLMs. Skyra-SFT and Skyra-RL use the
released official-timestamp prompt, and \BusterXpp\ uses its released evaluation
pipeline.}
\label{tab:seed_stability_main}

\noindent\begin{minipage}[t]{0.47\textwidth}
\vspace{0pt}
\centering
\footnotesize
\textbf{(a) Traditional-detector AUC}\par\vspace{3pt}
\fontsize{7.7}{8.6}\selectfont
\renewcommand{\arraystretch}{0.96}
\setlength{\tabcolsep}{1.0pt}
\begin{tabular*}{\linewidth}{@{\extracolsep{\fill}}c*{4}{c}@{}}
\toprule
\textbf{Seed} &
\makecell{\textbf{Wan2.2}\\\textbf{dynamic}} &
\makecell{\textbf{Wan2.2}\\\textbf{Lightning}} &
\textbf{LTX} & \textbf{OmniWeaving} \\
\midrule
0 & 50.85 & 54.24 & 57.26 & 56.12 \\
42 & 50.39 & 54.49 & 57.77 & 56.59 \\
123 & 50.72 & 54.47 & 56.72 & 56.12 \\
\(\Delta\) & 0.46 & 0.25 & \textbf{1.05} & 0.47 \\
\bottomrule
\end{tabular*}

\par\vspace{3pt}
\raggedright
\footnotesize
\(\Delta\) is the maximum minus the minimum across seeds; all AUC values are in \%.
Bold marks the largest \(\Delta\) in each panel. Full results are in
Appendix~\ref{app:seed-detail}.
\end{minipage}\hfill
\begin{minipage}[t]{0.50\textwidth}
\vspace{0pt}
\small
Traditional detectors vary little at the source level. The largest range among
the source-level seven-detector AUC means is 1.05 points, for LTX. Seed 0 differs from the
three-seed averages by at most 0.20 AUC points. Although individual
detector--source cells vary more, the 28-cell AUC pattern remains highly
consistent (pairwise Spearman 0.978--0.989). Only three cells cross 50\% AUC,
and all remain close to random ranking under every seed
(Appendix~\ref{app:seed-detail}). Sampling therefore changes some local
estimates without altering the broader detector--source pattern.
\end{minipage}
\end{minipage}

\Needspace{0.34\textheight}
\par\vspace{6pt}

\begin{minipage}{\textwidth}
\centering
\footnotesize
\textbf{(b) Fine-tuned MLLM FakeR}\par\vspace{4pt}
\fontsize{7.7}{8.6}\selectfont
\renewcommand{\arraystretch}{1.00}
\setlength{\tabcolsep}{1.0pt}
\begin{tabular*}{\textwidth}{@{\extracolsep{\fill}}l*{15}{c}@{}}
\toprule
& \multicolumn{5}{c}{\textbf{Skyra-SFT}}
& \multicolumn{5}{c}{\textbf{Skyra-RL}}
& \multicolumn{5}{c}{\textbf{\BusterXpp}} \\
\cmidrule(lr){2-6}
\cmidrule(lr){7-11}
\cmidrule(lr){12-16}
\textbf{Source}
& \textbf{0} & \textbf{42} & \textbf{123}
& \(\boldsymbol{\Delta}\textbf{F}\)
& \(\boldsymbol{\Delta}\textbf{B}\)
& \textbf{0} & \textbf{42} & \textbf{123}
& \(\boldsymbol{\Delta}\textbf{F}\)
& \(\boldsymbol{\Delta}\textbf{B}\)
& \textbf{0} & \textbf{42} & \textbf{123}
& \(\boldsymbol{\Delta}\textbf{F}\)
& \(\boldsymbol{\Delta}\textbf{B}\) \\
\midrule
Wan2.2 dynamic
& 33.77 & 32.84 & 31.42 & 2.35 & 1.17
& 38.20 & 39.45 & 37.65 & 1.80 & 0.90
& 5.96 & 6.61 & 6.34 & 0.66 & 0.33 \\
Wan2.2-Lightning
& 32.46 & 30.78 & 30.11 & 2.35 & 1.17
& 37.32 & 36.50 & 34.86 & \textbf{2.46} & \textbf{1.23}
& 9.07 & 9.18 & 8.58 & 0.60 & 0.30 \\
LTX
& 61.91 & 62.90 & 63.01 & 1.09 & 0.55
& 69.84 & 70.44 & 70.60 & 0.77 & 0.38
& 4.10 & 3.88 & 3.61 & 0.49 & 0.25 \\
OmniWeaving
& 16.45 & 17.32 & 16.07 & 1.26 & 0.63
& 20.22 & 20.66 & 20.49 & 0.44 & 0.22
& 6.12 & 6.83 & 6.39 & 0.71 & 0.36 \\
\bottomrule
\end{tabular*}

\par\vspace{3pt}
\raggedright
\footnotesize
\(\Delta\) is the maximum minus the minimum across seeds; F and B denote FakeR
and BAcc, respectively; all metrics are in \%. The largest \(\Delta\) within each
panel is bold. Detector-specific results and protocol details are in
Appendix~\ref{app:seed-detail}.
\end{minipage}

\smallskip

The fine-tuned MLLMs show similarly limited variation. The largest FakeR/BAcc
ranges are 2.35/1.17 points for Skyra-SFT, 2.46/1.23 for Skyra-RL, and 0.71/0.36
for \BusterXpp. Because the real-video control is shared, BAcc changes are driven
entirely by FakeR. Skyra's source differences persist under every seed, while
\BusterXpp\ remains below 10\% FakeR for every source--seed combination. Neither
pattern is therefore specific to the seed-0 realizations used in the primary
benchmark.

\paragraph{Takeaway.}
Across the three tested seeds, the source-level seven-detector mean AUC and fine-tuned MLLM FakeR vary only slightly.

\subsubsection{Summary of Generation Properties}
\label{sec:generation-side-summary}

Generation quality and conditioning affect detector families differently. Within sources, stronger Condition Fidelity and temporal consistency are associated with weaker evidence for the generated class, whereas dynamic content strengthens Gemini's fake evidence but leaves the traditional-detector mean nearly unchanged. Across T2V, first-frame I2V, and first+last-frame I2V, the mean FakeR of the evaluated MLLMs fine-tuned for AI-generated video detection falls from 70.5\% to 42.7\% and 28.3\%, while traditional-detector AUCs move in both directions. Across the three tested seeds, the source-level seven-detector mean AUC and fine-tuned MLLM FakeR vary only slightly.

\subsection{Human Perception and Social Dissemination}
\label{sec:human-reliability}

\subsubsection{Human Recognition Varies Sharply Across Generation Sources}
\label{sec:human}

Whether generated videos can mislead viewers is central to their societal risk, but detector performance alone does not measure human recognition. We therefore conduct a source-unaware human evaluation on RA-Bench. Twenty reviewers inspect videos in reviewer-specific randomized orders and choose among \emph{Real}, \emph{Uncertain}, and \emph{Generated}, with real and generated videos from different sources interleaved. Each video in the primary analysis receives three independent judgments. Reviewers label 60.3\% of generated-video judgments as \emph{Generated}. Real videos are recognized more often, with 71.9\% of judgments labeled \emph{Real}; however, 22.8\% are labeled \emph{Generated} and 5.3\% \emph{Uncertain}. Real videos depicting crisis events can therefore also be mistaken for generated content.

\noindent\begin{minipage}{\textwidth}
\captionof{table}{\textbf{Human recognition varies sharply by generation source.}
Stage~1 response shares are in \%. For generated sources, \emph{Generated} is
human FakeR. The \emph{Generated total} row pools judgments, whereas Open/Closed
averages weight generation sources equally.}
\label{tab:human_source_main}
\noindent\begin{minipage}[t]{0.50\textwidth}
\vspace{0pt}
\centering
\scriptsize
\renewcommand{\arraystretch}{0.96}
\setlength{\tabcolsep}{2.0pt}
\begin{tabular*}{\linewidth}{@{\extracolsep{\fill}}lccc@{}}
\toprule
\textbf{Video source} &
\textbf{Real} &
\textbf{Unc.} &
\textbf{Generated} \\
\midrule
Real videos      & \textbf{71.9} & 5.3 & 22.8 \\
Generated total  & 33.9 & 5.8 & \textbf{60.3} \\
\midrule
\rowcolor{blue!6}\multicolumn{4}{l}{\textit{Open-source generators}} \\
Wan2.2 dynamic   & 27.0 & 5.5 & 67.5 \\
Wan2.2 Lightning & 28.5 & 4.9 & 66.6 \\
LTX              & 30.3 & 5.1 & 64.6 \\
OmniWeaving      & 19.5 & 4.7 & 75.8 \\
\textbf{Open avg.} & \textbf{26.3} & \textbf{5.0} & \textbf{68.6} \\
\midrule
\rowcolor{orange!7}\multicolumn{4}{l}{\textit{Closed-source generators}} \\
HappyHorse       & 37.2 & 6.9 & 55.9 \\
Runway           & 34.8 & 5.6 & 59.7 \\
Kling            & 47.7 & 7.3 & 45.1 \\
Seedance2.0      & 51.9 & 7.4 & 40.7 \\
Hailuo           & 31.8 & 5.3 & 63.0 \\
\textbf{Closed avg.} & \textbf{40.6} & \textbf{6.5} & \textbf{52.9} \\
\bottomrule
\end{tabular*}
\end{minipage}\hfill
\begin{minipage}[t]{0.47\textwidth}
\vspace{0pt}
\small
The pooled result masks substantial differences across generation sources. The
source-equal FakeR is 68.6\% for the four open-source generators but 52.9\% for
the five closed-source generators. Seedance2.0 and Kling are the most difficult
to recognize as generated, with FakeR values of 40.7\% and 45.1\%, respectively,
whereas OmniWeaving reaches 75.8\%. This pattern is not driven by a small subset
of reviewers: all 20 reviewers obtain higher source-equal FakeR on open-source
than on closed-source generators, and leaving out any one reviewer preserves the
complete nine-source ranking. Detailed protocol, source-level counts, and
reviewer-level robustness checks are provided in Appendix~\ref{app:human-protocol}; generated
videos repeatedly judged \emph{Real} form the candidate pool studied next.
\end{minipage}
\end{minipage}

\paragraph{Takeaway.}
Reviewers identify 68.6\% of open-source videos as generated but only 52.9\% of closed-source videos, with the rates falling to 40.7\% for Seedance2.0 and 45.1\% for Kling.

\subsubsection{RA-Bench-HumanProof Remains Difficult Across Detector Families}
\label{sec:challenge-set}

We construct \textbf{RA-Bench-HumanProof} through two stages. Of the 16{,}038 generated videos in the standard review stream, Stage~1 retains 1{,}080 that all three assigned reviewers label \emph{Real}. Two additional reviewers independently reassess these candidates using the same source-unaware protocol, and a video is retained only when both again label it \emph{Real}. The resulting 633 generated videos have therefore been labeled \emph{Real} by all five reviewers. RA-Bench-HumanProof contains 119 open-source and 514 closed-source videos, including 160 from Kling and 159 from Seedance2.0.

For detector evaluation, each generated video in RA-Bench-HumanProof is paired with its matched real anchor. Because the source composition is determined by human selection rather than a predefined quota, we compare these results with a source-matched RA-Bench reference that uses the same source proportions. Table~\ref{tab:human_main} reports representative detectors from all three families. For discrete outputs, we report balanced accuracy (BAcc) and fake recall (FakeR); for continuous scores, we report paired AUC and T@5\%. The final column reports the same metric pair for the source-matched RA-Bench reference: BAcc/FakeR for discrete outputs and AUC/T@5\% for continuous scores. Full construction details, response pairs, source counts, coverage audits, and comparison rules are provided in Appendix~\ref{app:challenge-set}.

\begin{table*}[t]
\caption{\textbf{Human-deceptive videos remain difficult for current detectors.}
RA-Bench-HumanProof contains 633 generated videos, each paired with its matched real anchor. Discrete-output methods report BAcc and FakeR, while continuous-score methods report paired AUC and T@5\%. The final column reports BAcc/FakeR for discrete outputs and AUC/T@5\% for continuous scores on full RA-Bench after weighting its source-specific results by the RA-Bench-HumanProof source proportions. All values are percentages. Skyra official-timestamp results follow the released protocol; frame index is the timestamp-free control.}
\label{tab:human_main}
\centering
\scriptsize
\setlength{\tabcolsep}{8.0pt}
\renewcommand{\arraystretch}{1.08}
\begin{tabular}{lccccc}
\toprule
\multirow{3}{*}{\textbf{Configuration}} &
\multicolumn{4}{c}{\textbf{RA-Bench-HumanProof}} &
\multirow{3}{*}{\makecell{\textbf{Source-matched RA-Bench}\\\textbf{BAcc/FakeR or AUC/T@5\%}}} \\
\cmidrule(lr){2-5}
& \multicolumn{2}{c}{\textbf{Discrete output}} &
\multicolumn{2}{c}{\textbf{Continuous score}} & \\
\cmidrule(lr){2-3}\cmidrule(lr){4-5}
& \textbf{BAcc} & \textbf{FakeR} & \textbf{AUC} & \textbf{T@5\%} & \\
\midrule
\rowcolor{blue!6}
\multicolumn{6}{l}{\textbf{Traditional detectors}} \\
CNNSpot & -- & -- & 40.5 & 3.6 & 43.5 / 3.8 \\
NPR & -- & -- & 50.3 & 3.8 & 50.4 / 4.0 \\
UnivFD & -- & -- & 39.4 & 0.6 & 39.8 / 1.0 \\
ForgeLens & -- & -- & 54.9 & 8.8 & 58.4 / 10.4 \\
DeCoF & -- & -- & 59.0 & 1.9 & 59.2 / 1.7 \\
D3 & -- & -- & 33.1 & 6.8 & 39.5 / 6.2 \\
ReStraV & -- & -- & 55.2 & 5.2 & 55.6 / 4.9 \\
\cmidrule(lr){1-6}
\textbf{7-detector mean} & -- & -- & \textbf{47.5} & \textbf{4.4} & \textbf{49.5 / 4.6} \\
\midrule
\rowcolor{green!6}
\multicolumn{6}{l}{\textbf{Zero-shot multimodal models}} \\
Gemini-3.1-Pro-Preview Binary & 54.7 & 34.3 & -- & -- & 61.2 / 49.9 \\
Gemini-3.1-Pro-Preview Diagnostic & 54.5 & 30.0 & -- & -- & 61.0 / 47.3 \\
Gemini-3.1-Pro-Preview Rating & -- & -- & 54.9 & 4.3 & 61.5 / 5.6 \\
\midrule
\rowcolor{orange!7}
\multicolumn{6}{l}{\textbf{Fine-tuned MLLM detectors}} \\
Skyra-SFT, official timestamp & 72.0 & 55.1 & -- & -- & 73.9 / 60.4 \\
Skyra-SFT, frame index & 53.7 & 51.3 & -- & -- & 55.3 / 55.7 \\
Skyra-RL, official timestamp & 74.5 & 63.0 & -- & -- & 75.1 / 66.3 \\
Skyra-RL, frame index & 53.7 & 58.8 & -- & -- & 56.1 / 63.3 \\
\BusterXpp, released pipeline & 49.4 & 3.9 & -- & -- & 49.9 / 6.2 \\
\bottomrule
\end{tabular}
\end{table*}

The seven traditional detectors average 47.5\% AUC and 4.4\% T@5\% on RA-Bench-HumanProof, compared with 49.5\% and 4.6\% under source-matched RA-Bench weighting. This limited change does not imply reliable detection: both evaluations remain close to random ranking and provide little generated-video recall at a 5\% false-positive rate. Gemini-3.1-Pro-Preview exhibits a clearer association with human difficulty. Binary and Diagnostic BAcc decrease from 61.2\% and 61.0\% to 54.7\% and 54.5\%, while their FakeR decreases from 49.9\% and 47.3\% to 34.3\% and 30.0\%, respectively. Rating AUC decreases from 61.5\% to 54.9\%, and its T@5\% decreases from 5.6\% to 4.3\%.

The fine-tuned MLLMs change by at most 2.4 BAcc points relative to their source-matched references, while FakeR decreases by 2.3--5.3 points across all five configurations. This limited variation does not establish content-based robustness. The official-timestamp Skyra configurations retain the prior identified in Section~\ref{sec:finetuned-shortcut}; replacing timestamps with frame indices reduces both checkpoints to 53.7\% BAcc. \BusterXpp\ reaches 49.4\% BAcc, with 3.9\% FakeR and 94.9\% RealR, indicating a strong tendency to predict \emph{Real}. Neither behavior provides reliable detection of human-deceptive videos.

RA-Bench-HumanProof exposes different failure patterns across detector families: Gemini loses much of its fake-side evidence, traditional detectors remain weak before and after human selection, and the stronger official Skyra results retain the timestamp prior. Human and detector failures therefore overlap, but they are not equivalent. Together, these results show that videos that mislead reviewers are also difficult for current detectors. This overlap is particularly consequential for generated videos depicting crisis events, which repeatedly appear real to reviewers while receiving weak or unreliable fake evidence from the evaluated detector families.

\Needspace{4\baselineskip}
\paragraph{Takeaway.}
RA-Bench-HumanProof remains difficult across detector families: Gemini Binary and Diagnostic reach 54.7\% and 54.5\% BAcc, the seven traditional detectors average 47.5\% AUC, both timestamp-free Skyra checkpoints reach 53.7\% BAcc, and \BusterXpp\ reaches only 3.9\% FakeR.
\FloatBarrier

\subsubsection{Social Dissemination Simulation Further Weakens Detection}
\label{sec:propagation}

Videos rarely reach viewers or moderation systems in their standardized form. We therefore introduce \textbf{RA-Bench-LastMile}, a controlled social dissemination simulation for evaluating detector reliability. It contains 150 real-event anchors spanning 41 of the 44 L2 categories in RA-Bench, together with their matched videos from all nine generation sources. Original retains the standardized clips; T1 applies a common VP9-to-H.264 transcode; T1+T2, T1+T3, and T1+T4 add spatial downsampling, frame-rate reduction, or a synthetic news badge to T1; and Full combines all four operations. Every condition is applied identically to each generated video and its matched real anchor. Appendix~\ref{app:propagation-protocol} provides the complete protocol and detector-level results.

\begin{figure*}[t]
    \centering
    \includegraphics[width=\textwidth]{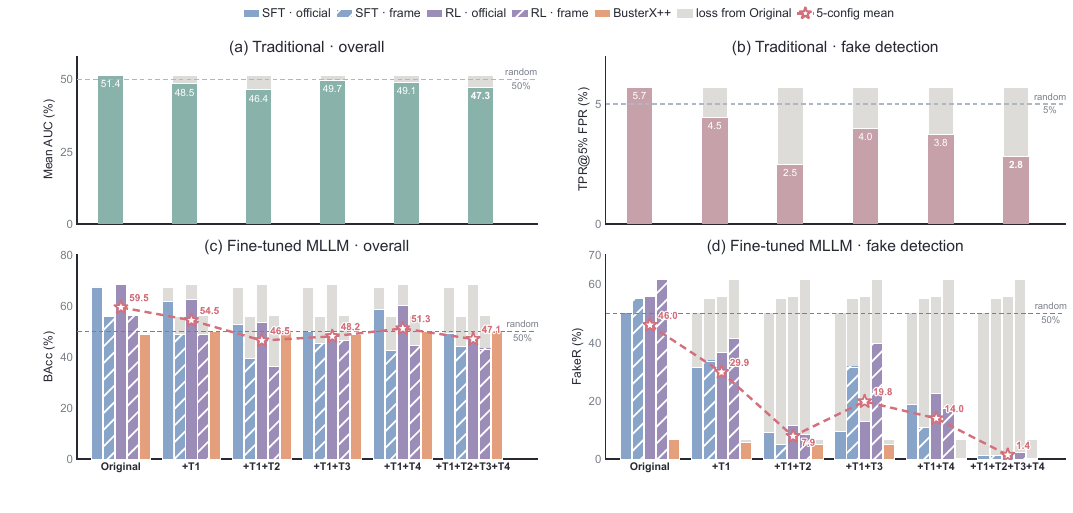}
    \caption{\textbf{Detection under the RA-Bench-LastMile social dissemination simulation.} Original retains the standardized clips; T1 denotes VP9-to-H.264 transcoding, T2 denotes $0.5\times$ spatial downsampling, T3 denotes conversion to 8 fps, and T4 denotes a synthetic news badge. T1+T2, T1+T3, and T1+T4 isolate each added operation, while Full combines T1--T4. (a--b) Mean AUC and T@5\% over the seven traditional detectors. (c--d) BAcc and FakeR for the fine-tuned MLLM detectors. Gray extensions show the loss from each configuration's Original result, hatched bars denote frame-index prompts, and the red dashed line gives the mean over the five fine-tuned configurations. Results are equal-source means over the nine RA-Bench generators.}
    \label{fig:propagation_main}
\end{figure*}

Traditional detectors are already weak on Original and respond inconsistently across the social dissemination simulation. As shown in Figure~\ref{fig:propagation_main}(a--b), their mean AUC decreases from 51.4\% on Original to 48.5\% after T1 and 47.3\% under Full, while mean T@5\% falls from 5.7\% to 2.8\%. Detector-specific responses differ sharply: ForgeLens decreases from 61.6\% to 35.6\%, whereas DeCoF increases from 59.9\% to 62.3\% and D3 from 49.3\% to 54.1\%. The Spearman correlation between the Original and Full AUC rankings is consequently only 0.07. The social dissemination simulation can therefore change which detector appears strongest without yielding reliable separation.

Fine-tuned MLLMs instead show a systematic shift toward \emph{Real} as the operations in the social dissemination simulation accumulate. Averaged over the five configurations, BAcc decreases from 59.5\% on Original to 54.5\% after T1 and 47.1\% under Full; FakeR falls from 46.0\% to 29.9\% and then to 1.4\%. Under Full, the four Skyra configurations retain only 1.2--2.4\% FakeR despite 84.0--97.3\% RealR, while \BusterXpp\ reaches 0.2\% FakeR and 100\% RealR. The near-50\% BAcc therefore reflects a collapse toward predicting \emph{Real}, not preserved generated-video detection.

The isolated additions to T1 clarify which operations drive this shift. Spatial downsampling lowers mean FakeR from 29.9\% to 7.9\%. Conversion to 8 fps reduces BAcc by 11.7--12.9 points for the official-timestamp Skyra configurations, but by only 2.4--3.4 points for their frame-index controls, consistent with the temporal-grid dependence in Section~\ref{sec:finetuned-shortcut}. The news badge lowers mean FakeR from 29.9\% to 14.0\% while increasing mean RealR from 79.1\% to 88.5\%. Because the underlying scene is unchanged, this presentation cue alone shifts predictions toward \emph{Real}. These results show that both signal degradation and presentation changes can suppress evidence for the generated class, and Full combines their effects.

\paragraph{Takeaway.}
On RA-Bench-LastMile, Full reduces mean FakeR across the five fine-tuned configurations from 46.0\% to 1.4\% as their predictions shift toward \emph{Real}.
\FloatBarrier

\subsubsection{Summary of Human Perception and Social Dissemination}
\label{sec:human-reliability-summary}

Human recognition varies substantially across generation sources. Reviewers identify 68.6\% of open-source videos as generated, but only 52.9\% of closed-source videos; the rates fall to 40.7\% for Seedance2.0 and 45.1\% for Kling. The 633 generated videos labeled \emph{Real} by all five reviewers form RA-Bench-HumanProof. On RA-Bench-HumanProof, Gemini Binary and Diagnostic reach only 54.7\% and 54.5\% BAcc, while the seven traditional detectors average 47.5\% AUC and 4.4\% T@5\%. Both timestamp-free Skyra checkpoints reach 53.7\% BAcc, and \BusterXpp\ identifies only 3.9\% of the videos as fake. On RA-Bench-LastMile, Full reduces mean FakeR across the five fine-tuned configurations from 46.0\% to 1.4\% as their predictions shift toward \emph{Real}. Generated videos depicting crisis events can therefore be difficult for both viewers and detectors, and the social dissemination simulation can make them still harder to flag.
\FloatBarrier
\section{Discussion}
\paragraph{Detector reliability is conditional.}
RA-Bench shows that detector reliability cannot be summarized by a single benchmark score. Traditional detectors fall sharply from their public-reference performance and change rank across generation sources, so performance on one generator does not establish transfer to another. Zero-shot multimodal models vary with prompt format and generation source, while MLLMs fine-tuned for AI-generated video detection depend on temporal-label format or strongly favor the \emph{Real} class. No evaluated detector family therefore provides a decision rule that transfers reliably across sources and evaluation protocols. As generators evolve, detectors tuned to fixed sources or protocols may lose reliability. Evaluation should therefore separate source generalization, prompt sensitivity, protocol dependence, and class bias. These results also motivate systems that combine evidence from multiple detectors and forensic tools through modular or agent-based workflows rather than relying on a single classifier.

\paragraph{Generation properties affect detector families differently.}
Within each generation source, stronger Condition Fidelity and temporal consistency are associated with weaker fake evidence, but the most relevant dimensions differ across detector families. Gemini is most sensitive to Condition Fidelity and assigns more fake evidence to dynamic clips, whereas the traditional-detector mean changes more with the VBench-I2V Quality Score, Subject Consistency, and Imaging Quality. Across T2V, first-frame I2V, and first+last-frame I2V, the mean FakeR of the evaluated MLLMs fine-tuned for AI-generated video detection falls from 70.5\% to 42.7\% and 28.3\%, while traditional-detector AUCs move in both directions. Detection difficulty therefore cannot be summarized by a single quality score or a common ordering of generation conditions. The small variation across three sampling seeds further indicates that these patterns are not specific to one sampled realization.

\paragraph{Human deception identifies socially consequential failures.}
Human review reveals a failure mode that detector scores alone cannot capture. Reviewers label 22.8\% of real-video judgments as \emph{Generated}, while 633 generated videos are labeled \emph{Real} by all five assigned reviewers. These 633 videos form \textbf{RA-Bench-HumanProof}, on which Gemini performs near chance under all three prompt formats. Traditional detectors show only a small additional decline, but their source-matched RA-Bench performance is already close to random. RA-Bench-HumanProof therefore isolates generated videos that repeatedly mislead viewers and remain difficult across detector families. When these videos depict crisis events, they may circulate as authentic evidence and distort public understanding before verification.

\paragraph{Social dissemination shifts detector predictions toward \emph{Real}.}
\textbf{RA-Bench-LastMile} uses a social dissemination simulation to measure how detector behavior changes without altering the underlying event content. Under Full, the mean AUC of the seven traditional detectors falls from 51.4\% to 47.3\%, and their ranking changes substantially. The evaluated MLLMs fine-tuned for AI-generated video detection show a sharper failure on generated videos: their mean FakeR falls from 46.0\% to 1.4\% as predictions shift toward \emph{Real}. Downsampling, frame-rate reduction, and a news-style badge each contribute to this shift. Generated videos depicting crisis events may therefore pass automated screening and reach viewers without an authenticity warning, which may give misinformation more time to spread before verification.

\paragraph{Limitations.}
RA-Bench provides a controlled evaluation of AI-generated video detection in real-world crisis settings rather than a complete account of how such content may be misused. First, the current release is necessarily a snapshot of a rapidly changing field. It covers nine I2V generation sources and visual-only detection, but new generators and detectors may quickly change the difficulty of the benchmark. RA-Bench should therefore be maintained through versioned updates that add new generation sources, detection methods, and social dissemination settings. Second, the benchmark studies I2V generation and social dissemination as separate, controlled stages, with the latter represented through a social dissemination simulation. In practice, malicious actors may use multi-stage forgery pipelines that combine generation with selective editing, audio synthesis, contextual captions, and repeated platform processing. Future versions should evaluate these end-to-end workflows while retaining traceable controls over each stage. Finally, human judgments are collected in a controlled interface without the surrounding social context that can influence credibility online.

\section{Future Directions}
Based on our proposed benchmark and comprehensive analysis, we present several promising directions to inspire future research:
    
\begin{itemize}
    \item \textbf{Developing robust detectors for increasingly realistic AI-generated videos.} Although significant progress has been achieved in AI-generated video detection, our analysis demonstrates a substantial performance gap when existing detection methods are applied to realistic videos produced by state-of-the-art generators~\citep{google2024veo2, seedance2026seedance}. More advanced detectors should be designed to 
    leverage multiple cues, such as temporal consistency, lighting variations, and object motion, to ensure robust performance across diverse video scenarios and post-processing operations.

    \item \textbf{Establishing evaluation benchmarks and approaches for detector interpretability.}
    MLLM-based detection methods can provide detailed evidence beyond the simple logits produced by traditional approaches, yet the reliability of such reasoning remains largely unverified. A promising future direction is to construct benchmarks that comprehensively evaluate the correctness and faithfulness of generated evidence. Building upon this, future work could explore MLLM-based agents~\citep{yao2025survey, xie2024large} and post-training approaches~\citep{guo2025deepseek, agarwal2024policy} to decompose AI-generated video detection into structured reasoning steps, thereby further enhancing transparency.

    \item \textbf{Incorporating active watermarking for reliable AI-generated video detection.}
To improve the reliability of AI-generated video detection, active watermarking techniques in video generation models~\citep{fernandez2024video, hu2025videoshield, su2026safe} represent another important research direction. Such methods can embed identifiable signals during generation, making AI-generated videos easier to detect. Our benchmark can be further extended to systematically evaluate the robustness of active watermarking methods against post-processing operations during social dissemination. Additionally, future work should move beyond solely relying on passive detectors and explore their co-design with active watermarking to keep pace with increasingly realistic AI-generated videos.

    \item \textbf{Extending AI-generated video detection to audio-visual settings.}
    Since existing methods mainly rely on visual evidence to detect AI-generated videos, our benchmark focuses on measuring this visual detection ability. However, as video generation models such as Seedance 2.0~\citep{seedance2026seedance} have started to natively generate audio-video aligned content rather than only silent video frames, AI-generated video detection should also move beyond visual cues alone. Future work could jointly explore audio and visual evidence for more reliable and comprehensive AI-generated video detection, and construct corresponding benchmarks to evaluate such multimodal capabilities.

    \item \textbf{Enhancing the realism of video generation models.}
    Although our benchmark is originally designed for detecting AI-generated videos, it can also be leveraged in the opposite direction to promote more realistic video generation. Our analysis reveals that higher-quality AI-generated videos are generally harder for both humans and detectors to identify, suggesting that detection results can serve as a useful signal for measuring generation realism. Therefore, the best-performing existing detectors, as well as future detectors developed under this benchmark, can be used as robust reward models during the post-training process~\citep{xu2026visionreward, liu2026improving} of video generation models.

\end{itemize}

\section[Conclusion \& References]{Conclusion}
We introduce \textbf{RA-Bench}, a benchmark for AI-generated video detection that uses \textbf{R}eal videos as \textbf{A}nchors. RA-Bench contains 17{,}886 videos, comprising 1{,}830 real-video anchors across 10 social-risk categories and 16{,}056 generated clips from four open-source and five closed-source generators. We evaluate detector generalization, generation properties, and human perception and social dissemination. None of the three detector families generalizes consistently across generation sources: traditional-detector performance falls sharply from public-reference results, zero-shot multimodal models remain sensitive to prompts and sources, and fine-tuned MLLMs depend on timestamp cues or exhibit strong class bias. Generation properties also affect detector families differently. Condition Fidelity, temporal consistency, dynamic content, and real-image conditioning show different relationships with traditional-detector and MLLM outputs, while source-level detection patterns remain stable across three sampling seeds. Human review identifies 633 generated videos judged \emph{Real} by all five assigned reviewers, forming \textbf{RA-Bench-HumanProof}, on which Gemini performs near chance under all three prompt formats. \textbf{RA-Bench-LastMile} further shows that the social dissemination simulation reduces mean FakeR across the evaluated MLLMs fine-tuned for AI-generated video detection from 46.0\% to 1.4\%. Together, these results show that current methods still cannot reliably detect realistic AI-generated videos in real-world crisis settings.
\FloatBarrier

{\small
\phantomsection
\bibliographystyle{bibstyle}
\bibliography{iclr2026_conference}
}

\clearpage
\appendix
\addtocontents{toc}{\protect\mainappendixgap}
\thispagestyle{empty}
\appendixcontents
\clearpage

\clearpage
\appendixsection{RA-Bench Construction and Data Documentation}

\appendixsubsection{L2 Taxonomy and Source Distribution}
\label{app:l2-taxonomy}

Section~\ref{sec:data-sources} organizes the 675 collected source videos using a two-level taxonomy. The 10 L1 domains represent broad areas of social risk, while the 44 L2 categories specify the event scope within each domain. Table~\ref{tab:l2-taxonomy} provides the complete category definitions and source counts.

Taxonomy assignments are made before scene segmentation and manual review. The counts therefore refer to source videos rather than review clips or released anchors. We retain the collected source distribution without imposing uniform category quotas or rebalancing at this stage.

\begin{table}[H]
\setlength{\abovecaptionskip}{2pt}
\setlength{\belowcaptionskip}{3pt}
\caption{\textbf{L2 taxonomy and source distribution of RA-Bench.}
Counts refer to the 675 collected source videos before scene segmentation and manual review. L1 subtotals are shown in parentheses.}
\label{tab:l2-taxonomy}
\centering
\fontsize{7pt}{8pt}\selectfont
\setlength{\tabcolsep}{2.4pt}
\setlength{\defaultaddspace}{2pt}
\renewcommand{\arraystretch}{0.86}
\begin{tabular}{@{}lp{0.76\linewidth}c@{}}
\toprule
L2 & Event scope & \# \\
\midrule
\multicolumn{3}{@{}l}{\textbf{L1-01 Weather and natural disasters} (75)} \\
L2-01a & Windstorms, hurricanes, and tornadoes & 17 \\
L2-01b & Wildfire spread & 15 \\
L2-01c & Earthquakes, building damage, and post-disaster street scenes & 23 \\
L2-01d & Flooding & 20 \\
\addlinespace
\multicolumn{3}{@{}l}{\textbf{L1-02 War and armed conflict} (60)} \\
L2-02a & Battlefield overviews and urban war damage & 15 \\
L2-02b & Airstrikes, missile launches, and distant explosions & 15 \\
L2-02c & Military aircraft, naval vessels, and armored vehicle formations & 15 \\
L2-02d & Official military exercises, parades, and public training & 15 \\
\addlinespace
\multicolumn{3}{@{}l}{\textbf{L1-03 Politics and governance} (60)} \\
L2-03a & Official press briefings and public announcements & 15 \\
L2-03b & Parliamentary and government-hall meetings & 15 \\
L2-03c & Public speeches by political leaders & 15 \\
L2-03d & Diplomatic meetings, signing ceremonies, and summit photo calls & 15 \\
\addlinespace
\multicolumn{3}{@{}l}{\textbf{L1-04 Public safety} (69)} \\
L2-04a & Urban lockdowns, police lines, and security deployment & 15 \\
L2-04b & Urban unrest and arson scenes in distant, CCTV, or aerial views & 24 \\
L2-04c & Counter-terrorism drills and large-scale emergency evacuation & 15 \\
L2-04d & Search-and-rescue operations in mountains, at sea, or urban areas & 15 \\
\addlinespace
\multicolumn{3}{@{}l}{\textbf{L1-05 Accidents and infrastructure failures} (93)} \\
L2-05a & Aviation incidents, including emergency landings, runway excursions, and airport emergency response & 22 \\
L2-05b & Rail transit incidents, including train derailments and subway evacuation & 11 \\
L2-05c & Highway multi-vehicle crashes and large bus or truck accidents in distant or aerial views & 16 \\
L2-05d & Industrial disasters, including factory explosions, fires, and chemical leaks & 17 \\
L2-05e & Major infrastructure damage, including bridge or high-rise collapse and repair operations & 27 \\
\addlinespace
\multicolumn{3}{@{}l}{\textbf{L1-06 Economic and social panic} (60)} \\
L2-06a & Cash-withdrawal queues or bank-run scenes outside bank branches & 12 \\
L2-06b & Trading floors or market screens showing abrupt stock-market drops & 12 \\
L2-06c & Empty supermarket shelves and panic buying & 12 \\
L2-06d & Energy shortages, including gas-station queues and city-scale power restrictions & 12 \\
L2-06e & Official or regulatory briefings and hearings on financial crises & 12 \\
\addlinespace
\multicolumn{3}{@{}l}{\textbf{L1-07 Public health} (60)} \\
L2-07a & Official briefings on epidemics and public-health emergencies & 15 \\
L2-07b & Hospital exteriors, emergency entrances, and ambulance dispatch without patient close-ups & 15 \\
L2-07c & Mass vaccination sites and medical queues in distant group views & 15 \\
L2-07d & City-scale public-health measures, including disinfection and health checkpoints & 15 \\
\addlinespace
\multicolumn{3}{@{}l}{\textbf{L1-08 Technology} (60)} \\
L2-08a & Technology or AI product launches with stage demonstrations & 13 \\
L2-08b & Robots and autonomous vehicles in public spaces, such as inspection or delivery & 18 \\
L2-08c & Network outages and data-center failure scenes, including machine-room and operations views & 16 \\
L2-08d & Technology and developer conference demos & 13 \\
\addlinespace
\multicolumn{3}{@{}l}{\textbf{L1-09 Space, exploration, and anomalies} (63)} \\
L2-09a & Rocket launches, booster recovery, and launch-site views & 12 \\
L2-09b & Space-station exterior views and public-license in-orbit satellite video & 13 \\
L2-09c & Rare astronomical phenomena, including eclipses and meteor showers & 14 \\
L2-09d & Polar and deep-sea scientific exploration video & 12 \\
L2-09e & Aerial anomalies & 12 \\
\addlinespace
\multicolumn{3}{@{}l}{\textbf{L1-10 Large public events} (75)} \\
L2-10a & Major sports broadcasts, including wide views and scoreboards & 15 \\
L2-10b & Post-game celebrations and fan gatherings in public squares & 15 \\
L2-10c & Large concerts and music festivals with stage and crowd views & 15 \\
L2-10d & Award ceremonies, premieres, and red-carpet events centered on public figures & 15 \\
L2-10e & City-scale festivals, lantern shows, New Year fireworks, and public celebrations & 15 \\
\midrule
\textbf{Total} & & \textbf{675} \\
\bottomrule
\end{tabular}
\end{table}

\clearpage
\appendixsubsection{Source Inventory and Rights Basis} \label{app:source-inventory}

Table~\ref{tab:source-inventory} summarizes the 675 source videos by provenance group and major contributor. Government and public-institution sources contribute 348 videos, and open-license repositories contribute 191; together they account for 539 of the 675 source videos (79.9\%). These counts are measured before scene segmentation and do not depend on how many clips are later retained from each source video.

The \emph{Rights basis} column condenses the source-level provenance and terms recorded during collection; it does not by itself establish permission to redistribute a clip. An aggregated row may contain more than one rights basis, while the final redistribution decision is made separately for each source during the rights review described in Appendix~\ref{app:postprocessing-details}.

\begin{table}[H]
\caption{\textbf{Source inventory and recorded rights basis.}
Counts refer to source videos before scene segmentation. Combined entries (e.g., ``public domain / official reuse'') indicate that the aggregated row contains sources with different recorded rights bases.}
\label{tab:source-inventory}
\centering
\scriptsize
\setlength{\tabcolsep}{2.4pt}
\renewcommand{\arraystretch}{0.94}
\begin{tabular}{@{}p{0.48\linewidth}cp{0.35\linewidth}@{}}
\toprule
Source & \# & Rights basis \\
\midrule
\multicolumn{3}{@{}l}{\textbf{Government \& public institutions} (348)} \\
\quad DVIDS & 240 & US public domain \\
\quad The White House & 22 & Open / free license \\
\quad NTSB & 16 & Platform terms \\
\quad EU institutions (Council, Parliament, Commission) & 38 & Official reuse terms \\
\quad NASA (incl.\ SVS) & 14 & Public domain / official reuse \\
\quad NOAA & 10 & US public domain \\
\quad Other agencies (USGS, NSF, Defense.gov) & 8 & Public domain / official reuse \\
\addlinespace
\multicolumn{3}{@{}l}{\textbf{Open-license repositories} (191)} \\
\quad Wikimedia Commons & 131 & Open / free license, public domain \\
\quad Pexels & 60 & Pexels license \\
\addlinespace
\multicolumn{3}{@{}l}{\textbf{News \& broadcast platforms} (105)} \\
\quad YouTube (general uploads) & 75 & Platform terms \\
\quad Reuters (Video, YouTube) & 16 & Platform / editorial terms \\
\quad Associated Press (AP, AP Archive) & 13 & Editorial terms \\
\quad Other broadcast & 1 & Editorial terms \\
\addlinespace
\multicolumn{3}{@{}l}{\textbf{Other official channels} (31)} \\
\quad Corporate channels (Microsoft, Google, AWS, NVIDIA, OpenAI, Meta, Apple, Samsung, and others) & 19 & Official reuse / case-by-case \\
\quad Transportation \& organizational channels (state DOTs, airports, others) & 12 & Platform terms \\
\midrule
\textbf{Total} & \textbf{675} & \\
\bottomrule
\end{tabular}
\end{table}

\FloatBarrier
\appendixsubsection{Automated Preprocessing Details} \label{app:preprocessing-details}

\paragraph{Scene segmentation.} We apply PySceneDetect's
\texttt{ContentDetector} with a threshold of \(27.0\) and a minimum scene length
of \(15\) frames. Detected scenes shorter than \(5.0\,\)s are merged with an
adjacent segment to avoid fragmenting a coherent event into very short clips.
Detection uses the PyAV decoding backend. We then split the source videos with
FFmpeg using re-encoding rather than stream copy, which avoids restricting the cut
points to existing keyframes. Applied to the \(675\) source videos, this procedure
produces the \(5{,}774\) clips reviewed in Section~\ref{sec:review}.

\paragraph{Near-duplicate prefilter.} We uniformly sample \(8\) frames from each
clip and encode them with a ResNet-18 pretrained on ImageNet-1K. Comparisons are
restricted to the next \(4\) clips from the same source video, since the main source
of local redundancy is repeated content across adjacent scenes. Two sampled frames
are treated as a match when their cosine similarity is at least \(0.90\). A clip
pair is then flagged only when its bidirectional matched-frame ratio is at least
\(0.50\), its mean cosine similarity over matched frames is at least \(0.90\), and
its clip-level cosine similarity is at least \(0.92\). The prefilter produces
\(1{,}269\) pairwise duplicate warnings grouped into \(350\) clusters. These
warnings are advisory: no clip is removed automatically, and the corresponding
pairs are presented to the Round~1 reviewers for confirmation.

\FloatBarrier
\appendixsubsection{Manual Review Details} \label{app:review-details}

\paragraph{Round~1 review.} Seven volunteers review all \(5{,}774\) clips. Each
clip is assessed independently by two reviewers, yielding \(11{,}548\)
clip--reviewer decisions. Assignments are balanced by semantic subcategory and
automatic prefilter status so that no reviewer receives a disproportionate share
of any event type or of the suspected near-duplicates.

\paragraph{Review criteria and actions.} Reviewers assess visual quality,
semantic fit to the assigned subcategory, duration suitability, duplicate content,
and source-related risks, such as privacy-sensitive content or intrusive platform
overlays. Each reviewer assigns one of three actions:
\emph{retain}, \emph{reject}, or \emph{uncertain}. The near-duplicate warnings from
Appendix~\ref{app:preprocessing-details} serve only as review aids; the reviewers
make the inclusion decision from the video content and the predefined criteria.

\paragraph{Round~2 adjudication.} A clip is routed to Round~2 when the two
Round~1 reviewers disagree or when either reviewer selects \emph{uncertain}. Four
additional volunteers serve as adjudicators, with each routed clip assigned to one
adjudicator for a final decision. After adjudication, \(2{,}426\) clips are retained
as the standard sample pool and \(3{,}348\) are rejected. Thus, all \(5{,}774\)
clips receive a resolved binary outcome before the retained clips enter
postprocessing (Section~\ref{sec:postprocess};
Appendix~\ref{app:postprocessing-details}).

\FloatBarrier
\appendixsubsection{Postprocessing Details}
\label{app:postprocessing-details}
This subsection reports the exact count-changing operations, encoding settings,
and source-level rights review used in Section~\ref{sec:postprocess}.

\paragraph{Duration bounding.} Each clip is bounded to a 3--15\,s window: clips
shorter than 3\,s are discarded, while clips longer than 15\,s are truncated. This
step removes one clip and truncates 444, leaving 2{,}425 length-bounded clips.

\paragraph{Encoding standardization.} We re-encode the video stream of every
retained clip with H.264 using \texttt{libx264}, preset \texttt{slow}, and CRF
\(17\). The same release encoding is later applied to generated clips so that codec
choice and encoder configuration do not serve as label cues.

\paragraph{Homogeneity pruning.} We apply the mutually exclusive per-source
schedule to reduce overrepresented uploads without imposing equal category quotas.
Sources with at most 10 retained clips are unchanged. For sources with 11--20,
21--30, 31--40, and 41--50 clips, we remove one clip in every 5, 4, 3, and 2,
respectively; for sources with more than 50 clips, we remove approximately 70\% at
random. Density pruning removes \(519\) clips. Removing seven additional groups of
manually identified repetitive clips discards another \(76\), for \(595\) removals
in total.

\paragraph{Rights review.} Rights are determined per source, with one decision
covering all clips from that source. Two of the seven reviewers handled a manual
queue of \(247\) sources whose licenses were not directly determinable; the
remaining sources had determinable licenses from their origin. Each anchor
carries its documented rights basis and release mode as metadata. The public
release distributes 1{,}319 anchors with \texttt{public\_cleared} or
\texttt{public\_conditional} status as media; the remaining 511 are represented
by metadata and source URLs only.

\paragraph{Count reconciliation.} Starting from the \(2{,}426\) clips retained
after manual review, duration bounding removes one clip and homogeneity pruning
removes \(595\). The resulting real set contains \(1{,}830\) clips from \(338\)
source videos, totaling \(18{,}448\,\)s with a mean duration of \(10.08\,\)s.

\FloatBarrier
\appendixsubsection{Generation Details} \label{app:generation-details}

\definecolor{vfdStageA}{HTML}{7776A6}
\definecolor{vfdStageB}{HTML}{C97963}
\definecolor{vfdExample}{HTML}{4D8294}
\definecolor{vfdInk}{HTML}{24384B}

\tcbset{
  vfd prompt/.style={
    enhanced, listing only,
    listing options={basicstyle=\scriptsize\ttfamily, breaklines=true,
      breakindent=0pt, columns=fullflexible, keepspaces=true,
      showstringspaces=false},
    coltitle=vfdInk, fonttitle=\bfseries\footnotesize,
    boxrule=0.6pt, arc=1.5pt,
    left=5pt, right=5pt, top=2pt, bottom=2pt,
    before skip=6pt, after skip=8pt
  }
}
\newtcblisting{stageapromptbox}[1]{
  vfd prompt,
  colback=vfdStageA!4, colframe=vfdStageA!75!black,
  colbacktitle=vfdStageA!18,
  title={#1}
}
\newtcblisting{stagebpromptbox}[1]{
  vfd prompt,
  colback=vfdStageB!4, colframe=vfdStageB!75!black,
  colbacktitle=vfdStageB!18,
  title={#1}
}
\newtcolorbox{vfdexamplebox}[1]{
  enhanced,
  colback=vfdExample!3, colframe=vfdExample!75!black,
  colbacktitle=vfdExample!17, coltitle=vfdInk,
  fonttitle=\bfseries\footnotesize,
  fontupper=\footnotesize,
  title={#1}, boxrule=0.6pt, arc=1.5pt,
  left=7pt, right=7pt, top=5pt, bottom=5pt,
  before skip=6pt, after skip=8pt
}

\paragraph{Two-stage caption and prompt construction.}
Captioning and prompt construction use Gemini 3.1 Pro
Preview~\citep{google2026gemini31pro}, queried through its native video endpoint
under the model identifier \texttt{gemini-3.1-pro-preview}.
Stage~A converts the
audio-stripped real anchor and anonymized clip metadata into the structured visual
caption defined in Table~\ref{tab:caption-schema}. Stage~B converts this caption and
the target duration into a generator-agnostic English prompt, a negative prompt, a
Chinese translation retained only for bookkeeping, and a duration hint. The English
prompt and negative prompt are shared across generation sources. In the reproduced
instructions, ``text-to-video prompt'' refers to this textual condition; every
benchmark generation call also receives the real anchor's first frame and is
executed as I2V.

\paragraph{Structured output and interface.}
The Stage~A user message supplies only anonymized identifiers, L1/L2 labels and
definitions, duration, frame rate, and resolution. Its output follows
Table~\ref{tab:caption-schema}. Stage~B receives this JSON together with
\texttt{target\_duration\_sec}. Numeric duration is excluded from the generation
prompt itself and appears only in \texttt{duration\_hint}, formatted as
\texttt{"\{t\}s"}.

\begin{table}[H]
\caption{\textbf{Stage~A structured caption schema.}
The six core fields describe visible content; the auxiliary fields support prompt
construction, uncertainty tracking, and downstream claim-level analysis.}
\label{tab:caption-schema}
\centering
\footnotesize
\setlength{\tabcolsep}{4pt}
\renewcommand{\arraystretch}{1.02}
\begin{tabular}{@{}p{0.32\linewidth}p{0.62\linewidth}@{}}
\toprule
Field & Content \\
\midrule
\multicolumn{2}{@{}l}{\textcolor{vfdStageA}{\textbf{Core visual fields}}} \\
\texttt{short\_caption}        & one-sentence summary \\
\texttt{main\_object\_caption} & foreground subject(s) \\
\texttt{background\_caption}   & scene / setting \\
\texttt{camera\_caption}       & shot type and camera motion \\
\texttt{style\_caption}        & capture / visual style \\
\texttt{action\_caption}       & event and visible motion \\
\addlinespace
\multicolumn{2}{@{}l}{\textcolor{vfdExample}{\textbf{Auxiliary fields}}} \\
\texttt{dense\_caption}        & integrated 120--200 word paragraph \\
\texttt{text\_in\_video}       & on-screen text as \{text, position, role\}, PII redacted \\
\texttt{salient\_entities}     & common-noun entity list \\
\texttt{salient\_actions}      & visible-action list \\
\texttt{claim\_candidates}     & \{claim, evidence\_type, confidence\} list \\
\texttt{uncertainty\_notes}    & free-text unknowns \\
\texttt{category\_consistency} & \{matches\_declared\_category, notes\} \\
\bottomrule
\end{tabular}
\end{table}

\paragraph{Exact prompt text.}
The two boxes below reproduce the complete Stage~A and Stage~B messages. The
bracketed headers separate the system instruction from the user template for
presentation and were not included in either message. Braced expressions denote
fields populated separately for each clip.

\begin{stageapromptbox}{Stage A: Structured visual captioning}
[SYSTEM INSTRUCTION]
You are an expert video annotator constructing a research benchmark for AI-generated
video detection. Your task is to produce a structured, visually grounded caption from
a real-world video clip. The caption will serve as benchmark metadata and will be used to derive the text
prompt for video generation models (Wan, Hunyuan, LTX, Veo, Kling, Sora).

Adopt the following principles strictly:
1. VISUAL-ONLY. The input video has no audio track. Describe only what is visible.
2. RE-GENERATION ORIENTED. The caption must let a text-to-video model reproduce a
   visually equivalent clip. Prioritize subject appearance, action, location, lighting,
   camera behavior, and visible motion.
3. GROUNDED. Describe only what is directly observable. Do not infer news context,
   identities, causal explanations, or event significance from visual style alone.
   When an attribute is not visually supported, mark it "unknown" in uncertainty_notes.
4. NEUTRAL TONE. Avoid evaluative adjectives (tragic, shocking, heroic). Stay descriptive.
5. NO NAMED IDENTITIES. Do not name real people, organizations, locations, countries,
   brands, or events, even if recognizable. Use generic descriptors.
6. PRIVACY-SAFE ON-SCREEN TEXT. Record visible burned-in text in text_in_video, but
   redact identifying strings and PII with bracketed generic tokens.
7. SHOT STRUCTURE AWARENESS. State whether the clip is one continuous shot or contains
   hard cuts / multiple distinct shots.
8. L2 CATEGORY CONSISTENCY. Judge category_consistency against the declared L2 label,
   not only the broad L1 category.
9. TEMPORAL COVERAGE. Describe the clip as a whole, including visible temporal progression.
10. ENGLISH OUTPUT. All caption fields are in English.

Output a single JSON object conforming to the schema. No preamble, no code fences.

[USER TEMPLATE]
<task>
Annotate the attached real-world video clip into a structured caption following the
MiraData / Any2Caption schema. The caption is benchmark metadata and the source for
downstream text-to-video prompt construction.
</task>

<context>
clip_id: {clip_id}
parent_video_id: {parent_video_id}
category_l1: {l1_code} ({l1_label})
category_l2: {l2_code} ({l2_label})
category_l2_definition: {l2_label}
clip_duration_sec: {duration_sec}
fps: {fps}
resolution: {width}x{height}
notes: sensitive-impact taxonomy; apply identity-free anonymization.
</context>

Return a single JSON object with the 13 schema fields (six core caption fields plus
seven auxiliary fields).
\end{stageapromptbox}

\begin{stagebpromptbox}{Stage B: Generator-agnostic prompt construction}
[SYSTEM INSTRUCTION]
You convert a structured video caption into a single generator-agnostic text prompt for
the text-to-video generation stage of an AI-generated video detection benchmark.

Single objective: produce one English prompt, plus one Chinese translation for metadata
bookkeeping, from the same scene description and target duration for use with
text-to-video models (Wan, Hunyuan, LTX, Veo, Kling, Sora). The same prompt is sent to
every generator so prompt wording is fixed across the comparison; do not tailor wording
to any single generator.

Strict requirements for the English prompt:
- GROUNDED. Use only content present in the source structured caption.
- COMPLETE. Integrate subject, action, setting, camera, style, and lighting/quality.
- CONCISE. Target 100-180 words, ideally 120-160.
- VISUAL. Concrete nouns, verbs, spatial / lighting / camera terms.
- IDENTITY-FREE. No real names, brands, places, countries, organizations, events,
  phone numbers, addresses, license plates, or bracket tokens.
- TEMPORALLY SPECIFIC BUT NOT NUMERIC. Use natural pace words, no numeric durations.
- SINGLE-SHOT. Assume one continuous take.
- VISUAL-ONLY. Do not describe sounds, dialogue, music, or audio cues.
- GENERATOR-NEUTRAL. Plain descriptive English that common T2V encoders can parse.

Output a single JSON object: English prompt, Chinese translation, negative prompt,
duration hint.

[USER TEMPLATE]
<source_structured_caption>
{stage_a_json}
</source_structured_caption>

<generation_target>
target_duration_sec: {target_duration_sec}
</generation_target>

<instructions>
Compose the English generation prompt as a single coherent paragraph ordered as:
1. [Subject + Action]   2. [Setting + Environment]   3. [Camera]   4. [Style]

Length 100-180 words (ideally 120-160); never below 80. Do not pad.

SINGLE-SHOT FRAMING: if the source caption mentions cuts, describe only the first shot.

Then produce:
- generation_prompt_zh: a faithful Chinese translation of generation_prompt_en.
- negative_prompt: a generator-neutral English negative prompt, <=40 words.
- duration_hint: format "{target_duration_sec:g}s".

Hard rules:
- Do NOT introduce content absent from the caption, or copy from claim_candidates.
- Do NOT use named identities, places, brands, organizations, events, or bracket tokens.
- Do NOT mention audio, dialogue, music, or numeric seconds in the prompts.
- Do NOT tailor the prompt to any specific generator. Output JSON only, no fences.
</instructions>
\end{stagebpromptbox}

\clearpage
\paragraph{End-to-end worked example.}
The following example shows how a Stage~A visual description is converted into the
shared Stage~B prompt. Only the short-caption field is shown for Stage~A; the full
record follows Table~\ref{tab:caption-schema}.

\begin{vfdexamplebox}{Worked example: structured caption to generation prompt}
\begin{tabular}{@{}p{0.55\linewidth}p{0.35\linewidth}@{}}
\textbf{Clip ID} & \textbf{Duration hint} \\
\texttt{L1-01\_L2-01a\_0003\_\_scene\_003} & \texttt{6.602s} \\
\end{tabular}

\medskip
{\color{vfdStageA}\bfseries Stage A output: \texttt{short\_caption}}\par
An aerial view from an aircraft window showing a spinning turboprop engine against
a backdrop of dense clouds.

\medskip
\noindent\textcolor{vfdExample!45}{\rule{\linewidth}{0.5pt}}
\medskip

{\color{vfdStageB}\bfseries Stage B output: \texttt{generation\_prompt\_en}}\par
On the right side of the frame, a turboprop engine with a light-colored cylindrical
housing and dark propeller blades operates continuously. Throughout the clip, the
dark blades spin rapidly, creating a distinct motion blur as the aircraft maintains
a steady forward flight path. The scene is an aerial view set high above a vast,
textured layer of whitish-grey clouds below. In the upper sky, a dense layer of dark
grey clouds looms, while a faint band of orange light glows at the distant horizon
under natural, muted lighting. The shot is captured from an interior passenger
perspective, looking out through an aircraft window. The camera remains mostly
static, featuring only minor, steady vibrations typical of in-flight recording. The
footage is presented in an amateur smartphone video style, exhibiting slight
low-light noise in the darker cloud areas alongside the realistic motion blur of the
active propeller.

\medskip
{\color{vfdInk}\bfseries Stage B output: \texttt{negative\_prompt}}\par
low quality, blurry, distorted face, warped hands, extra limbs, flicker,
oversaturation, text watermark, logo, jittery camera, cartoon, anime, painting,
illustration.
\end{vfdexamplebox}

\paragraph{Generator execution.}
All generation sources receive the same Stage~B prompt and the real anchor's first
frame. Table~\ref{tab:generation-source-outputs} summarizes the generator,
request settings, realized output format, and retained counts. Open-source settings
follow Wan2.2~\citep{wan2025wan}, LTX-2~\citep{hacohen2026ltx2}, and
OmniWeaving~\citep{pan2026omniweaving}; provider calls follow the corresponding
HappyHorse~\citep{cloudflare2026happyhorse10i2v}, Runway~\citep{runway2026modelsapi},
Kling~\citep{klingai2026openplatform}, Seedance~\citep{seedance2026seedance}, and
Hailuo~\citep{minimax_hailuo23_2025} interfaces. Internal inference parameters are
available only for the locally executed sources and are reported separately in
Table~\ref{tab:open-source-inference}.

\paragraph{Duration mapping.}
For an anchor of length \(t_{\text{real}}\in[3,15]\)\,s, the shared target is
\(t_{\text{gen}}=\mathrm{clamp}(t_{\text{real}}\cdot 8/15,\,2,\,8)\)\,s.
This target preserves duration variation across anchors while remaining compatible
with the operating ranges of the evaluated generators.

Each source then maps \(t_{\text{gen}}\) to its supported temporal grid. The Wan2.2
and OmniWeaving sources use a \(4n{+}1\) grid at 16\,fps, whereas LTX uses an
\(8n{+}1\) grid at 24\,fps; both policies impose a floor of 33 frames. The separate
Wan2.2 fixed-duration control uses 81 frames (5.06\,s) and is excluded from the
benchmark count. Seedance accepts integer durations from 4 to 15\,s and therefore
realizes 4--8\,s under our target policy.

The Hailuo 768P I2V endpoint exposes only 6- and 10-second outputs.\footnote{MiniMax
Image-to-Video API documentation: \url{https://platform.minimax.io/docs/api-reference/video-generation-i2v}.}
We therefore use 7.5\,s as the boundary: target durations at or above 7.5\,s are
generated as 10-second clips, whereas shorter targets are generated as 6-second
clips. This provider-specific discretization explains why Hailuo does not follow the
continuous 2--8\,s target range reported for the shared policy.

\begin{table}[H]
\caption{\textbf{Generation sources and realized outputs.}
Execution strings and request settings are taken from the recorded generation manifests.
\(n\) counts clips in the primary paired benchmark set; additional seed repeats and
the Wan2.2 fixed-duration control are excluded from the 16{,}056 benchmark clips.}
\label{tab:generation-source-outputs}
\centering
\footnotesize
\setlength{\tabcolsep}{3pt}
\renewcommand{\arraystretch}{1.04}
\begin{tabular}{@{}
  >{\raggedright\arraybackslash}p{0.14\linewidth}
  >{\raggedright\arraybackslash}p{0.30\linewidth}
  >{\centering\arraybackslash}p{0.13\linewidth}
  c
  >{\raggedright\arraybackslash}p{0.18\linewidth}
  c@{}}
\toprule
Source & Execution string / request setting & Output res. & fps & Temporal setting & \(n\) \\
\midrule
\rowcolor{vfdStageA!10}
\multicolumn{6}{@{}l}{\textit{Open-source benchmark generators}} \\
Wan2.2 dynamic & \texttt{Wan2.2-I2V-A14B} & 832$\times$480 & 16 & $4n{+}1$ frames ($\geq$33) & 1{,}830 \\
Wan2.2-Lightning & \makecell[l]{\texttt{Wan2.2-Lightning-}\\\texttt{I2V-A14B-NFE4-V1}} & 1280$\times$720 & 16 & $4n{+}1$ frames ($\geq$33) & 1{,}830 \\
LTX & \texttt{ltx-2.3-22b-dev} & 1536$\times$1024 & 24 & $8n{+}1$ frames ($\geq$33) & 1{,}830 \\
OmniWeaving & \texttt{HY-OmniWeaving} & 848$\times$480 & 16 & $4n{+}1$ frames ($\geq$33) & 1{,}830 \\
\midrule
\rowcolor{vfdStageB!9}
\multicolumn{6}{@{}l}{\textit{Closed-source benchmark generators}} \\
HappyHorse & \makecell[l]{\texttt{happyhorse-1.0-i2v}\\720P, 16:9} & 1264$\times$730 & 24 & 3--8\,s & 1{,}787 \\
Runway & \makecell[l]{\texttt{gen4.5}\\1280:720} & 1280$\times$720 & 24 & 2--8\,s & 1{,}805 \\
Kling & \makecell[l]{\texttt{kling-v3}\\mode \texttt{std}} & 1264$\times$728 & 24 & 3--8\,s & 1{,}790 \\
Seedance2.0 & \makecell[l]{\texttt{doubao-seedance-}\\\texttt{2-0-260128}; ratio-conditioned} & mixed\textsuperscript{\dag} & 24 & 4--8\,s & 1{,}524 \\
Hailuo & \makecell[l]{\texttt{MiniMax-Hailuo-}\\\texttt{2.3-Fast}; 768P} & 1330$\times$768 & 24 & 6 or 10\,s & 1{,}830 \\
\midrule
\rowcolor{vfdExample!9}
\multicolumn{6}{@{}l}{\textit{Auxiliary duration control}} \\
Wan2.2 fixed\textsuperscript{*} & \texttt{Wan2.2-I2V-A14B} & 832$\times$480 & 16 & 81 frames (5.06\,s) & 1{,}830 \\
\bottomrule
\end{tabular}
\parbox{\linewidth}{\vspace{2pt}\footnotesize
\textsuperscript{\dag}Seedance outputs comprise 1{,}452 clips at
1280$\times$720, 41 at 960$\times$960, and 31 at 720$\times$1280.
\textsuperscript{*}The fixed-duration clips are used only as an auxiliary control.
Each provider was queried on all 1{,}830 anchors; smaller retained counts reflect
provider-specific content-safety rejections rather than deliberate subsampling.}
\end{table}

\begin{table}[H]
\caption{\textbf{Internal inference settings for locally executed sources.}
Provider APIs do not expose corresponding sampler-level controls. The dynamic and
fixed Wan2.2 settings differ only in their temporal policy.}
\label{tab:open-source-inference}
\centering
\footnotesize
\setlength{\tabcolsep}{3pt}
\renewcommand{\arraystretch}{1.04}
\begin{tabular}{@{}
  >{\raggedright\arraybackslash}p{0.18\linewidth}
  >{\raggedright\arraybackslash}p{0.17\linewidth}
  c
  >{\raggedright\arraybackslash}p{0.51\linewidth}@{}}
\toprule
Source & Conditioning size & Steps & Sampler / inference setting \\
\midrule
Wan2.2 dynamic/fixed & output resolution & 40 & UniPC; shift 5.0; guidance (3.5, 3.5) \\
Wan2.2-Lightning & output resolution & 4 & Euler + 4-step LoRA; guidance (1.0, 1.0) \\
LTX & 768$\times$512 & 30 & distilled LoRA 0.8; cfg/STG/rescale 3.0/1.0/0.7 \\
OmniWeaving & output resolution & 30 & 480p, 16:9 \\
\bottomrule
\end{tabular}
\parbox{\linewidth}{\vspace{2pt}\footnotesize\raggedright
LTX generates at its base resolution and applies
\nolinkurl{ltx-2.3-spatial-upscaler-x2-1.1} to reach 1536$\times$1024; its text
model is \nolinkurl{gemma-3-12b-it-qat-q4_0-unquantized}. Wan2.2-Lightning applies
\nolinkurl{Wan2.2-I2V-A14B-4steps-lora-rank64-Seko-V1} to the
\texttt{Wan2.2-I2V-A14B} base.}
\end{table}

\FloatBarrier
\Needspace{13\baselineskip}
\appendixsubsection{Qualitative Real-versus-Generated Examples} \label{app:qualitative-examples}

\begin{figure}[H]
\centering
\includegraphics[width=\linewidth]{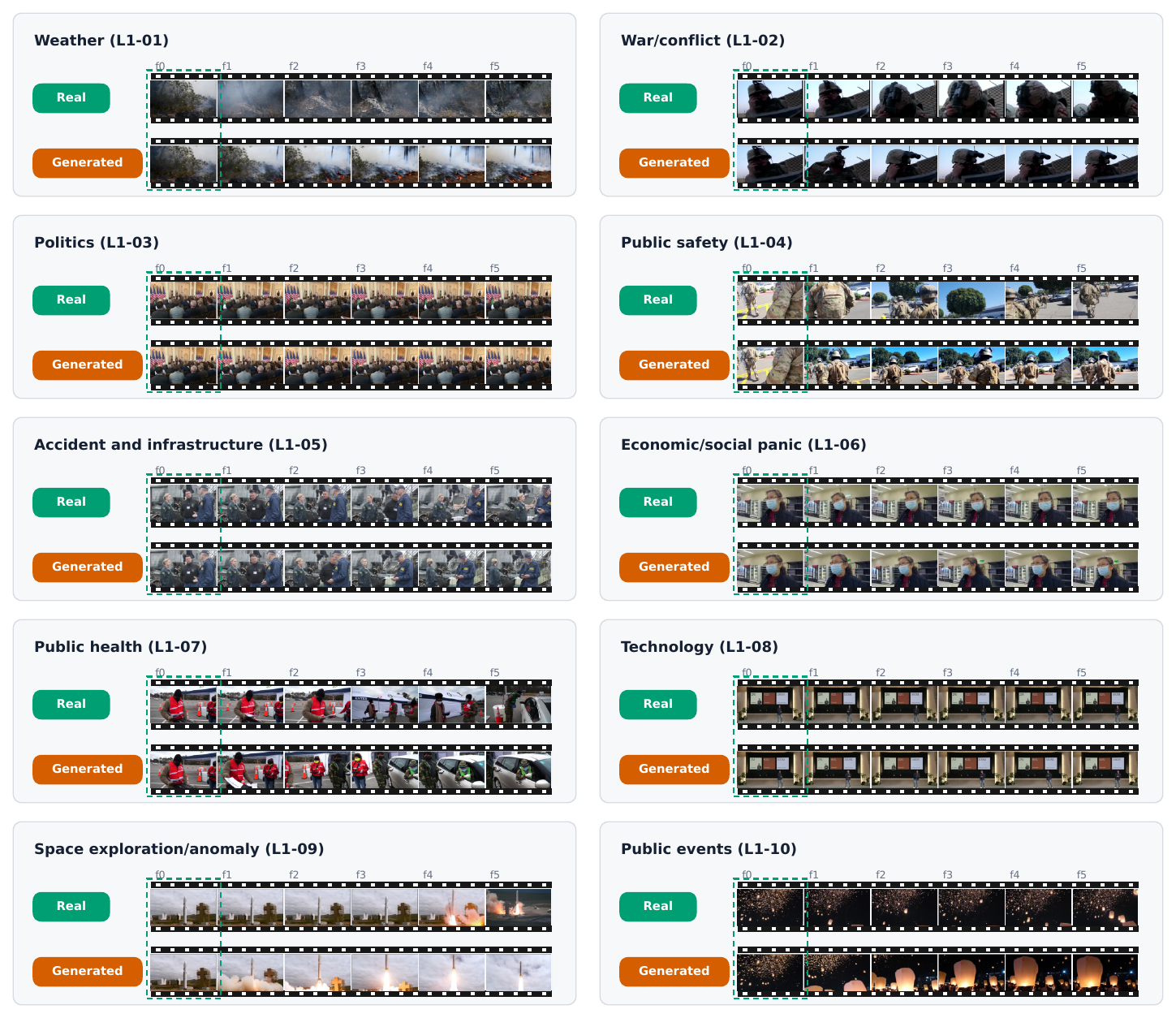}
\caption{\textbf{Qualitative coverage across all ten social-risk domains.}
Each panel pairs a real anchor (top) with one generated counterpart (bottom);
each continuation starts from the highlighted conditioning frame, retains the
event-specific scene appearance, and introduces new motion and content.}
\label{fig:gallery}
\end{figure}

\begin{figure}[H]
\centering
\includegraphics[width=\linewidth]{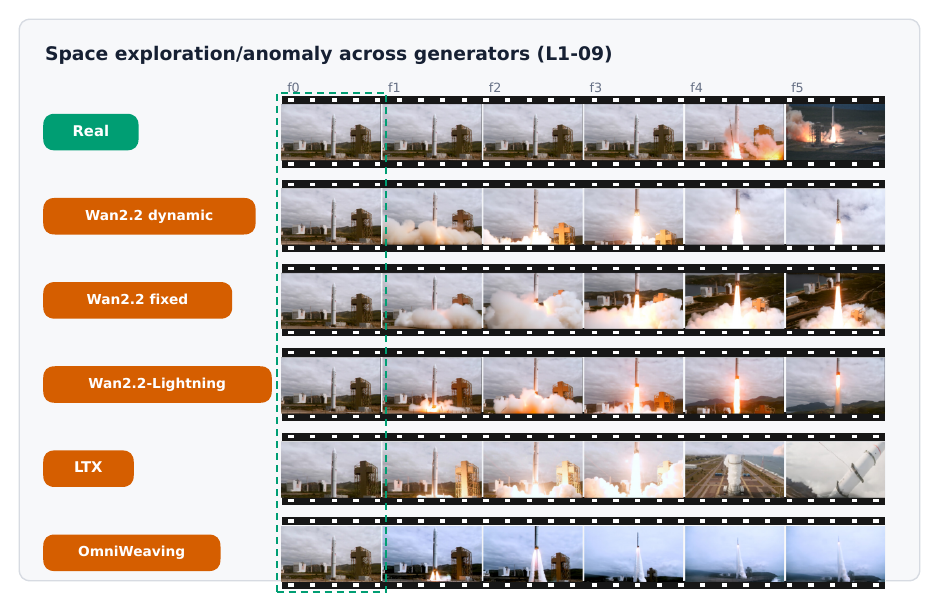}
\caption{\textbf{Cross-generator comparison under matched conditioning.}
The top row is a real space-launch anchor; subsequent rows show Wan2.2 dynamic,
its auxiliary fixed-duration control, Wan2.2-Lightning, LTX, and OmniWeaving
under the same prompt and first-frame reference. The outputs preserve the
initial launch-pad layout but differ in subsequent motion and visual drift.}
\label{fig:cross_generator}
\end{figure}

Figures~\ref{fig:gallery} and~\ref{fig:cross_generator} extend the paired
examples in Figure~\ref{fig:real_vs_gen_main} along category coverage and
generator variation. Figure~\ref{fig:gallery} covers all ten L1 domains using
one matched pair per domain, sampled from several benchmark generation sources.
Figure~\ref{fig:cross_generator} instead holds the anchor, prompt, and
first-frame reference fixed while varying the open-source generation
configuration. Each filmstrip contains six uniformly sampled frames. The dashed
outline marks the real first frame supplied to the generator; columns indicate
relative positions within each clip and are not timestamp-aligned across rows.

\FloatBarrier
\Needspace{6\baselineskip}
\appendixsection{Detector Evaluation Protocols and Extended Results}
\appendixsubsection{Detector Evaluation Protocol and Model Inputs}
\label{app:evaluation-implementation}

For each generation source, we evaluate every generated video together with its
matched real anchor. Each open-source generator contributes 1{,}830 such pairs;
for each closed-source generator, evaluation is restricted to the anchors for
which the provider returned a generated video. Benchmark-level results give
equal weight to the nine RA-Bench generation sources. The fixed-duration Wan2.2
control is marked with an asterisk wherever it is reported and is excluded from
these averages.

Unless an ablation states otherwise, we preserve each method's released
checkpoint or API version, visual preprocessing, temporal sampling policy, and
score-extraction rule. Continuous-score methods are evaluated with paired AUC
and TPR at 5\% FPR (T@5\%); discrete outputs are evaluated with BAcc, macro-F1,
and FakeR. For continuous outputs, the released score direction is fixed across
all sources; scores are not inverted post hoc when AUC falls below 50\%. We
do not apply source-specific inversion because it would require knowing the
generator identity and would not measure cross-source transfer. We additionally
report real recall (RealR) when analyzing class preference.
Zero-shot multimodal models receive sampled video frames only; no
filename, generator identity, seed, dataset label, or other auxiliary metadata
is provided. Table~\ref{tab:app_method_impl} records the visual input and output
used for each method family.

\paragraph{Zero-shot model identifiers and decoding.}
The local Qwen3.5 checkpoints are \texttt{Qwen3.5-0.8B},
\texttt{Qwen3.5-2B}, \texttt{Qwen3.5-4B}, \texttt{Qwen3.5-9B},
\texttt{Qwen3.5-27B}, \texttt{Qwen3.5-35B-A3B}, and
\texttt{Qwen3.5-122B-A10B}. Their inference disables thinking and uses greedy
decoding. API evaluations use the invoked identifiers \texttt{qwen3.7-plus}
(thinking mode), \texttt{gemini-3.1-pro-preview}, and \texttt{gpt-5.5}; Gemini
uses temperature 0. These API evaluations were conducted in June--July 2026.
Because the providers expose model aliases rather than immutable public
snapshots, we report the exact invoked identifiers and access period rather
than infer unrecorded version numbers.

\begin{table}[H]
\caption{\textbf{Detector inputs and evaluated outputs.}
Unless a control is explicitly noted, visual preprocessing and temporal sampling
follow the released implementations. The table reports the visual evidence
presented to each method and the output used to compute the common metrics;
prompt-specific fields and interventions are detailed in the corresponding
zero-shot and fine-tuned appendices.}
\label{tab:app_method_impl}
\centering
\footnotesize
\setlength{\tabcolsep}{4.2pt}
\renewcommand{\arraystretch}{1.08}
\begin{tabularx}{\linewidth}{@{}>{\raggedright\arraybackslash}p{3.05cm}
  >{\raggedright\arraybackslash}X >{\raggedright\arraybackslash}X@{}}
\toprule
\textbf{Method} & \textbf{Visual input / temporal sampling} &
\textbf{Evaluation output} \\
\midrule
\rowcolor{vfdStageA!9}
\multicolumn{3}{@{}l}{\textit{Traditional detectors}} \\
\makecell[l]{CNNSpot, NPR,\\UnivFD, ForgeLens}
& Eight uniformly spaced full-video frames; official ImageNet or CLIP
normalization
& Per-frame fake probabilities averaged into one clip score \\
DeCoF
& Eight frames within the first 32 decoded frames; center crop and
$224\!\times\!224$ resize
& Softmax fake probability from the eight-frame stack \\
D3
& One 3\,s window sampled at 8\,fps; 16 frames when available; 10\% center
crop and $224\!\times\!224$ resize
& Continuous \texttt{dis\_2nd\_std} score \\
ReStraV
& Center 2\,s window; 24 frames resized to $224\!\times\!224$; DINOv2
ViT-S/14 features
& Temporal-geometry MLP score,
\(\texttt{prob\_fake}=1-\texttt{prob\_real}\) \\
\midrule
\rowcolor{vfdStageB!9}
\multicolumn{3}{@{}l}{\textit{Zero-shot multimodal models}} \\
\makecell[l]{Qwen3.5 family\\Qwen3.7-Plus\\Gemini-3.1-Pro-\\Preview\\GPT-5.5}
& Sixteen temporally ordered frames, short side 256; no auxiliary metadata
& Binary verdict, 0--100 Rating with verdict, or five-aspect Diagnostic with
verdict \\
\midrule
\rowcolor{vfdExample!9}
\multicolumn{3}{@{}l}{\textit{MLLMs fine-tuned for AI-generated video detection}} \\
Skyra-SFT / Skyra-RL
& Sixteen frames at \(\operatorname{round}(i(N-1)/15)\), short side 256;
official timestamps or frame-index control
& Real/fake verdict parsed from \texttt{<answer>} \\
\BusterXpp~\citep{wen2025busterxpp}
& Released image-mode pipeline sampled at 2\,fps
& A/B multiple-choice answer parsed from \texttt{\textbackslash boxed\{\}} \\
\bottomrule
\end{tabularx}
\end{table}

The methods do not observe identical temporal evidence. Frame-level models
sample isolated frames across the clip, DeCoF emphasizes the opening frames,
D3 and ReStraV evaluate local windows, and the multimodal models receive broader
multi-frame summaries. These differences are part of the released inference
protocols, so the main comparison treats each detector as an end-to-end system.

\paragraph{Temporal-sampling control.}
The controlled experiment in Section~\ref{sec:traditional} changes only the
temporal allocation of a fixed eight-frame input. Uniform-8 selects eight frames
at uniformly spaced positions. For Global--Local-8, each clip is first scanned
at 4\,fps with a 160-pixel short side. We construct a camera-compensated eventness
curve from residual motion, structural change, and their temporal variation,
then select four consecutive frames centered on the strongest response and four
uniformly spaced frames. The coarse scan is used only for routing and is not
passed to the detector; detector
weights, detector-side preprocessing, and score aggregation remain unchanged.
The experiment covers eight fixed-length RA-Bench sources and the auxiliary
fixed-duration Wan2.2 control, yielding nine evaluated settings;
dynamic-duration Wan2.2 is
excluded. We report macro-average AUC over these settings and estimate the
five-detector mean confidence interval with a source-block bootstrap. This
ablation average is separate from the benchmark-level RA-Bench averages defined
above.

\FloatBarrier
\Needspace{0.45\textheight}
\appendixsubsection{Traditional-Detector Reference Transfer and Operating Points}
\label{app:traditional-extended}

\paragraph{Reference alignment and rank transfer.}
The public-reference column in Table~\ref{tab:traditional_main} provides
high-performance context rather than a matched-domain baseline. Six detectors
share the LTX-I2V evaluation reported by AIGVDBench
\citep{ma2026aigvdbench}; ReStraV instead uses its reported VidProM AUROC
\citep{interno2025restrav}. The source paper calls this metric AUROC; because it
is equivalent to ROC AUC, we denote it as AUC throughout. We therefore restrict
the rank-transfer analysis to
the six detectors evaluated on the same public reference. ReStraV remains in
the absolute-performance comparison but does not enter the correlation.

\begin{table}[H]
\caption{\textbf{Public detector rankings do not transfer to RA-Bench.}
Public AUC and rank use the shared AIGVDBench LTX-I2V reference. RA-Bench AUC
is the source-equal mean over the nine benchmark generation sources; the
fixed-duration Wan2.2 control is excluded. \(\Delta\)Rank is public rank minus
RA-Bench rank, so a positive value denotes an improved relative rank on
RA-Bench. AUC values are in \%.}
\label{tab:app_public_rank_transfer}
\centering
\footnotesize
\setlength{\tabcolsep}{6.0pt}
\renewcommand{\arraystretch}{1.04}
\begin{tabular}{lccccc}
\toprule
\textbf{Detector} & \textbf{Public AUC} & \textbf{Public rank} &
\textbf{RA-Bench AUC} & \textbf{RA-Bench rank} &
\(\boldsymbol{\Delta}\)\textbf{Rank} \\
\midrule
ForgeLens & 92.9 & 1 & 61.1 & 1 & 0 \\
UnivFD    & 89.9 & 2 & 40.7 & 6 & \textbf{$-4$} \\
DeCoF     & 81.5 & 3 & 60.9 & 2 & $+1$ \\
CNNSpot   & 81.4 & 4 & 44.4 & 4 & 0 \\
D3        & 77.7 & 5 & 43.5 & 5 & 0 \\
NPR       & 67.6 & 6 & 52.2 & 3 & \textbf{$+3$} \\
\bottomrule
\end{tabular}
\end{table}

The public and RA-Bench mean rankings have a Spearman correlation of only
\(0.26\).  UnivFD moves from second to sixth, whereas NPR moves from sixth to
third.  This is not a uniform performance loss: such a loss would reduce AUC
while largely preserving detector order.  The per-source correlations in
Table~\ref{tab:traditional_main} range from \(-0.37\) to \(0.54\), with a
median of \(0.31\); Seedance2.0 reverses the public ordering most strongly.
Together with the source-specific leaders reported in the main text, these
changes show that the preferred detector varies with the generation source.

\paragraph{Practical operating points.}

Table~\ref{tab:app_traditional_operating_points} complements the AUC and T@5\%
results in Table~\ref{tab:traditional_main} with stricter endpoints from the
same ROC sweep. T@1\% and T@5\% denote TPR at 1\% and 5\% FPR, respectively,
whereas F@95\% denotes FPR at 95\% TPR. These values do not use a detector's
default threshold. For random ranking, the corresponding values are 1\%, 5\%,
and 95\%. The auxiliary Wan2.2 fixed-duration control is excluded from all
Open averages, consistent with the main evaluation.

\begin{table}[H]
\caption{\textbf{Traditional detectors remain weak at both low-FPR and high-recall operating points.}
Cells are source-equal means over the four open-source and five closed-source
generators in RA-Bench, in \%. Higher is better for T@1\% and T@5\%;
lower is better for F@95\%. The Wan2.2 fixed-duration control is excluded.
Bold marks the best value within each source group.}
\label{tab:app_traditional_operating_points}
\centering
\footnotesize
\setlength{\tabcolsep}{5.2pt}
\renewcommand{\arraystretch}{1.03}
\begin{tabular}{l ccc ccc}
\toprule
& \multicolumn{3}{c}{\textbf{Open-source generators}}
& \multicolumn{3}{c}{\textbf{Closed-source generators}} \\
\cmidrule(lr){2-4}\cmidrule(lr){5-7}
\textbf{Detector}
& \textbf{T@1\%$\uparrow$} & \textbf{T@5\%$\uparrow$} & \textbf{F@95\%$\downarrow$}
& \textbf{T@1\%$\uparrow$} & \textbf{T@5\%$\uparrow$} & \textbf{F@95\%$\downarrow$} \\
\midrule
\rowcolor{vfdStageA!9}\multicolumn{7}{l}{\textit{Image / frame-level detectors}} \\
CNNSpot   & 0.5 & 5.1  & 91.1 & 0.2 & 3.4 & 92.4 \\
NPR       & 0.7 & 4.6  & 91.1 & 0.6 & 4.2 & 93.3 \\
UnivFD    & 0.2 & 1.0  & 95.8 & 0.2 & 1.1 & 96.6 \\
ForgeLens & \textbf{5.3} & \textbf{18.3} & 84.7
          & \textbf{2.5} & \textbf{9.7} & 85.3 \\
\midrule
\rowcolor{vfdExample!9}\multicolumn{7}{l}{\textit{Video / temporal-level detectors}} \\
DeCoF     & 0.6 & 3.9  & \textbf{73.2} & 0.4 & 2.0 & \textbf{77.6} \\
D3        & 0.6 & 4.5  & 93.0 & 1.2 & 5.3 & 98.4 \\
ReStraV   & 1.8 & 9.6  & 83.7 & 1.0 & 5.0 & 79.0 \\
\bottomrule
\end{tabular}
\end{table}

\begin{figure}[t]
    \centering
    \includegraphics[width=0.97\linewidth]{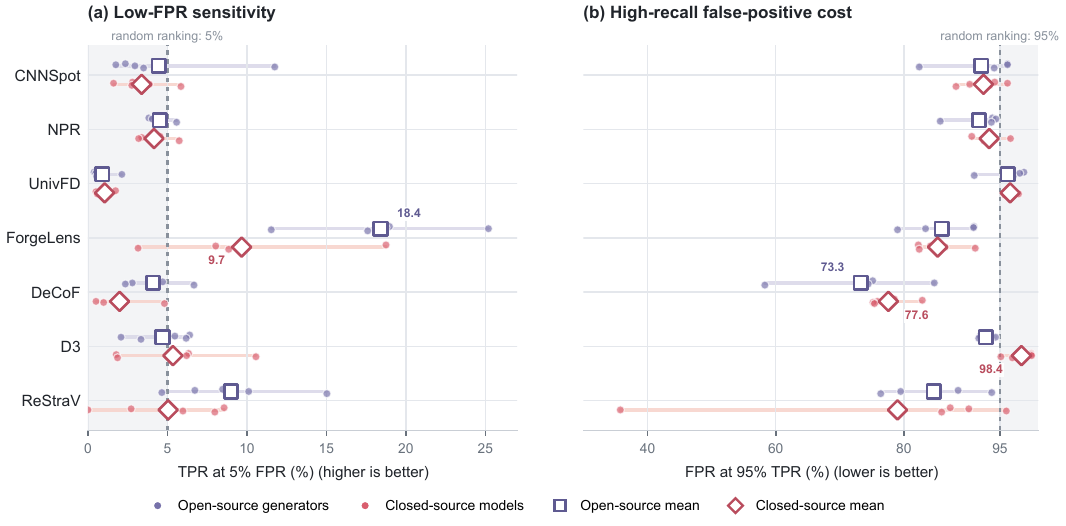}
    \caption{\textbf{Operating-point variability across generation sources.}
    Each filled point is one detector--source pair; horizontal segments span the
    four open-source or five closed-source generators in RA-Bench, and
    hollow markers denote source-equal means. The Wan2.2 fixed-duration control
    is excluded. (a) TPR at 5\% FPR. (b) FPR at 95\% TPR. Dashed lines mark
    random-ranking values, and shaded regions indicate worse-than-random
    operating points.}
    \label{fig:app_traditional_operating_points}
\end{figure}

The stricter endpoints show that the low AUCs in Table~\ref{tab:traditional_main}
translate into weak practical operating regions. Across the nine RA-Bench
sources, the seven-detector mean ranges from 0.4\% to 1.7\% at 1\% FPR and from
2.8\% to 7.5\% at 5\% FPR. Conversely, reaching 95\% TPR requires a mean FPR of
83.1--91.8\%. At the detector--source level, 54 of 63 pairs remain below 10\%
T@5\%, and 52 require at least 80\% FPR to reach 95\% TPR. Even LTX and
OmniWeaving, the two sources with the highest seven-detector mean AUC, reach
only 6.0\% and 7.5\% T@5\%, while their F@95\% remains 85.4\% and 85.5\%.
These results are obtained from complete ROC curves; they therefore cannot be
attributed to an unfavorable default threshold.

Similar AUC can also conceal different operating trade-offs. ForgeLens and
DeCoF are nearly tied in mean AUC on both the open-source generators
(64.2\% versus 63.4\%) and closed-source generators (58.7\% versus 58.9\%).
Across the open/closed groups, ForgeLens provides higher T@5\%
(18.3\%/9.7\%) but requires F@95\% of 84.7\%/85.3\%. DeCoF provides lower
T@5\% (3.9\%/2.0\%) but reduces F@95\% to 73.2\%/77.6\%.
Figure~\ref{fig:app_traditional_operating_points} further shows
that group means can hide source-specific instability: ReStraV's F@95\% ranges
from 35.8\% on Kling to 96.0\% on Seedance2.0. No evaluated detector therefore
maintains both low false-positive rates and high recall across generation
sources.

\FloatBarrier
\appendixsubsection{Zero-Shot Multimodal Model Protocol and Extended Analysis}
\label{app:zeroshot-full}
\label{app:zeroshot-prompts}

\paragraph{Unified evaluation protocol.}
All zero-shot multimodal models receive the same 16 temporally ordered frames
at a short-side resolution of 256 pixels. No filename, source, seed, dataset,
watermark, or other metadata is provided. We evaluate three prompt formats:
Binary requests a categorical verdict, Diagnostic requests five aspect scores
and a verdict, and Rating requests a continuous AI-generation score and a
verdict. The visual evidence and real-versus-generated decision task remain
fixed across the three formats.

All aggregate results give equal weight to generation sources rather than
individual clips. Open covers the four open-source RA-Bench generators, Closed
covers the five closed-source generators, and All covers all nine sources. The
fixed-duration Wan2.2 setting is an auxiliary temporal control and is excluded
from these averages.

\begin{table}[H]
\caption{\textbf{Output fields and metrics for the three zero-shot prompts.}
Binary and Diagnostic use the explicit \emph{Overall Verdict} for their primary classification metrics. Rating uses the continuous score for paired AUC; its verdict and the Diagnostic aspect scores are analyzed separately.}
\label{tab:app_zeroshot_metric_map}
\centering
\scriptsize
\setlength{\tabcolsep}{4.0pt}
\renewcommand{\arraystretch}{1.08}
\begin{tabularx}{\linewidth}{@{}l X p{3.05cm} p{3.65cm}@{}}
\toprule
\textbf{Prompt} & \textbf{Requested output} & \textbf{Primary metric} & \textbf{Additional analysis} \\
\midrule
Binary & Overall Verdict & BAcc / macro-F1 & FakeR and RealR \\
Rating & 0--100 rating and verdict & Paired rating AUC & Verdict macro-F1, FakeR, and RealR \\
Diagnostic & Five 0--100 aspect scores and verdict & Verdict BAcc / macro-F1 & Per-aspect paired AUC and exact-five collapse \\
\bottomrule
\end{tabularx}
\end{table}

\paragraph{Canonical prompt templates.}
The following templates define the three reported settings. The shared prefix fixes the visual input, prohibits metadata, and asks the model to judge the clip as a whole. Rating and Diagnostic scores measure the apparent degree of AI generation rather than confidence.

\tcbset{
  zs prompt/.style={
    enhanced,
    listing only,
    coltitle=vfdInk,
    fonttitle=\bfseries\footnotesize,
    boxrule=0.6pt,
    arc=1.5pt,
    left=5pt,
    right=5pt,
    top=2pt,
    bottom=2pt,
    before skip=6pt,
    after skip=7pt
  }
}
\newtcblisting{zsbinarypromptbox}[1]{
  zs prompt,
  colback=vfdStageA!4,
  colframe=vfdStageA!75!black,
  colbacktitle=vfdStageA!18,
  title={#1},
  listing options={
    basicstyle=\scriptsize\ttfamily,
    breaklines=true,
    breakindent=0pt,
    columns=fullflexible,
    keepspaces=true,
    showstringspaces=false
  }
}
\newtcblisting{zsratingpromptbox}[1]{
  zs prompt,
  colback=vfdStageB!4,
  colframe=vfdStageB!75!black,
  colbacktitle=vfdStageB!18,
  title={#1},
  listing options={
    basicstyle=\scriptsize\ttfamily,
    breaklines=true,
    breakindent=0pt,
    columns=fullflexible,
    keepspaces=true,
    showstringspaces=false
  }
}
\newtcblisting{zsdiagnosticpromptbox}[1]{
  zs prompt,
  colback=vfdExample!3,
  colframe=vfdExample!75!black,
  colbacktitle=vfdExample!17,
  title={#1},
  listing options={
    basicstyle=\scriptsize\ttfamily,
    breaklines=true,
    breakindent=0pt,
    columns=fullflexible,
    keepspaces=true,
    showstringspaces=false
  }
}

\Needspace{0.24\textheight}
\begin{zsbinarypromptbox}{Prompt 1: Binary verdict}
You are evaluating whether a video clip is AI-generated.
You will receive 16 frames sampled from the video clip in temporal order.
Base your judgment only on the visible content of these frames.
Do not use or assume any metadata, filename, source, seed, dataset, or watermark information.
Judge the clip as a whole.

Output exactly the following line and nothing else:
Overall Verdict: real/fake
\end{zsbinarypromptbox}

\Needspace{0.48\textheight}
\begin{zsratingpromptbox}{Prompt 2: AI-generation rating}
You are evaluating whether a video clip is AI-generated.
You will receive 16 frames sampled from the video clip in temporal order.
Base your judgment only on the visible content of these frames.
Do not use or assume any metadata, filename, source, seed, dataset, or watermark information.
Judge the clip as a whole.

AI-Generation Rating: an integer from 0 to 100, divided into five levels based on the degree to which the video appears to be AI-generated.
0-19: No visible signs of AI generation.
20-39: Low degree of AI generation, with only weak or isolated signs.
40-59: Moderate degree of AI generation, with mixed or inconclusive signs.
60-79: High degree of AI generation, with clear signs from one or more visual cues.
80-100: Very high degree of AI generation, with strong and consistent signs.
The rating reflects how AI-generated the content appears, not how confident you are.

Output exactly the following two lines and nothing else:
AI-Generation Rating: <integer 0-100>
Overall Verdict: real/fake
\end{zsratingpromptbox}

\Needspace{0.60\textheight}
\begin{zsdiagnosticpromptbox}{Prompt 3: Five-aspect diagnostic}
You are evaluating whether a video clip is AI-generated.
You will receive 16 frames sampled from the video clip in temporal order.
Base your judgment only on the visible content of these frames.
Do not use or assume any metadata, filename, source, seed, dataset, or watermark information.
Judge the clip as a whole.

Rate the clip on the five aspects below. For each aspect, give an integer from 0 to 100 for how strong the signs of AI generation are on that aspect, in five levels:
0-19: No visible signs of AI generation.
20-39: Only weak or isolated signs of AI generation.
40-59: Mixed or inconclusive signs of AI generation.
60-79: Clear signs of AI generation from one or more visual cues.
80-100: Strong and consistent signs of AI generation.
The rating reflects how AI-generated the content appears, not how confident you are.

The five aspects:
1. Texture & Material: surface textures and materials, such as oversmoothing, repetition, melting, or implausible details.
2. Structure & Local: object, body, face, hand, and text structure, such as warping, incorrect proportions, extra or missing parts, or splicing.
3. Lighting, Color & Optical: lighting, shadows, colors, reflections, and other optical effects.
4. Temporal: temporal appearance across the ordered frames, such as flicker, popping, duplication, frozen regions, or identity inconsistency.
5. Motion & Physical: motion, interactions, trajectories, and physical plausibility across the ordered frames.

Output exactly the following lines and nothing else:
Texture & Material: <integer 0-100>
Structure & Local: <integer 0-100>
Lighting, Color & Optical: <integer 0-100>
Temporal: <integer 0-100>
Motion & Physical: <integer 0-100>
Overall Verdict: real/fake
\end{zsdiagnosticpromptbox}

\paragraph{Aggregate performance.}
Table~\ref{tab:app_zeroshot_aggregate} extends
Table~\ref{tab:zeroshot_main} to every evaluated zero-shot model. Aggregate
performance remains close to chance for many model--prompt pairs, and no prompt
improves consistently with model scale. Qwen3.5-122B-A10B improves from
53.0/45.7 BAcc/macro-F1 under Binary to 54.8/50.8 under Diagnostic, whereas
Qwen3.5-27B declines from 53.2/46.7 to 51.3/37.2. The two Qwen3.5
mixture-of-experts models perform better under Rating, while the smaller dense
models remain near chance. Qwen3.7-Plus and Gemini-3.1-Pro-Preview also vary
between source groups: their Rating AUC decreases from 62.6 to 55.4 and from
66.6 to 61.2, respectively, between open- and closed-source generators.

\Needspace{0.72\textheight}
\noindent
\begin{minipage}[t]{0.485\linewidth}
\vspace{0pt}
\captionsetup{type=table,font=footnotesize,skip=4pt}
\captionof{table}{\textbf{Source-equal zero-shot results.}
Binary and Diagnostic report BAcc/macro-F1; Rating reports paired AUC/verdict macro-F1 (top/bottom). The Open and Closed columns average four and five sources, respectively, and All averages all nine. The fixed-duration Wan2.2 control does not enter these averages. Values are in \%.}
\label{tab:app_zeroshot_aggregate}
\centering
\fontsize{6.2}{6.7}\selectfont
\setlength{\tabcolsep}{1.2pt}
\renewcommand{\arraystretch}{0.90}
\begin{tabular*}{\linewidth}{@{\extracolsep{\fill}}llccc@{}}
\toprule
\textbf{Model} & \textbf{Prompt} & \textbf{Open} & \textbf{Closed} & \textbf{All} \\
\midrule
\rowcolor{vfdStageA!9}
\multicolumn{5}{l}{\textit{Qwen3.5 dense models}} \\
Qwen3.5-0.8B & Binary & \makecell{55.2\\[-0.45mm]\textcolor{black!62}{49.5}} & \makecell{53.6\\[-0.45mm]\textcolor{black!62}{46.9}} & \makecell{54.3\\[-0.45mm]\textcolor{black!62}{48.1}} \\
 & Diagnostic & \makecell{49.8\\[-0.45mm]\textcolor{black!62}{34.7}} & \makecell{49.9\\[-0.45mm]\textcolor{black!62}{34.8}} & \makecell{49.8\\[-0.45mm]\textcolor{black!62}{34.8}} \\
 & Rating & \makecell{50.4\\[-0.45mm]\textcolor{black!62}{33.3}} & \makecell{52.3\\[-0.45mm]\textcolor{black!62}{33.3}} & \makecell{51.5\\[-0.45mm]\textcolor{black!62}{33.3}} \\
\addlinespace[1pt]
Qwen3.5-2B & Binary & \makecell{50.1\\[-0.45mm]\textcolor{black!62}{33.6}} & \makecell{50.3\\[-0.45mm]\textcolor{black!62}{33.9}} & \makecell{50.2\\[-0.45mm]\textcolor{black!62}{33.8}} \\
 & Diagnostic & \makecell{51.4\\[-0.45mm]\textcolor{black!62}{43.8}} & \makecell{50.3\\[-0.45mm]\textcolor{black!62}{42.5}} & \makecell{50.8\\[-0.45mm]\textcolor{black!62}{43.0}} \\
 & Rating & \makecell{52.8\\[-0.45mm]\textcolor{black!62}{48.8}} & \makecell{51.9\\[-0.45mm]\textcolor{black!62}{48.5}} & \makecell{52.3\\[-0.45mm]\textcolor{black!62}{48.7}} \\
\addlinespace[1pt]
Qwen3.5-4B & Binary & \makecell{53.7\\[-0.45mm]\textcolor{black!62}{47.5}} & \makecell{51.0\\[-0.45mm]\textcolor{black!62}{45.4}} & \makecell{52.2\\[-0.45mm]\textcolor{black!62}{46.3}} \\
 & Diagnostic & \makecell{50.8\\[-0.45mm]\textcolor{black!62}{38.8}} & \makecell{50.0\\[-0.45mm]\textcolor{black!62}{37.1}} & \makecell{50.3\\[-0.45mm]\textcolor{black!62}{37.8}} \\
 & Rating & \makecell{52.4\\[-0.45mm]\textcolor{black!62}{47.5}} & \makecell{50.6\\[-0.45mm]\textcolor{black!62}{45.9}} & \makecell{51.4\\[-0.45mm]\textcolor{black!62}{46.6}} \\
\addlinespace[1pt]
Qwen3.5-9B & Binary & \makecell{54.1\\[-0.45mm]\textcolor{black!62}{50.2}} & \makecell{52.4\\[-0.45mm]\textcolor{black!62}{48.8}} & \makecell{53.2\\[-0.45mm]\textcolor{black!62}{49.4}} \\
 & Diagnostic & \makecell{50.2\\[-0.45mm]\textcolor{black!62}{33.9}} & \makecell{50.0\\[-0.45mm]\textcolor{black!62}{33.6}} & \makecell{50.1\\[-0.45mm]\textcolor{black!62}{33.7}} \\
 & Rating & \makecell{49.9\\[-0.45mm]\textcolor{black!62}{34.8}} & \makecell{48.9\\[-0.45mm]\textcolor{black!62}{34.5}} & \makecell{49.4\\[-0.45mm]\textcolor{black!62}{34.6}} \\
\addlinespace[1pt]
Qwen3.5-27B & Binary & \makecell{55.3\\[-0.45mm]\textcolor{black!62}{49.8}} & \makecell{51.6\\[-0.45mm]\textcolor{black!62}{44.2}} & \makecell{53.2\\[-0.45mm]\textcolor{black!62}{46.7}} \\
 & Diagnostic & \makecell{51.9\\[-0.45mm]\textcolor{black!62}{38.6}} & \makecell{50.7\\[-0.45mm]\textcolor{black!62}{36.1}} & \makecell{51.3\\[-0.45mm]\textcolor{black!62}{37.2}} \\
 & Rating & \makecell{52.3\\[-0.45mm]\textcolor{black!62}{35.5}} & \makecell{50.3\\[-0.45mm]\textcolor{black!62}{34.6}} & \makecell{51.2\\[-0.45mm]\textcolor{black!62}{35.0}} \\
\midrule
\rowcolor{vfdExample!9}
\multicolumn{5}{l}{\textit{Qwen3.5 mixture-of-experts models}} \\
\makecell[l]{Qwen3.5-\\[-0.5mm]35B-A3B} & Binary & \makecell{55.8\\[-0.45mm]\textcolor{black!62}{48.0}} & \makecell{52.4\\[-0.45mm]\textcolor{black!62}{42.4}} & \makecell{53.9\\[-0.45mm]\textcolor{black!62}{44.9}} \\
 & Diagnostic & \makecell{52.8\\[-0.45mm]\textcolor{black!62}{40.8}} & \makecell{50.7\\[-0.45mm]\textcolor{black!62}{36.6}} & \makecell{51.6\\[-0.45mm]\textcolor{black!62}{38.5}} \\
 & Rating & \makecell{60.3\\[-0.45mm]\textcolor{black!62}{59.1}} & \makecell{53.4\\[-0.45mm]\textcolor{black!62}{52.6}} & \makecell{56.5\\[-0.45mm]\textcolor{black!62}{55.5}} \\
\addlinespace[1pt]
\makecell[l]{Qwen3.5-\\[-0.5mm]122B-A10B} & Binary & \makecell{54.2\\[-0.45mm]\textcolor{black!62}{47.5}} & \makecell{52.0\\[-0.45mm]\textcolor{black!62}{44.2}} & \makecell{53.0\\[-0.45mm]\textcolor{black!62}{45.7}} \\
 & Diagnostic & \makecell{58.9\\[-0.45mm]\textcolor{black!62}{56.3}} & \makecell{51.5\\[-0.45mm]\textcolor{black!62}{46.4}} & \makecell{54.8\\[-0.45mm]\textcolor{black!62}{50.8}} \\
 & Rating & \makecell{59.9\\[-0.45mm]\textcolor{black!62}{59.2}} & \makecell{52.5\\[-0.45mm]\textcolor{black!62}{52.7}} & \makecell{55.8\\[-0.45mm]\textcolor{black!62}{55.6}} \\
\midrule
\rowcolor{vfdStageB!9}
\multicolumn{5}{l}{\textit{Frontier multimodal models}} \\
\makecell[l]{Qwen3.7-Plus\\[-0.5mm](thinking)} & Binary & \makecell{57.2\\[-0.45mm]\textcolor{black!62}{51.0}} & \makecell{53.3\\[-0.45mm]\textcolor{black!62}{45.2}} & \makecell{55.0\\[-0.45mm]\textcolor{black!62}{47.8}} \\
 & Diagnostic & \makecell{57.9\\[-0.45mm]\textcolor{black!62}{53.0}} & \makecell{53.3\\[-0.45mm]\textcolor{black!62}{46.2}} & \makecell{55.3\\[-0.45mm]\textcolor{black!62}{49.2}} \\
 & Rating & \makecell{62.6\\[-0.45mm]\textcolor{black!62}{53.4}} & \makecell{55.4\\[-0.45mm]\textcolor{black!62}{47.3}} & \makecell{58.6\\[-0.45mm]\textcolor{black!62}{50.0}} \\
\addlinespace[1pt]
\makecell[l]{Gemini-3.1-\\[-0.5mm]Pro-Preview} & Binary & \makecell{66.5\\[-0.45mm]\textcolor{black!62}{66.3}} & \makecell{60.9\\[-0.45mm]\textcolor{black!62}{60.2}} & \makecell{63.4\\[-0.45mm]\textcolor{black!62}{62.9}} \\
 & Diagnostic & \makecell{66.3\\[-0.45mm]\textcolor{black!62}{65.9}} & \makecell{60.6\\[-0.45mm]\textcolor{black!62}{59.7}} & \makecell{63.1\\[-0.45mm]\textcolor{black!62}{62.5}} \\
 & Rating & \makecell{66.6\\[-0.45mm]\textcolor{black!62}{66.7}} & \makecell{61.2\\[-0.45mm]\textcolor{black!62}{60.0}} & \makecell{63.6\\[-0.45mm]\textcolor{black!62}{63.0}} \\
\addlinespace[1pt]
GPT-5.5 & Binary & \makecell{52.5\\[-0.45mm]\textcolor{black!62}{39.1}} & \makecell{51.2\\[-0.45mm]\textcolor{black!62}{36.6}} & \makecell{51.8\\[-0.45mm]\textcolor{black!62}{37.7}} \\
 & Diagnostic & \makecell{51.8\\[-0.45mm]\textcolor{black!62}{37.4}} & \makecell{50.8\\[-0.45mm]\textcolor{black!62}{35.4}} & \makecell{51.2\\[-0.45mm]\textcolor{black!62}{36.3}} \\
 & Rating & \makecell{64.3\\[-0.45mm]\textcolor{black!62}{38.0}} & \makecell{60.3\\[-0.45mm]\textcolor{black!62}{36.0}} & \makecell{62.1\\[-0.45mm]\textcolor{black!62}{36.9}} \\
\bottomrule
\end{tabular*}
\end{minipage}\hfill
\begin{minipage}[t]{0.49\linewidth}
\vspace{0pt}
\centering
\includegraphics[width=\linewidth]{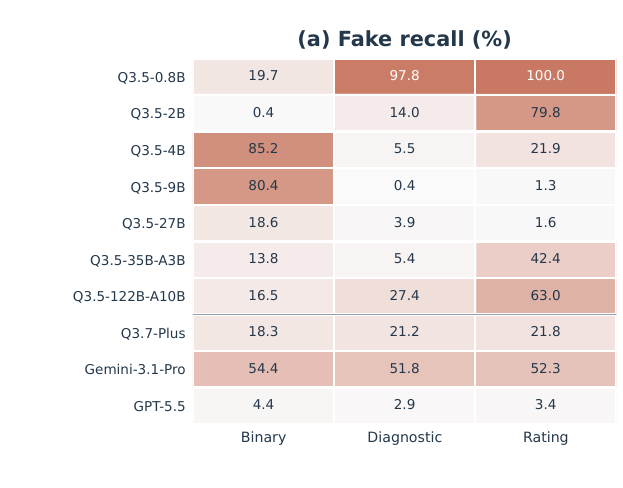}\\[-1.5mm]
\includegraphics[width=\linewidth]{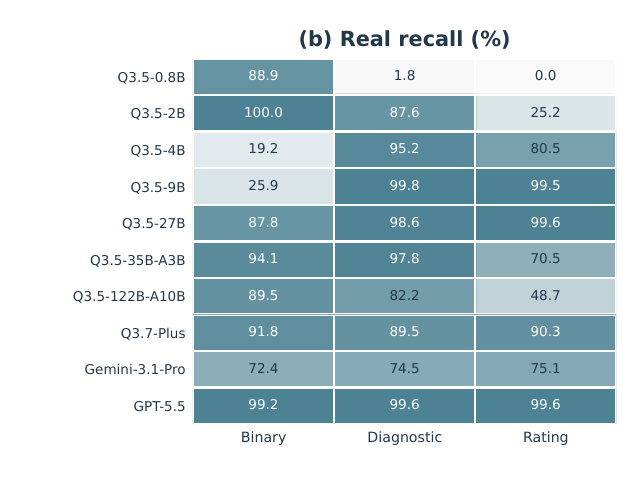}
\captionsetup{type=figure,font=footnotesize,skip=4pt}
\captionof{figure}{\textbf{Prompt-dependent class recall.}
Each heatmap reports a source-equal mean over the nine RA-Bench generators; Rating uses its explicit Overall Verdict. Several Qwen3.5 models reverse their class preference across prompts. Gemini remains comparatively balanced, whereas GPT-5.5 consistently predicts \emph{Real}.}
\label{fig:app_zeroshot_prompt_bias}
\end{minipage}
\par\medskip

\paragraph{Prompt-dependent operating points.}
Figure~\ref{fig:app_zeroshot_prompt_bias} shows why BAcc alone is insufficient for comparing the three prompts. Qwen3.5-0.8B changes from a Real-favoring Binary rule to predicting nearly every video as fake under Diagnostic and Rating. Qwen3.5-4B and Qwen3.5-9B move in the opposite direction: Binary favors fake, whereas the structured prompts favor real. These models can all remain near 50\% BAcc because errors on one class offset correct decisions on the other, despite representing qualitatively different operating points.

Gemini-3.1-Pro-Preview changes little across prompt formats: its FakeR remains between 51.8\% and 54.4\%, while RealR remains between 72.4\% and 75.1\%. Qwen3.7-Plus is also stable but remains tilted toward real. GPT-5.5 shows why prompt stability alone is insufficient: its FakeR remains below 4.5\% under every prompt while RealR exceeds 99\%. Reporting FakeR and RealR is therefore necessary to distinguish prompt-invariant detection behavior from a persistent class preference.

\begin{figure}[t]
    \centering
    \includegraphics[width=0.94\linewidth]{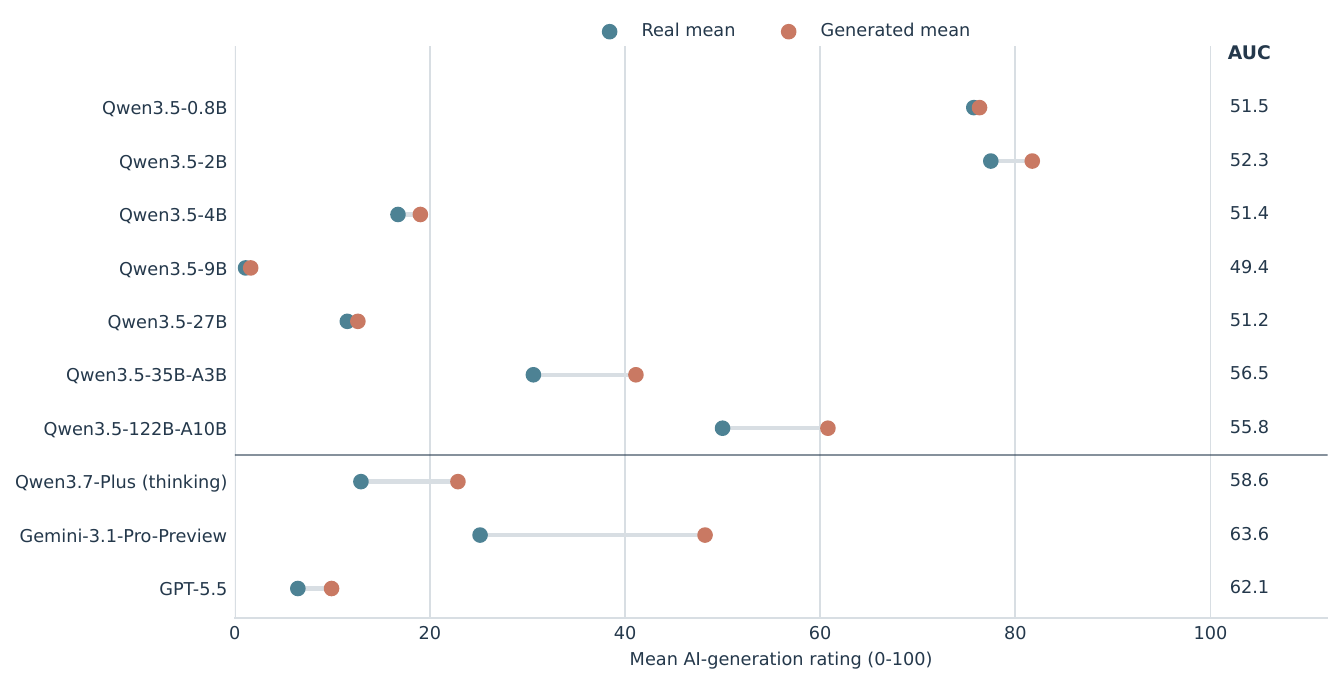}
    \caption{\textbf{Model-specific use of the 0--100 Rating scale.}
    Gray-blue and coral markers show source-equal means for matched real and generated videos; the right column reports paired AUC. Large differences in score location across models do not imply corresponding differences in separability.}
    \label{fig:app_zeroshot_rating}
\end{figure}

\Needspace{12\baselineskip}
\paragraph{Absolute Rating scores are not directly comparable across models.}
Figure~\ref{fig:app_zeroshot_rating} shows that models use the 0--100 scale very differently. Qwen3.5-2B assigns mean scores of 81.7 to generated videos and 77.5 to real videos, whereas Qwen3.5-9B assigns 1.6 and 1.1. Both remain close to chance AUC. A high absolute rating therefore does not imply stronger separation, and scores from different models should not be compared as generation probabilities.

Continuous ranking and categorical decisions can also diverge within one model. GPT-5.5 obtains 62.1\% paired AUC despite assigning low scores to both classes, but its explicit verdict yields only 36.9\% macro-F1 because it predicts almost every video as real. Gemini-3.1-Pro-Preview is the only evaluated model with both paired AUC and verdict macro-F1 above 60\%. We therefore report paired AUC for the continuous score and evaluate the explicit verdict separately rather than deriving an additional prediction with a post hoc threshold.

\begin{figure}[t]
    \centering
    \includegraphics[width=0.98\linewidth]{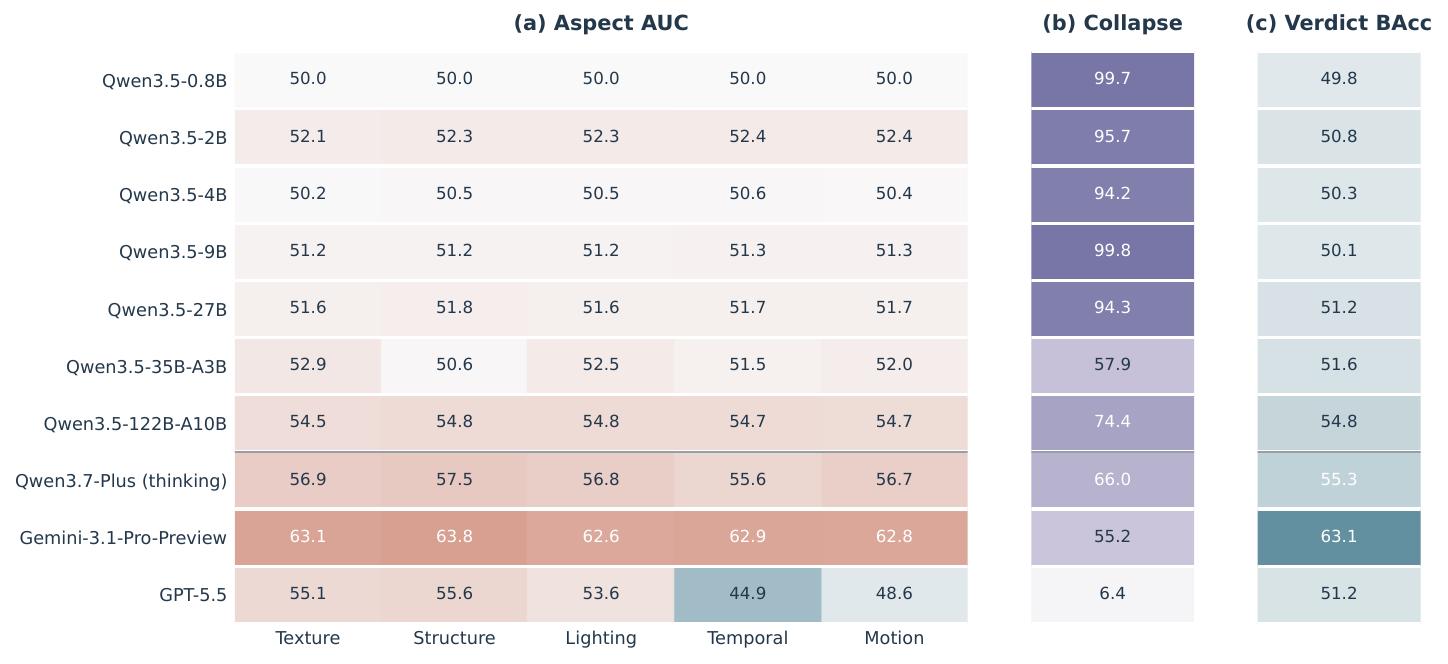}
    \caption{\textbf{Separability and collapse of the five Diagnostic scores.}
    Panel (a) reports source-equal paired AUC for each requested dimension. Panel (b) reports the fraction of complete outputs that assign exactly the same value to all five dimensions. Panel (c) reports BAcc from the explicit Overall Verdict.}
    \label{fig:app_zeroshot_diagnostic}
\end{figure}

\paragraph{The Diagnostic prompt often collapses to one global score.}
Figure~\ref{fig:app_zeroshot_diagnostic} evaluates each Diagnostic dimension as a continuous signal. For Qwen3.5-0.8B, 4B, 9B, and 27B, aspect AUC remains near chance while exact-five collapse exceeds 94\%. These models usually repeat one global score across all requested dimensions rather than provide aspect-specific judgments.

Qwen3.7-Plus reaches 55.6--57.5\% aspect AUC, and Qwen3.5-122B-A10B reaches 54.5--54.8\%, but both still collapse a majority of complete outputs. Gemini-3.1-Pro-Preview reaches 62.6--63.8\% aspect AUC, yet assigns the same value to all five dimensions in 55.2\% of complete outputs. GPT-5.5 rarely collapses the five values, but its Temporal and Motion scores reach only 44.9\% and 48.6\% AUC. A useful diagnostic therefore requires both distinct outputs and dimension-specific separation.

\Needspace{7\baselineskip}
These analyses separate three behaviors that aggregate BAcc cannot distinguish. FakeR and RealR reveal prompt-induced class preference, Rating AUC measures continuous ranking, and Diagnostic AUC together with collapse tests whether structured responses contain aspect-specific information. We therefore report the class-conditional and structured-output analyses alongside Table~\ref{tab:zeroshot_main} rather than treating one summary score as sufficient evidence of zero-shot detection ability.

\FloatBarrier
\appendixsubsection{Fine-Tuned MLLM Protocol Sensitivity and Class-Conditional Behavior}
\label{app:finetuned-full}

We evaluate Skyra under two temporal-label settings. The official prompt prefixes each of the 16 sampled frames with its absolute timestamp. The frame-index prompt replaces these prefixes with \texttt{[Frame=01]} through \texttt{[Frame=16]}, while preserving the images and their order. Both settings cover the nine RA-Bench generation sources and the auxiliary Wan2.2 fixed-duration control. Because the visual input is unchanged, the comparison measures sensitivity to the temporal-label representation.

\providecommand{\genlogo}[1]{\raisebox{-0.25em}{\includegraphics[height=1.15em]{figures/logos/#1}}}
\providecommand{\genrow}[2]{\makecell[l]{\genlogo{#1}\hspace{0.35em}#2}}

\begin{table*}[t]
\caption{\textbf{Temporal-label controls for Skyra.}
FakeR and BAcc are source-equal results over the nine RA-Bench generation sources, whereas Full RealR is measured on all 1{,}830 real anchors; Wan2.2 fixed$^{*}$ is excluded from the benchmark averages. $\Delta_{5}$ is the source-equal fake-recall difference between exact-5-second and other clips over the eight RA-Bench generation sources containing both strata. Values are in \%, and $\Delta_{5}$ is in percentage points.}
\label{tab:app_skyra_summary}
\centering
\scriptsize
\renewcommand{\arraystretch}{1.04}
\setlength{\tabcolsep}{2.4pt}
\begin{tabular}{llcccc}
\toprule
\textbf{Model} & \textbf{Temporal label} & \textbf{Full RealR} & \textbf{FakeR} & \textbf{BAcc} & $\boldsymbol{\Delta}_{\mathbf{5}}$ \\
\midrule
Skyra-SFT & official timestamp & 87.4 & 49.6 & 68.5 & $+48.6$ \\
          & frame index & 55.4 & 53.6 & 54.4 & $+1.5$ \\
\midrule
Skyra-RL  & official timestamp & 84.0 & 55.1 & 69.5 & $+43.4$ \\
          & frame index & 49.5 & 60.6 & 54.9 & $+1.4$ \\
\bottomrule
\end{tabular}
\end{table*}

Table~\ref{tab:app_skyra_summary} shows that replacing timestamps with frame indices changes the class preference rather than reducing recall for both classes. Skyra-SFT gains 4.0 points in source-equal FakeR but loses 32.0 points in full-set RealR; Skyra-RL gains 5.5 points in FakeR but loses 34.5 points in full-set RealR. Both BAcc values consequently fall to about 55\%.

Our audit of the released ViF-CoT-4K and ViF-Bench metadata~\citep{li2025skyra} finds a label-correlated temporal grid. In ViF-CoT-4K, the final timestamp is exactly 5.00 seconds for 81 of 2{,}017 real samples (4.0\%) and 1{,}074 of 2{,}017 generated samples (53.2\%). In ViF-Bench, the corresponding proportions are 97 of 165 real samples (58.8\%) and 2{,}994 of 2{,}997 generated samples (99.9\%). Because these timestamps are exposed directly in the prompt, the final temporal tag can act as a label prior.

\begin{figure*}[t]
    \centering
    \includegraphics[width=0.96\textwidth]{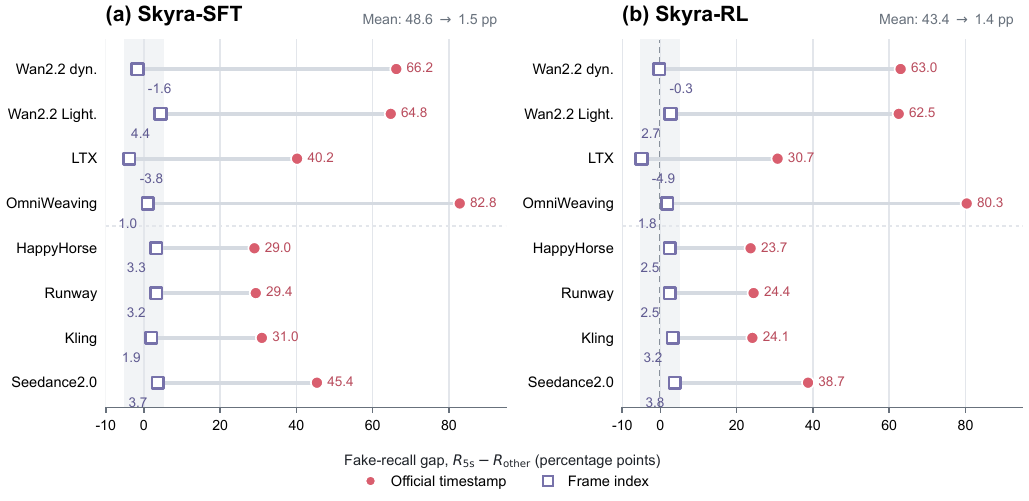}
    \caption{\textbf{Exact-5-second fake-recall gap under the Skyra prompt ablation.} For each of the eight RA-Bench generation sources containing both strata, points show $\Delta_{5}=R_{5\mathrm{s}}-R_{\mathrm{other}}$. Replacing timestamps with frame indices reduces the source-equal mean gap from 48.6 to 1.5 points for Skyra-SFT and from 43.4 to 1.4 points for Skyra-RL. Both prompts receive the same sampled frames. The auxiliary Wan2.2 fixed$^{*}$ control and Hailuo are excluded because each contains only one stratum.}
    \label{fig:app_skyra_temporal_gap}
\end{figure*}

Table~\ref{tab:app_skyra_grid} and Figure~\ref{fig:app_skyra_temporal_gap} show the corresponding behavior on RA-Bench. Under the official-timestamp prompt, every eligible source has higher fake recall for exact-5-second clips than for other clips. The source-equal gap ranges from 29.0 to 82.8 points for Skyra-SFT and from 23.7 to 80.3 points for Skyra-RL. Replacing timestamps with frame indices reduces the mean gap from 48.6 to 1.5 points and from 43.4 to 1.4 points, respectively.

\begin{table*}[t]
\caption{\textbf{Fake recall by final temporal tag.} Results cover the nine benchmark sources and the auxiliary Wan2.2 fixed-duration control. \emph{5}
denotes clips whose final official timestamp token is exactly
$[\mathrm{T}{=}5.00\,\mathrm{s}]$; \emph{other} denotes all remaining clips. Wan2.2 fixed$^{*}$ contains
only the 5-second stratum, whereas Hailuo contains no exact-5-second clip. Replacing timestamps with
frame indices removes the large 5-versus-other gap on all eight RA-Bench generation sources containing both strata.}
\label{tab:app_skyra_grid}
\centering
\scriptsize
\setlength{\tabcolsep}{3.5pt}
\renewcommand{\arraystretch}{1.05}
\begin{tabular}{lcc cc cc cc cc}
\toprule
 & \multicolumn{2}{c}{\textbf{Count}} & \multicolumn{4}{c}{\textbf{Skyra-SFT}}
 & \multicolumn{4}{c}{\textbf{Skyra-RL}} \\
\cmidrule(lr){2-3}\cmidrule(lr){4-7}\cmidrule(lr){8-11}
 & & & \multicolumn{2}{c}{\textbf{Timestamp}} & \multicolumn{2}{c}{\textbf{Frame index}}
 & \multicolumn{2}{c}{\textbf{Timestamp}} & \multicolumn{2}{c}{\textbf{Frame index}} \\
\textbf{Source} & \textbf{5} & \textbf{other}
 & \textbf{5} & \textbf{other} & \textbf{5} & \textbf{other}
 & \textbf{5} & \textbf{other} & \textbf{5} & \textbf{other} \\
\midrule
\rowcolor{vfdStageA!9}\multicolumn{11}{l}{\textit{Open-source generators}} \\
\genrow{wan.png}{Wan2.2 dynamic} & 74 & 1{,}756 & 97.3 & 31.1 & 63.5 & 65.1 & 98.6 & 35.6 & 70.3 & 70.6 \\
\rowcolor{orange!7}\genrow{wan.png}{Wan2.2 fixed$^{*}$} & 1{,}830 & 0 & 96.4 & -- & 62.8 & -- & 97.7 & -- & 68.4 & -- \\
\genrow{wan_lightning.png}{Wan2.2-Lightning} & 74 & 1{,}756 & 94.6 & 29.8 & 66.2 & 61.8 & 97.3 & 34.8 & 70.3 & 67.6 \\
\genrow{ltx.png}{LTX} & 96 & 1{,}734 & 100.0 & 59.8 & 3.1 & 6.9 & 99.0 & 68.2 & 5.2 & 10.1 \\
\genrow{omni.png}{OmniWeaving} & 74 & 1{,}756 & 95.9 & 13.1 & 47.3 & 46.3 & 97.3 & 17.0 & 58.1 & 56.3 \\
\midrule
\rowcolor{vfdStageB!9}\multicolumn{11}{l}{\textit{Closed-source generators}} \\
\genrow{happyhorse.png}{HappyHorse} & 302 & 1{,}485 & 97.4 & 68.4 & 66.6 & 63.3 & 98.3 & 74.7 & 72.5 & 70.0 \\
\genrow{runway.png}{Runway} & 300 & 1{,}505 & 97.7 & 68.3 & 69.3 & 66.1 & 98.7 & 74.2 & 75.3 & 72.8 \\
\genrow{kling.png}{Kling} & 298 & 1{,}492 & 98.7 & 67.7 & 64.1 & 62.2 & 99.0 & 74.9 & 74.2 & 70.9 \\
\genrow{seedance.png}{Seedance2.0} & 269 & 1{,}255 & 97.0 & 51.6 & 55.4 & 51.7 & 97.8 & 59.0 & 64.7 & 60.9 \\
\genrow{hailuo.png}{Hailuo} & 0 & 1{,}830 & -- & 22.5 & -- & 56.6 & -- & 28.3 & -- & 64.6 \\
\bottomrule
\end{tabular}
\end{table*}

The two single-stratum sources provide complementary evidence. Wan2.2 fixed contains only exact-5-second clips; frame indices reduce FakeR from 96.4\% to 62.8\% for Skyra-SFT and from 97.7\% to 68.4\% for Skyra-RL. Hailuo has no exact-5-second clips, and FakeR instead rises from 22.5\% to 56.6\% and from 28.3\% to 64.6\%. These opposite shifts are consistent with a temporal prior that favors a fake verdict at 5 seconds and a real verdict otherwise.

Switching to frame indices does not eliminate source dependence. Under frame indices, source-wise BAcc still ranges from 31.1\% to 61.0\% for Skyra-SFT and from 29.7\% to 61.3\% for Skyra-RL. The source patterns of SFT and RL remain strongly aligned under both timestamps (Spearman $\rho=0.93$) and frame indices ($\rho=0.95$). In particular, both variants retain very low frame-index FakeR on LTX. RL therefore shifts the operating point but does not remove the shared source-specific failure pattern.

\BusterXpp\ exhibits a different class-conditional failure. Across the nine benchmark sources and Wan2.2 fixed$^{*}$, its released pipeline attains only 4.1--9.1\% FakeR while RealR remains at 93.0--93.8\%. It predicts \emph{Real} for approximately 93.6\% of the paired inputs, largely independent of generation source. This behavior explains why its macro-F1 remains low despite high RealR in Table~\ref{tab:finetuned_main}. The Skyra and \BusterXpp\ results therefore show why balanced and class-conditional metrics are necessary when evaluating fine-tuned MLLMs.

\FloatBarrier
\Needspace{0.45\textheight}
\appendixsection{Generation Factors and Robustness Analysis}
\appendixsubsection{Generation Quality and Detectability}
\label{app:generation-quality}

\paragraph{VBench++ I2V protocol.}
We run the released VBench++ I2V evaluators on all 16{,}056 generated clips. The evaluation covers Subject Consistency, Background Consistency, Motion Smoothness, Dynamic Degree, Aesthetic Quality, and Imaging Quality, together with Video--Image Subject Consistency and Video--Image Background Consistency. The official VBench-I2V Quality Score combines the first six dimensions after normalization with the released ranges, using weight 0.5 for Dynamic Degree and weight 1 for each remaining dimension. We define Condition Fidelity as the mean of the two normalized Video--Image consistency scores and Combined Quality as the equal-weight mean of the VBench-I2V Quality Score and Condition Fidelity. Camera Motion is excluded because RA-Bench prompts do not provide the controlled camera-motion labels required by that evaluator. Condition Fidelity and Combined Quality are therefore benchmark-specific aggregates rather than the official VBench-I2V I2V Score and Total Score.

All eight dimension scores are produced by the released evaluators on RA-Bench inputs. Because RA-Bench uses real-event conditioning images and prompts rather than the VBench++ Image Suite, these results characterize the present benchmark and are not intended for direct comparison with the official leaderboard. Source-level Subject and Background Consistency retain the released frame-level aggregation. The association analysis instead uses one score vector per clip so that video duration does not determine statistical weight. All eight dimensions are complete for all 16{,}056 clips.

\paragraph{Association analysis.}
For each continuous score, we compute percentile ranks within every generation source and Dynamic Degree group. We fit a model with fixed effects for each source--group combination, give every source equal total weight, and report the estimated change associated with an interquartile increase. This specification retains the continuous scores rather than dividing clips into discrete quality groups. Gemini Binary and Diagnostic outputs are represented by fake indicators, while Rating is scaled to $[0,1]$. Traditional-detector fake scores are converted to empirical percentiles within each detector and source before averaging across the seven detectors, preventing detector-specific score scales from dominating the mean. The valid sample counts are 15{,}347 for Binary, 15{,}890 for Diagnostic, 15{,}451 for Rating, and 16{,}052 for the traditional-detector mean.

Dynamic Degree is binary in the released evaluator and is analyzed separately through a dynamic-minus-static contrast with generation-source and real-anchor fixed effects. Confidence intervals use cluster-robust standard errors at the real-anchor level to account for multiple generated clips derived from the same anchor. All estimates are interpreted as adjusted associations within RA-Bench rather than causal effects of quality on detector behavior.

\begin{table}[H]
\caption{\textbf{Source-level VBench++ profile on RA-Bench.}
(a) Aggregate scores and conditioning consistency. The VBench-I2V Quality Score follows the released six-dimension normalization and weighting. Condition Fidelity averages normalized Video--Image Subject and Background Consistency, and Combined Quality gives the two aggregates equal weight. I2V-S and I2V-B are the corresponding raw evaluator outputs. (b) Raw means for the six dimensions of the VBench-I2V Quality Score. Source-level Subject and Background Consistency follow the released frame-level aggregation. All values except $N$ are percentages. These are custom-input results on RA-Bench rather than VBench++ Image Suite leaderboard scores.}
\label{tab:app-generation-quality-source}
\label{tab:app-generation-quality-dimensions}
\centering
\scriptsize
\renewcommand{\arraystretch}{1.04}

\textbf{(a) Aggregate scores and conditioning consistency}\par\smallskip
\setlength{\tabcolsep}{4.5pt}
\begin{tabular*}{\textwidth}{@{\extracolsep{\fill}}lcccccc@{}}
\toprule
\textbf{Generation source} & \textbf{$N$} & \shortstack{\textbf{Quality}\\\textbf{Score}} & \shortstack{\textbf{Condition}\\\textbf{Fidelity}} & \shortstack{\textbf{Combined}\\\textbf{Quality}} & \textbf{I2V-S} & \textbf{I2V-B} \\
\midrule
\rowcolor{vfdStageA!9}\multicolumn{7}{l}{\textit{Open-source models}} \\
Wan2.2 dynamic    & 1{,}830 & 76.2 & 94.0 & 85.1 & 94.4 & 95.9 \\
Wan2.2-Lightning  & 1{,}830 & 76.4 & 96.4 & 86.4 & 96.8 & 97.4 \\
LTX                & 1{,}830 & 74.9 & 92.5 & 83.7 & 93.5 & 94.5 \\
OmniWeaving        & 1{,}830 & 72.9 & 95.3 & 84.1 & 95.8 & 96.6 \\
\midrule
\rowcolor{vfdStageB!9}\multicolumn{7}{l}{\textit{Closed-source models}} \\
HappyHorse         & 1{,}787 & 76.9 & 92.6 & 84.8 & 93.6 & 94.6 \\
Runway             & 1{,}805 & 78.0 & 95.0 & 86.5 & 95.5 & 96.5 \\
Kling              & 1{,}790 & 76.7 & 93.1 & 84.9 & 93.9 & 95.0 \\
Seedance2.0        & 1{,}524 & 76.4 & 91.1 & 83.7 & 92.3 & 93.5 \\
Hailuo             & 1{,}830 & 77.0 & 96.3 & 86.7 & 96.7 & 97.4 \\
\bottomrule
\end{tabular*}

\medskip
\textbf{(b) Six VBench-I2V Quality Score dimensions}\par\smallskip
\begin{tabular*}{\textwidth}{@{\extracolsep{\fill}}lcccccc@{}}
\toprule
\textbf{Generation source} & \textbf{Subject} & \textbf{Background} & \textbf{Motion} & \textbf{Dynamic} & \textbf{Aesthetic} & \textbf{Imaging} \\
\midrule
\rowcolor{vfdStageA!9}\multicolumn{7}{l}{\textit{Open-source models}} \\
Wan2.2 dynamic    & 88.1 & 92.0 & 98.0 & 69.4 & 50.6 & 66.1 \\
Wan2.2-Lightning  & 89.9 & 92.1 & 98.2 & 63.7 & 50.0 & 67.4 \\
LTX                & 86.0 & 89.9 & 98.7 & 74.4 & 47.4 & 62.8 \\
OmniWeaving        & 87.5 & 91.0 & 99.0 & 52.8 & 47.9 & 58.1 \\
\midrule
\rowcolor{vfdStageB!9}\multicolumn{7}{l}{\textit{Closed-source models}} \\
HappyHorse         & 88.4 & 91.8 & 98.7 & 66.7 & 50.2 & 69.4 \\
Runway             & 88.9 & 91.3 & 98.8 & 76.8 & 50.5 & 69.5 \\
Kling              & 90.3 & 93.4 & 99.0 & 63.4 & 49.3 & 65.1 \\
Seedance2.0        & 88.5 & 92.2 & 98.6 & 63.6 & 49.6 & 67.9 \\
Hailuo             & 89.6 & 93.0 & 99.0 & 63.3 & 50.0 & 66.3 \\
\bottomrule
\end{tabular*}
\end{table}

The source-level profile is descriptive rather than the basis of the main analysis. The VBench-I2V Quality Score spans 72.9--78.0, Condition Fidelity spans 91.1--96.4, and Combined Quality spans 83.7--86.7. These narrow ranges do not imply similar detection difficulty. LTX and Seedance2.0, for example, both obtain a Combined Quality score of 83.7, while their Gemini Diagnostic fake recalls are 74.5\% and 32.3\%, respectively. We therefore rely on within-source clip-level associations rather than source-level quality rankings.

\begin{table}[H]
\caption{\textbf{Adjusted associations between VBench++ scores and fake-side detector outputs.}
Except for Dynamic Degree, each cell reports the estimated change associated with an interquartile increase within each source and Dynamic Degree group, followed by its 95\% confidence interval. Negative values indicate weaker fake evidence at higher scores. Dynamic Degree reports the adjusted dynamic-minus-static contrast. Gemini Binary and Diagnostic values are percentage points of fake recall, Gemini Rating is expressed on a 0--100 scale, and traditional values are percentile points of the mean normalized fake score across seven detectors.}
\label{tab:app-generation-quality-effects}
\centering
\scriptsize
\setlength{\tabcolsep}{3.6pt}
\renewcommand{\arraystretch}{1.06}
\resizebox{\textwidth}{!}{%
\begin{tabular}{l cccc}
\toprule
\textbf{Quality score or dimension} & \textbf{Gemini Binary} & \textbf{Gemini Diagnostic} & \textbf{Gemini Rating} & \textbf{7-detector mean} \\
\midrule
\rowcolor{vfdStageA!8}\multicolumn{5}{l}{\textit{Aggregate scores}} \\
VBench-I2V Quality Score & $-6.1\,[-8.2,-3.9]$ & $-9.6\,[-11.7,-7.5]$ & $-6.4\,[-8.1,-4.6]$ & $-11.5\,[-12.3,-10.8]$ \\
Condition Fidelity       & $-10.6\,[-12.5,-8.6]$ & $-14.4\,[-16.3,-12.5]$ & $-10.2\,[-11.8,-8.5]$ & $-9.5\,[-10.1,-8.8]$ \\
Combined Quality         & $-8.2\,[-10.3,-6.1]$ & $-12.4\,[-14.4,-10.4]$ & $-8.3\,[-10.0,-6.6]$ & $-10.9\,[-11.6,-10.2]$ \\
\midrule
\rowcolor{vfdStageB!8}\multicolumn{5}{l}{\textit{Condition fidelity and temporal consistency}} \\
Video--Image subject consistency    & $-9.6\,[-11.6,-7.7]$ & $-13.2\,[-15.1,-11.3]$ & $-9.4\,[-11.0,-7.7]$ & $-9.3\,[-10.0,-8.7]$ \\
Video--Image background consistency & $-10.9\,[-12.8,-8.9]$ & $-14.9\,[-16.7,-13.0]$ & $-10.3\,[-11.9,-8.7]$ & $-9.0\,[-9.7,-8.3]$ \\
Subject consistency                  & $-11.4\,[-13.4,-9.3]$ & $-14.8\,[-16.7,-12.9]$ & $-10.7\,[-12.4,-9.0]$ & $-10.1\,[-10.8,-9.4]$ \\
Background consistency               & $-9.4\,[-11.4,-7.4]$ & $-13.6\,[-15.5,-11.7]$ & $-9.3\,[-10.9,-7.6]$ & $-7.6\,[-8.3,-6.9]$ \\
Motion smoothness                    & $-7.8\,[-9.7,-5.9]$ & $-7.9\,[-9.7,-6.0]$ & $-7.0\,[-8.6,-5.5]$ & $-2.1\,[-2.9,-1.4]$ \\
\midrule
\rowcolor{vfdStageA!8}\multicolumn{5}{l}{\textit{Frame-level quality}} \\
Aesthetic quality & $1.4\,[-0.7,3.6]$ & $-0.4\,[-2.6,1.7]$ & $-0.8\,[-2.6,1.0]$ & $-4.1\,[-4.9,-3.2]$ \\
Imaging quality   & $1.7\,[-0.4,3.9]$ & $0.6\,[-1.5,2.7]$ & $1.8\,[0.1,3.6]$ & $-9.4\,[-10.2,-8.7]$ \\
\midrule
\rowcolor{vfdStageB!8}\multicolumn{5}{l}{\textit{Motion amount}} \\
Dynamic Degree    & $6.5\,[4.0,9.0]$ & $8.4\,[5.8,10.9]$ & $5.5\,[3.3,7.7]$ & $0.1\,[-0.4,0.6]$ \\
\bottomrule
\end{tabular}%
}
\end{table}

Table~\ref{tab:app-generation-quality-effects} shows three distinct patterns. Condition Fidelity has the largest aggregate association with all three Gemini outputs, whereas the VBench-I2V Quality Score has the largest association with the traditional-detector mean. The four subject and background consistency measures are negative across every detector output, while Aesthetic Quality and Imaging Quality show no clear association with Gemini Diagnostic. Dynamic Degree changes in the opposite direction for Gemini but is nearly null for the traditional-detector mean. A single aggregate quality score therefore combines dimensions with different detector-specific associations.

\FloatBarrier
\Needspace{6\baselineskip}
\appendixsubsection{Generation Settings and Detectability}
\label{app:conditioning}

\paragraph{Generation protocol.}
We evaluate the same 1{,}830 anchor-derived prompts under T2V, first-frame I2V, and first+last-frame I2V generation. T2V uses no real-image condition, while the two I2V settings use the matched first frame or the matched first and last frames, respectively; the first-frame setting is the protocol used by RA-Bench. We use the same anchor-derived prompts and seeds across all three settings.

We generate all three settings for seeds 0, 42, and 123. The main comparison uses seed 0 and evaluates the seven traditional detectors, both official-timestamp and frame-index variants of Skyra-SFT and Skyra-RL, and the released \BusterXpp\ pipeline. Traditional-detector AUC and T@5\% use the same frozen score vectors for the 1{,}830 paired real anchors. For the fine-tuned MLLMs, the real-video control is fixed within each configuration, and we report BAcc, FakeR, and macro-F1. Unparseable \BusterXpp\ responses are retained as abstentions and counted as incorrect. Table~\ref{tab:app_conditioning_finetuned} gives the complete seed-0 classification results.

\begin{table}[H]
\caption{\textbf{Complete seed-0 fine-tuned MLLM results across generation settings.}
The real-video control is fixed within each configuration. RealR is listed in
official-timestamp/frame-index order for Skyra. All values are percentages;
$\Delta$ is first+last minus first frame.
\BusterXpp\ abstentions are counted as incorrect.}
\label{tab:app_conditioning_finetuned}
\centering
\scriptsize
\setlength{\tabcolsep}{4.0pt}
\renewcommand{\arraystretch}{0.98}
\begin{tabular*}{\textwidth}{@{\extracolsep{\fill}}llcccc@{}}
\toprule
\textbf{Configuration} & \textbf{Metric}
& \textbf{T2V}
& \makecell{\textbf{First frame}\\[-0.2ex]\textbf{(I2V)}}
& \makecell{\textbf{First+last frames}\\[-0.2ex]\textbf{(I2V)}}
& \makecell{\textbf{End-frame effect}\\[-0.2ex]$\boldsymbol{\Delta}$} \\
\midrule
\rowcolor{blue!6}\multicolumn{6}{l}{\textbf{Skyra-SFT} \; (RealR: 87.4 / 55.4)} \\
\multirow{3}{*}{\makecell[l]{Official\\timestamp}}
  & BAcc     & 75.3 & 60.6 & 51.9 & $-8.7$ \\
  & FakeR    & 63.2 & 33.8 & 16.4 & $-17.4$ \\
  & macro-F1 & 74.9 & 57.5 & 45.0 & $-12.6$ \\
\multirow{3}{*}{\makecell[l]{Frame\\index}}
  & BAcc     & 75.2 & 60.2 & 50.6 & $-9.6$ \\
  & FakeR    & 94.9 & 65.1 & 45.8 & $-19.3$ \\
  & macro-F1 & 74.1 & 60.1 & 50.5 & $-9.7$ \\
\midrule
\rowcolor{cyan!6}\multicolumn{6}{l}{\textbf{Skyra-RL} \; (RealR: 84.0 / 49.5)} \\
\multirow{3}{*}{\makecell[l]{Official\\timestamp}}
  & BAcc     & 77.2 & 61.1 & 52.3 & $-8.7$ \\
  & FakeR    & 70.4 & 38.2 & 20.7 & $-17.5$ \\
  & macro-F1 & 77.1 & 58.9 & 47.0 & $-11.9$ \\
\multirow{3}{*}{\makecell[l]{Frame\\index}}
  & BAcc     & 73.3 & 60.0 & 52.2 & $-7.8$ \\
  & FakeR    & 97.2 & 70.5 & 54.9 & $-15.6$ \\
  & macro-F1 & 71.7 & 59.6 & 52.2 & $-7.4$ \\
\midrule
\rowcolor{orange!7}\multicolumn{6}{l}{\textbf{\BusterXpp} \; (RealR: 93.7)} \\
\multirow{3}{*}{\makecell[l]{Released\\pipeline}}
  & BAcc     & 60.4 & 49.8 & 48.9 & $-1.0$ \\
  & FakeR    & 27.0 & 6.0 & 4.0 & $-2.0$ \\
  & macro-F1 & 55.4 & 37.9 & 36.0 & $-1.9$ \\
\bottomrule
\end{tabular*}
\end{table}

Because the real-video control is fixed within each configuration, differences in BAcc across generation settings are driven by FakeR. Both Skyra prompt variants follow the same direction when the matched last frame is added: FakeR decreases by 17.4 and 19.3 points for Skyra-SFT and by 17.5 and 15.6 points for Skyra-RL. \BusterXpp\ decreases by only 2.0 points because its FakeR is already 6.0\% under first-frame I2V.

\begin{figure*}[!t]
    \centering
    \includegraphics[width=0.92\textwidth]{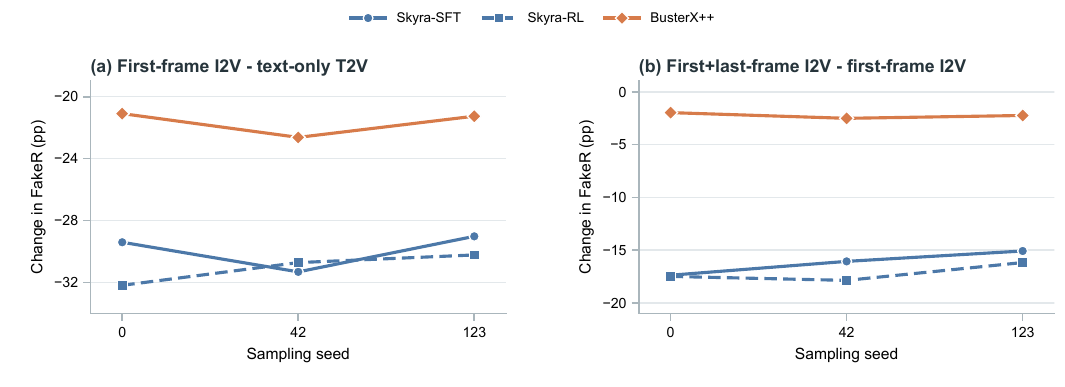}
    \caption{\textbf{Cross-seed stability of the generation-setting effects.}
    Panel (a) reports first-frame-I2V-minus-T2V changes in FakeR, and panel (b) reports first+last-frame-I2V-minus-first-frame-I2V changes. Skyra-SFT and Skyra-RL share the same color and are distinguished by solid and dashed lines, while \BusterXpp\ is shown in orange. Negative values indicate lower FakeR in the second setting. The panels use different vertical ranges to show cross-seed variation at the scale of each contrast.}
    \label{fig:app_conditioning_seed_stability}
\end{figure*}

For the cross-seed analysis, Table~\ref{tab:app_conditioning_seed_stability} and Figure~\ref{fig:app_conditioning_seed_stability} report the released official-timestamp prompts for Skyra-SFT and Skyra-RL together with \BusterXpp. This matches the protocol used for the seed analysis in Section~\ref{sec:seed} while retaining one released setting for each fine-tuned MLLM detector.

\begin{table*}[!t]
\caption{\textbf{Fine-tuned MLLM FakeR across generation settings and seeds.}
Each setting contains the same 1{,}830 anchors for seeds 0, 42, and 123.
Skyra uses the official-timestamp prompt, matching the paper's cross-seed
protocol. End-frame effect is first+last minus first frame;
Seed range is the maximum minus the minimum across the three seeds. All values
are percentages. \BusterXpp\ abstentions are counted as incorrect.}
\label{tab:app_conditioning_seed_stability}
\centering
\scriptsize
\setlength{\tabcolsep}{5.0pt}
\renewcommand{\arraystretch}{1.00}
\begin{tabular*}{\textwidth}{@{\extracolsep{\fill}}llcccc@{}}
\toprule
\textbf{Configuration} & \textbf{Generation setting}
& \textbf{Seed 0} & \textbf{Seed 42} & \textbf{Seed 123}
& \textbf{Seed range} \\
\midrule
\rowcolor{blue!6}\multicolumn{6}{l}{\textbf{Skyra-SFT}} \\
Official timestamp & T2V               & 63.2 & 64.2 & 60.4 & 3.7 \\
                   & First frame       & 33.8 & 32.8 & 31.4 & 2.4 \\
                   & First+last frames & 16.4 & 16.8 & 16.3 & 0.4 \\
                   & \textit{End-frame effect} & $-17.4$ & $-16.1$ & $-15.1$ & 2.3 \\
\midrule
\rowcolor{cyan!6}\multicolumn{6}{l}{\textbf{Skyra-RL}} \\
Official timestamp & T2V               & 70.4 & 70.2 & 67.9 & 2.5 \\
                   & First frame       & 38.2 & 39.5 & 37.7 & 1.8 \\
                   & First+last frames & 20.7 & 21.6 & 21.5 & 0.9 \\
                   & \textit{End-frame effect} & $-17.5$ & $-17.9$ & $-16.2$ & 1.7 \\
\midrule
\rowcolor{orange!7}\multicolumn{6}{l}{\textbf{\BusterXpp}} \\
Released pipeline  & T2V               & 27.0 & 29.2 & 27.6 & 2.2 \\
                   & First frame       & 6.0 & 6.6 & 6.3 & 0.7 \\
                   & First+last frames & 4.0 & 4.1 & 4.1 & 0.1 \\
                   & \textit{End-frame effect} & $-2.0$ & $-2.5$ & $-2.2$ & 0.5 \\
\bottomrule
\end{tabular*}
\end{table*}

The direction of each change is unchanged across the three seeds. For first+last-frame versus first-frame I2V, the FakeR decrease ranges from 15.1 to 17.4 points for Skyra-SFT, from 16.2 to 17.9 points for Skyra-RL, and from 2.0 to 2.5 points for \BusterXpp. The small cross-seed ranges show that the seed-0 results in the main paper do not arise from a single sampled realization.

\FloatBarrier
\Needspace{6\baselineskip}
\appendixsubsection{Stability Across Generation Seeds}
\label{app:seed-detail}

\paragraph{Controlled comparison.}
The primary RA-Bench clips from the four open-source generators use seed 0. We generate two additional sets with seeds 42 and 123, each covering the same 1{,}830 real-video anchors per generator. For every source--anchor pair, the prompt, conditioning image, and all generation settings other than the seed are unchanged. The paired real-video set is also shared across the three evaluations. Closed-source providers are omitted because their APIs do not expose a controllable seed, and the fixed-duration Wan2.2 control is omitted because this analysis concerns the four open-source RA-Bench generators.

\paragraph{Traditional detectors.}
The source-level seven-detector AUC means in Table~\ref{tab:seed_stability_main}(a) vary by at most 1.05 points across seeds. The seed-0 source means differ from their three-seed averages by at most 0.20 AUC points. Table~\ref{tab:app_seed_traditional} reports all 28 detector--source comparisons and shows where the remaining variation occurs.

The largest detector-specific range is 5.74 AUC points for ReStraV on LTX. The 28-cell detector--source pattern nevertheless remains highly similar across seeds. Pairwise Spearman correlations over all 28 AUC values are \(0.981\), \(0.978\), and \(0.989\) for seed pairs \(0/42\), \(0/123\), and \(42/123\), respectively. Only NPR on Wan2.2 dynamic, D3 on Wan2.2-Lightning, and UnivFD on LTX cross 50\% AUC, and each remains close to random ranking under all three seeds.

\paragraph{Fine-tuned MLLMs.}
The seed changes only the generated-video side of each paired evaluation, so RealR is constant for a given model. The resulting BAcc range is therefore half of the FakeR range. Table~\ref{tab:app_seed_mllm} reports the complete source-wise comparison.

The largest FakeR ranges are 2.35 points for Skyra-SFT, 2.46 points for Skyra-RL, and 0.71 points for \BusterXpp; the corresponding BAcc ranges are 1.17, 1.23, and 0.36 points. Skyra retains the same pronounced source differences under all three seeds. \BusterXpp\ remains below 10\% FakeR for every source--seed combination. The source-level conclusions are therefore not driven by the seed-0 generations used in the primary benchmark.

\begin{table}[H]
\caption{\textbf{Detector-specific traditional results across generation seeds.}
Each cell reports AUC. \(\Delta\) is the maximum minus the minimum across seeds. All values are percentages; bold marks the largest range in the table.}
\label{tab:app_seed_traditional}
\centering
\scriptsize
\setlength{\tabcolsep}{5.0pt}
\renewcommand{\arraystretch}{0.98}
\begin{tabular*}{\textwidth}{@{\extracolsep{\fill}}lcccc@{}}
\toprule
\textbf{Detector}
& \textbf{Seed 0}
& \textbf{Seed 42}
& \textbf{Seed 123}
& \(\boldsymbol{\Delta}\) \\
\midrule
\rowcolor{vfdStageA!9}\multicolumn{5}{l}{\textit{Wan2.2 dynamic}} \\
CNNSpot    & 35.50 & 34.78 & 35.38 & 0.72 \\
NPR        & 50.64 & 49.79 & 50.38 & 0.85 \\
UnivFD     & 32.72 & 33.23 & 33.84 & 1.11 \\
ForgeLens  & 59.99 & 60.23 & 59.89 & 0.34 \\
DeCoF      & 62.47 & 63.18 & 63.29 & 0.82 \\
D3         & 55.36 & 53.70 & 54.93 & 1.67 \\
ReStraV    & 59.28 & 57.84 & 57.32 & 1.96 \\
\addlinespace[1pt]
\rowcolor{vfdStageA!9}\multicolumn{5}{l}{\textit{Wan2.2-Lightning}} \\
CNNSpot    & 43.78 & 44.36 & 43.54 & 0.82 \\
NPR        & 55.17 & 54.89 & 55.03 & 0.29 \\
UnivFD     & 42.96 & 44.28 & 43.76 & 1.32 \\
ForgeLens  & 69.61 & 70.04 & 69.85 & 0.43 \\
DeCoF      & 56.70 & 56.95 & 57.05 & 0.36 \\
D3         & 49.09 & 50.05 & 49.65 & 0.97 \\
ReStraV    & 62.41 & 60.86 & 62.40 & 1.55 \\
\addlinespace[1pt]
\rowcolor{vfdStageA!9}\multicolumn{5}{l}{\textit{LTX}} \\
CNNSpot    & 63.99 & 62.02 & 60.29 & 3.70 \\
NPR        & 58.65 & 59.74 & 58.62 & 1.11 \\
UnivFD     & 51.16 & 50.03 & 49.54 & 1.62 \\
ForgeLens  & 64.11 & 62.07 & 61.87 & 2.24 \\
DeCoF      & 62.47 & 63.42 & 61.07 & 2.35 \\
D3         & 58.01 & 58.90 & 57.77 & 1.13 \\
ReStraV    & 42.45 & 48.19 & 47.87 & \textbf{5.74} \\
\addlinespace[1pt]
\rowcolor{vfdStageA!9}\multicolumn{5}{l}{\textit{OmniWeaving}} \\
CNNSpot    & 45.76 & 45.99 & 45.03 & 0.96 \\
NPR        & 50.93 & 51.94 & 51.69 & 1.01 \\
UnivFD     & 39.88 & 39.94 & 39.94 & 0.06 \\
ForgeLens  & 62.95 & 62.83 & 63.34 & 0.51 \\
DeCoF      & 72.11 & 72.26 & 71.48 & 0.77 \\
D3         & 52.85 & 52.91 & 52.70 & 0.22 \\
ReStraV    & 68.34 & 70.23 & 68.64 & 1.89 \\
\bottomrule
\end{tabular*}
\end{table}

\begin{table}[H]
\caption{\textbf{Source-wise seed sensitivity of the fine-tuned MLLMs.}
Skyra-SFT and Skyra-RL use the official-timestamp prompt, and \BusterXpp\ uses its released evaluation pipeline. FakeR and BAcc are reported in \%; \(\Delta\) denotes the maximum minus the minimum across seeds. Bold marks the largest range for each metric.}
\label{tab:app_seed_mllm}
\centering
\scriptsize
\setlength{\tabcolsep}{5.0pt}
\renewcommand{\arraystretch}{1.02}
\begin{tabular*}{\textwidth}{@{\extracolsep{\fill}}lccccc@{}}
\toprule
\textbf{Generation source}
& \multicolumn{3}{c}{\textbf{FakeR}}
& \makecell{\(\boldsymbol{\Delta}\)\\[-0.2ex]\textbf{FakeR}}
& \makecell{\(\boldsymbol{\Delta}\)\\[-0.2ex]\textbf{BAcc}} \\
\cmidrule(lr){2-4}
& \textbf{Seed 0} & \textbf{Seed 42} & \textbf{Seed 123} & & \\
\midrule
\rowcolor{orange!7}\multicolumn{6}{l}{\textit{Skyra-SFT}} \\
Wan2.2 dynamic    & 33.77 & 32.84 & 31.42 & 2.35 & 1.17 \\
Wan2.2-Lightning  & 32.46 & 30.78 & 30.11 & 2.35 & 1.17 \\
LTX               & 61.91 & 62.90 & 63.01 & 1.09 & 0.55 \\
OmniWeaving       & 16.45 & 17.32 & 16.07 & 1.26 & 0.63 \\
\addlinespace[1pt]
\rowcolor{orange!7}\multicolumn{6}{l}{\textit{Skyra-RL}} \\
Wan2.2 dynamic    & 38.20 & 39.45 & 37.65 & 1.80 & 0.90 \\
Wan2.2-Lightning  & 37.32 & 36.50 & 34.86 & \textbf{2.46} & \textbf{1.23} \\
LTX               & 69.84 & 70.44 & 70.60 & 0.77 & 0.38 \\
OmniWeaving       & 20.22 & 20.66 & 20.49 & 0.44 & 0.22 \\
\addlinespace[1pt]
\rowcolor{orange!7}\multicolumn{6}{l}{\textit{\BusterXpp}} \\
Wan2.2 dynamic    & 5.96 & 6.61 & 6.34 & 0.66 & 0.33 \\
Wan2.2-Lightning  & 9.07 & 9.18 & 8.58 & 0.60 & 0.30 \\
LTX               & 4.10 & 3.88 & 3.61 & 0.49 & 0.25 \\
OmniWeaving       & 6.12 & 6.83 & 6.39 & 0.71 & 0.36 \\
\bottomrule
\end{tabular*}
\end{table}

\FloatBarrier
\appendixsection{Human Evaluation and Social Dissemination}
\appendixsubsection{Human Evaluation Protocol and Source-Level Recognition}
\label{app:human-protocol}

\paragraph{Analysis set.}
The human study is drawn from the 17{,}886-video RA-Bench pool: 1{,}830 real anchors and 16{,}056 generated clips from four open-source generators and five closed-source providers. The reported Stage~1 results use the standard three-review stream, which contains 17{,}850 videos and 53{,}550 judgments. This analysis set comprises 1{,}812 real videos and 16{,}038 generated videos, with each video assigned to three different reviewers. The remaining 36 videos were reserved for internal assignment-level quality control and are excluded from the reported recognition statistics and RA-Bench-HumanProof construction.

\paragraph{Reviewers, training, and assignment.}
The 20 Stage~1 reviewers and two additional Stage~2 reviewers are undergraduate and graduate students recruited from universities in China and abroad. Before annotation, all reviewers complete the same standardized training on the review interface and label definitions. In Stage~1, the interface presents one video at a time together with an optional 16-frame contact sheet and asks the reviewer to choose \emph{Real}, \emph{Uncertain}, or \emph{Generated}. No source or label information is shown. Real and generated videos from different sources are interleaved in a reviewer-specific randomized order, and assignments are balanced across reviewers. Reviewers may replay a video and revise their immediately preceding response. The interface does not enforce a minimum viewing time, but reviewers are instructed to inspect each video before submitting.

\paragraph{Source-level recognition.}
A judgment is counted as correct when a real video is labeled \emph{Real} or a generated video is labeled \emph{Generated}; \emph{Uncertain} is reported separately and is not counted as correct. Table~\ref{tab:app_human_source} reports both the number of reviewed items and the response distribution for every source.

\begin{table}[H]
\caption{\textbf{Stage-1 human recognition by source.}
The standard analysis stream contains three judgments per video. Response shares are in \%; for generated sources, \emph{Judged Generated} is human FakeR. Open and Closed averages give equal weight to each generation source, whereas Generated total pools all generated-video judgments.}
\label{tab:app_human_source}
\centering
\scriptsize
\setlength{\tabcolsep}{4.0pt}
\renewcommand{\arraystretch}{1.04}
\textbf{Source-level response distribution}\par\smallskip
\begin{tabular*}{\textwidth}{@{\extracolsep{\fill}}lccccc@{}}
\toprule
\textbf{Video source} & \textbf{Videos} & \textbf{Judgments} &
\makecell{\textbf{Judged}\\\textbf{Real}} & \textbf{Uncertain} &
\makecell{\textbf{Judged}\\\textbf{Generated}} \\
\midrule
Real videos & 1{,}812 & 5{,}436 & \textbf{71.9} & 5.3 & 22.8 \\
\midrule
\rowcolor{vfdStageA!9}\multicolumn{6}{l}{\textit{Open-source generators}} \\
Wan2.2 dynamic    & 1{,}828 & 5{,}484 & 27.0 & 5.5 & 67.5 \\
Wan2.2-Lightning  & 1{,}828 & 5{,}484 & 28.5 & 4.9 & 66.6 \\
LTX                & 1{,}828 & 5{,}484 & 30.3 & 5.1 & 64.6 \\
OmniWeaving        & 1{,}828 & 5{,}484 & 19.5 & 4.7 & 75.8 \\
\textbf{Open avg.} & -- & -- & \textbf{26.3} & \textbf{5.0} & \textbf{68.6} \\
\midrule
\rowcolor{vfdStageB!9}\multicolumn{6}{l}{\textit{Closed-source generators}} \\
HappyHorse          & 1{,}785 & 5{,}355 & 37.2 & 6.9 & 55.9 \\
Runway              & 1{,}803 & 5{,}409 & 34.8 & 5.6 & 59.7 \\
Kling               & 1{,}788 & 5{,}364 & 47.7 & 7.3 & 45.1 \\
Seedance2.0         & 1{,}522 & 4{,}566 & 51.9 & 7.4 & 40.7 \\
Hailuo              & 1{,}828 & 5{,}484 & 31.8 & 5.3 & 63.0 \\
\textbf{Closed avg.} & -- & -- & \textbf{40.6} & \textbf{6.5} & \textbf{52.9} \\
\midrule
\textbf{Generated total} & \textbf{16{,}038} & \textbf{48{,}114} &
\textbf{33.9} & \textbf{5.8} & \textbf{60.3} \\
\bottomrule
\end{tabular*}
\end{table}

Overall, 60.3\% of judgments on generated videos and 71.9\% of judgments on real videos are correct. Generated-video recognition varies substantially by source. The source-equal human FakeR is 68.6\% for open-source generators and 52.9\% for closed-source providers. Seedance2.0 and Kling are the most difficult to recognize as generated, with FakeR values of 40.7\% and 45.1\%, whereas OmniWeaving reaches 75.8\%. The pooled human FakeR of 60.3\% therefore masks these source-level differences.

\paragraph{Reviewer-level variation and source-level stability.}
Reviewers use different response thresholds: individual FakeR ranges from 18.8\% to 87.5\%, RealR from 13.7\% to 93.3\%, and the share of \emph{Uncertain} responses from 0.0\% to 30.4\%. The source-level result is nevertheless stable. Every reviewer obtains higher source-equal FakeR on open-source than on closed-source generators; the reviewer-level difference has a median of 15.7 points and ranges from 0.6 to 38.2 points. Leaving out any one reviewer preserves the complete nine-source ranking, and the largest change in a source-level response rate is 2.9 points.

\FloatBarrier
\appendixsubsection{RA-Bench-HumanProof Construction and Detector Evaluation}
\label{app:challenge-set}

\paragraph{Two-stage selection.}
Stage~1 retains a generated video as a candidate only when all three assigned reviewers label it \emph{Real}. This criterion selects 1{,}080 of the 16{,}038 generated videos in the primary analysis. Two additional reviewers then independently reassess all candidates using the same interface and response options, without source or class labels. A candidate enters RA-Bench-HumanProof only when both additional reviewers also label it \emph{Real}; any \emph{Uncertain} or \emph{Generated} response excludes the video. The second stage retains 633 videos, or 58.6\% of the Stage-1 candidates. Each retained video therefore receives five \emph{Real} judgments.

Table~\ref{tab:app_stage2_pairs} reports the source composition before and after Stage~2 together with the complete response-pair counts.

\begin{table}[H]
\caption{\textbf{Two-stage construction and source composition of RA-Bench-HumanProof.}
(a) Source-wise retention from generated videos labeled \emph{Real} by all three Stage~1 reviewers to those also labeled \emph{Real} by both Stage~2 reviewers. Retention rates are in \%. (b) Complete Stage~2 response pairs; only the Real/Real cell enters RA-Bench-HumanProof.}
\label{tab:app_stage2_pairs}
\centering
\scriptsize
\renewcommand{\arraystretch}{1.04}
\begin{minipage}[t]{0.575\textwidth}
\vspace{0pt}
\centering
\textbf{(a) Source-wise retention}\par\smallskip
\setlength{\tabcolsep}{3.2pt}
\begin{tabular*}{\linewidth}{@{\extracolsep{\fill}}lccc@{}}
\toprule
\textbf{Source} & \makecell{\textbf{Stage~1}\\\textbf{3/3 Real}} &
\makecell{\textbf{Final}\\\textbf{5/5 Real}} & \textbf{Retained} \\
\midrule
\rowcolor{vfdStageA!9}\multicolumn{4}{l}{\textit{Open-source generators}} \\
Wan2.2 dynamic    & 49 & 21 & 42.9 \\
Wan2.2-Lightning  & 77 & 40 & 51.9 \\
LTX                & 97 & 49 & 50.5 \\
OmniWeaving        & 23 & 9  & 39.1 \\
\textbf{Open total} & \textbf{246} & \textbf{119} & \textbf{48.4} \\
\midrule
\rowcolor{vfdStageB!9}\multicolumn{4}{l}{\textit{Closed-source generators}} \\
HappyHorse          & 134 & 74  & 55.2 \\
Runway              & 122 & 74  & 60.7 \\
Kling               & 235 & 160 & 68.1 \\
Seedance2.0         & 249 & 159 & 63.9 \\
Hailuo              & 94  & 47  & 50.0 \\
\textbf{Closed total} & \textbf{834} & \textbf{514} & \textbf{61.6} \\
\midrule
\textbf{All sources} & \textbf{1{,}080} & \textbf{633} & \textbf{58.6} \\
\bottomrule
\end{tabular*}
\end{minipage}\hfill
\begin{minipage}[t]{0.395\textwidth}
\vspace{0pt}
\centering
\textbf{(b) Stage-2 response pairs}\par\smallskip
\setlength{\tabcolsep}{2.8pt}
\begin{tabular*}{\linewidth}{@{\extracolsep{\fill}}llcc@{}}
\toprule
\makecell{\textbf{Reviewer}\\\textbf{1}} &
\makecell{\textbf{Reviewer}\\\textbf{2}} & \textbf{Count} & \textbf{Kept} \\
\midrule
Real      & Real      & 633 & Yes \\
Real      & Generated & 276 & No \\
Generated & Real      & 86  & No \\
Generated & Uncertain & 1   & No \\
Generated & Generated & 84  & No \\
\bottomrule
\end{tabular*}
\end{minipage}
\end{table}

RA-Bench-HumanProof contains 119 open-source and 514 closed-source videos. Closed-source candidates have a higher retention rate than open-source candidates (61.6\% versus 48.4\%). Kling and Seedance2.0 contribute 160 and 159 videos, respectively, and together account for 50.4\% of RA-Bench-HumanProof. This composition follows directly from the five-reviewer selection criterion rather than a predefined source quota.

\paragraph{Paired evaluation and source-matched reference.}
Every generated video in RA-Bench-HumanProof retains the identifier of its real anchor. Detector evaluation therefore uses 633 generated--real pairs, corresponding to 511 unique real videos because several generated videos may share an anchor. For discrete outputs, we report BAcc together with FakeR and RealR. For continuous scores, we report paired AUC and the fake-video true-positive rate at a 5\% real-video false-positive rate (T@5\%).

RA-Bench-HumanProof contains a larger share of Seedance2.0 and Kling videos than full RA-Bench. An unweighted comparison would therefore combine human-selection effects with a change in source composition. For each metric, we compute the RA-Bench reference as \(\sum_s n_s M_s / \sum_s n_s\), where \(M_s\) is the full-RA-Bench result for source \(s\) and \(n_s\) is its RA-Bench-HumanProof count. This reference matches the RA-Bench-HumanProof source proportions while retaining all videos within each RA-Bench source. It controls the source mixture, but not the conditional selection induced by the five human judgments.

\begin{table}[H]
\caption{\textbf{Complete detector results on RA-Bench-HumanProof.}
(a) Continuous-score configurations report paired AUC and T@5\%. (b) Discrete-output configurations report BAcc, FakeR, and RealR. \emph{Matched ref.} reports the corresponding AUC/T@5\% or BAcc/FakeR pair on source-matched RA-Bench, and \(\Delta\) is the RA-Bench-HumanProof AUC or BAcc minus its matched reference. All values are in \%. Skyra official-timestamp results follow the released protocol; frame index removes the numerical timestamp values.}
\label{tab:app_challenge_detector_full}
\centering
\scriptsize
\renewcommand{\arraystretch}{1.05}

\textbf{(a) Continuous-score configurations}\par\smallskip
\setlength{\tabcolsep}{4.0pt}
\begin{tabular*}{0.96\textwidth}{@{\extracolsep{\fill}}lccccc@{}}
\toprule
\textbf{Configuration} & \makecell{\textbf{RA-Bench-}\\\textbf{HumanProof}\\\textbf{AUC}} &
\textbf{T@5\%} & \makecell{\textbf{Matched ref.}\\\textbf{AUC}} &
\makecell{\textbf{Matched ref.}\\\textbf{T@5\%}} &
\(\boldsymbol{\Delta}\) \\
\midrule
\rowcolor{vfdStageA!9}\multicolumn{6}{l}{\textit{Traditional detectors}} \\
CNNSpot    & 40.5 & 3.6 & 43.5 & 3.8  & -3.0 \\
NPR        & 50.3 & 3.8 & 50.4 & 4.0  & -0.1 \\
UnivFD     & 39.4 & 0.6 & 39.8 & 1.0  & -0.4 \\
ForgeLens  & 54.9 & 8.8 & 58.4 & 10.4 & -3.5 \\
DeCoF      & 59.0 & 1.9 & 59.2 & 1.7  & -0.2 \\
D3         & 33.1 & 6.8 & 39.5 & 6.2  & -6.4 \\
ReStraV    & 55.2 & 5.2 & 55.6 & 4.9  & -0.4 \\
\textbf{7-detector mean} & \textbf{47.5} & \textbf{4.4} & \textbf{49.5} & \textbf{4.6} & \textbf{-2.0} \\
\midrule
\rowcolor{vfdExample!9}\multicolumn{6}{l}{\textit{Zero-shot multimodal models}} \\
Gemini-3.1-Pro-Preview Rating & 54.9 & 4.3 & 61.5 & 5.6 & -6.6 \\
\bottomrule
\end{tabular*}

\medskip
\textbf{(b) Discrete-output configurations}\par\smallskip
\setlength{\tabcolsep}{3.5pt}
\begin{tabular*}{\textwidth}{@{\extracolsep{\fill}}lcccccc@{}}
\toprule
\textbf{Configuration} & \textbf{BAcc} & \textbf{FakeR} & \textbf{RealR} &
\makecell{\textbf{Matched ref.}\\\textbf{BAcc}} &
\makecell{\textbf{Matched ref.}\\\textbf{FakeR}} &
\(\boldsymbol{\Delta}\) \\
\midrule
\rowcolor{vfdExample!9}\multicolumn{7}{l}{\textit{Zero-shot multimodal models}} \\
Gemini-3.1-Pro-Preview Binary     & 54.7 & 34.3 & 75.2 & 61.2 & 49.9 & -6.5 \\
Gemini-3.1-Pro-Preview Diagnostic & 54.5 & 30.0 & 79.0 & 61.0 & 47.3 & -6.5 \\
\midrule
\rowcolor{vfdStageB!9}\multicolumn{7}{l}{\textit{Fine-tuned MLLM detectors}} \\
Skyra-SFT, official timestamp & 72.0 & 55.1 & 88.8 & 73.9 & 60.4 & -1.9 \\
Skyra-SFT, frame index        & 53.7 & 51.3 & 56.0 & 55.3 & 55.7 & -1.6 \\
Skyra-RL, official timestamp  & 74.5 & 63.0 & 85.9 & 75.1 & 66.3 & -0.6 \\
Skyra-RL, frame index         & 53.7 & 58.8 & 48.7 & 56.1 & 63.3 & -2.4 \\
\BusterXpp, released pipeline  & 49.4 & 3.9  & 94.9 & 49.9 & 6.2  & -0.5 \\
\bottomrule
\end{tabular*}
\end{table}

All configurations in Table~\ref{tab:app_challenge_detector_full} contain predictions for all 633 RA-Bench-HumanProof videos. The traditional detectors, Gemini, Skyra-RL, and \BusterXpp\ also contain 633 matched-real predictions. Skyra-SFT contains 632 valid matched-real predictions after excluding one invalid record; its BAcc and RealR use the available predictions.

\paragraph{Detector-family results.}
The seven traditional detectors show a modest mean AUC decrease of 2.0 points, from 49.5\% under source-matched RA-Bench weighting to 47.5\% on RA-Bench-HumanProof. Their individual AUCs span only 33.1\%--59.0\%, and their mean T@5\% changes from 4.6\% to 4.4\%. These small or heterogeneous changes do not indicate robustness: the source-matched baseline is already close to random ranking and provides little generated-video recall at a 5\% false-positive rate.

Gemini shows the clearest alignment with human difficulty. Binary and Diagnostic BAcc each decrease by 6.5 points, while their FakeR decreases by 15.6 and 17.3 points, respectively. Rating AUC decreases by 6.6 points and T@5\% by 1.3 points. The loss is concentrated on the generated side: Binary and Diagnostic FakeR fall to 34.3\% and 30.0\%, while RealR remains 75.2\% and 79.0\%. On RA-Bench-HumanProof, Gemini therefore provides substantially weaker fake evidence without a comparable loss on the matched real videos.

The fine-tuned MLLMs require a different interpretation. Their BAcc changes by at most 2.4 points, and FakeR decreases by 2.3--5.3 points across all five configurations. Skyra-SFT and Skyra-RL retain 72.0\% and 74.5\% BAcc with official timestamps, but these configurations preserve the timestamp prior identified in Section~\ref{sec:finetuned-shortcut}. Replacing timestamps with frame indices reduces both checkpoints to 53.7\% BAcc. \BusterXpp\ reaches 49.4\% BAcc by labeling only 3.9\% of generated videos as fake while retaining 94.9\% of real videos. Their limited changes from the source-matched reference therefore reflect an already weak visual decision rule or a strong class preference, not reliable detection of human-deceptive content.

\paragraph{Interpretation and scope.}
RA-Bench-HumanProof separates three failure patterns: Gemini loses much of its fake-side evidence on videos that mislead reviewers; traditional detectors remain weak before and after human selection; and the higher official-timestamp Skyra results retain a protocol prior. Human and detector failures therefore overlap, but they are not equivalent. The most consequential cases lie at their intersection: videos that repeatedly appear real to reviewers also receive weak or unreliable fake evidence from the evaluated detector families.

RA-Bench-HumanProof is constructed conditionally and does not replace full RA-Bench. Stage~2 responses determine membership and therefore cannot provide an independent estimate of human accuracy on the retained videos. Its source distribution is skewed toward Seedance2.0 and Kling, which motivates the source-matched reference. The 633 generated--real pairs also include repeated real anchors; interval estimates should therefore resample the real-anchor identifier rather than treat all pairs as independent.

\FloatBarrier

\appendixsubsection{RA-Bench-LastMile Protocol and Complete Results}
\label{app:propagation-protocol}

\paragraph{Evaluation subset.}
RA-Bench-LastMile uses anchors for which a real video and generated videos from all nine RA-Bench sources are available. To preserve coverage without allowing large L2 subcategories to dominate, let $N_l$ denote the number of common anchors in subcategory $l$. We retain $q_l=\lfloor 0.1N_l+0.5\rfloor$ anchors when $N_l>10$, one anchor when $1\leq N_l\leq10$, and none when $N_l=0$. A fixed SHA-256 ordering of normalized clip identifiers determines the retained anchors within each subcategory. This procedure selects 150 real-event anchors spanning 41 of the 44 L2 subcategories. Each condition contains these 150 real videos and 1{,}350 matched generated videos, for 1{,}500 videos per condition and 9{,}000 video instances across the six conditions. The Wan2.2 fixed-duration variant serves as an auxiliary control and is not included as a separate generation source.

\paragraph{Social dissemination simulation.}
All operations are applied identically to a generated video and its matched real anchor. Table~\ref{tab:app_propagation_conditions} summarizes the six conditions. T1 encodes each video with VP9 (CRF 36, \texttt{-b:v 0}) and then H.264 (CRF 28, \texttt{medium} preset). The spatial, temporal, and presentation operations are each evaluated as an addition to this common transcode, and Full applies all four operations in sequence.

\begin{table}[H]
\caption{\textbf{Social dissemination simulation conditions.} Each condition contains 150 real videos and 1{,}350 matched generated videos. T2--T4 are evaluated as additions to the common T1 transcode.}
\label{tab:app_propagation_conditions}
\centering
\footnotesize
\setlength{\tabcolsep}{5.2pt}
\renewcommand{\arraystretch}{1.04}
\begin{tabular}{@{}lp{0.78\textwidth}@{}}
\toprule
\textbf{Condition} & \textbf{Operations, in order} \\
\midrule
Original & Standardized input clip; no additional encoding. \\
T1 & VP9 encoding followed by H.264 transcoding. \\
T1+T2 & $0.5\times$ spatial downsampling, followed by T1. \\
T1+T3 & Conversion to 8 fps, followed by T1. \\
T1+T4 & Synthetic news badge, followed by T1. \\
Full & Spatial downsampling, 8-fps conversion, news badge, and T1. \\
\bottomrule
\end{tabular}
\end{table}

\begin{figure}[H]
\centering
\includegraphics[width=0.98\linewidth]{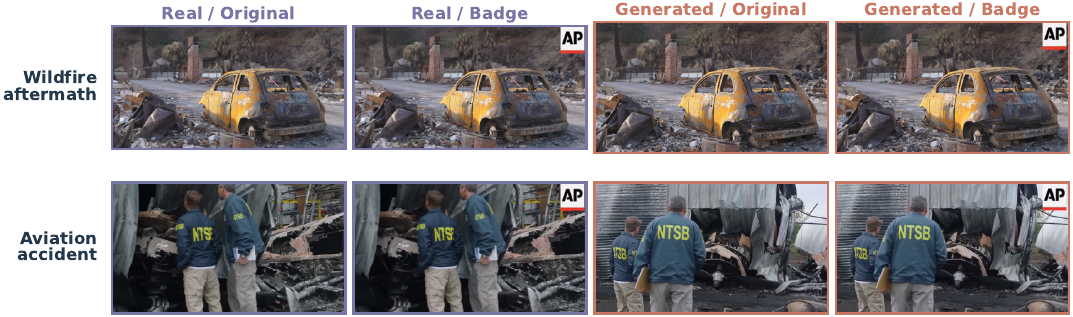}
\caption{\textbf{Illustration of the news-badge transformation.}
Matched real and Seedance2 videos are shown before and after adding the synthetic news badge. The same overlay rule is applied to both classes while preserving the remaining scene content. We show a wildfire aftermath scene (top) and an aviation accident investigation (bottom).}
\label{fig:app_propagation_badge}
\end{figure}

As illustrated in Figure~\ref{fig:app_propagation_badge}, the synthetic badge is placed at the upper-right corner of the active picture. Its width is 10.04\% of the active-picture width, with a 1.67\% right margin. The overlay is an experimental presentation cue rather than source attribution and does not imply affiliation with or endorsement by any news organization. If a clip already contains an upper-right news badge, the existing badge is retained and no second badge is added. Audio tracks and container metadata are removed from every transformed condition; all transformed outputs use H.264, \texttt{yuv420p}, and \texttt{faststart}.

\paragraph{Evaluation and uncertainty.}
For each traditional detector, we compute metrics separately on each generation source and its matched real anchors, then average the nine source-level values with equal weight. Classification metrics for the fine-tuned MLLM detectors follow the same protocol. Confidence intervals use 2{,}000 bootstrap replicates, stratified by L1 category and clustered by real-event anchor, with random seed 20260718. All reported configurations contain complete predictions for the nine sources under all six conditions. We repeat the analysis after excluding manually identified clips with a pre-existing upper-right badge. This exclusion does not materially change the results: across method--condition cells, the largest absolute changes are 2.8 percentage points for AUC and 4.3 points for FakeR.

\begin{table}[t]
\captionsetup{width=\textwidth}
\caption{\textbf{Complete detector results on RA-Bench-LastMile.}
Each cell reports the primary metric on top and the secondary metric below: AUC/T@5\% for traditional detectors and BAcc/FakeR for fine-tuned MLLM detectors. All values are percentages and equal-source means over the nine RA-Bench generation sources.}
\label{tab:app_propagation_full}
\centering
\scriptsize
\setlength{\tabcolsep}{8.0pt}
\renewcommand{\arraystretch}{1.00}
\begin{tabular}{@{}l *{6}{c}@{}}
\toprule
\textbf{Detector / configuration} & \textbf{Original} & \textbf{T1} & \textbf{T1+T2} & \textbf{T1+T3} & \textbf{T1+T4} & \textbf{Full} \\
\midrule
\rowcolor{vfdStageA!9}
\textbf{Traditional detectors} & \multicolumn{6}{c}{\textbf{AUC} \textcolor{vfdInk!58}{/ T@5\%}} \\
CNNSpot & \shortstack{44.5\\[-0.2ex]\textcolor{vfdInk!58}{2.4}} & \shortstack{48.2\\[-0.2ex]\textcolor{vfdInk!58}{2.4}} & \shortstack{48.0\\[-0.2ex]\textcolor{vfdInk!58}{2.4}} & \shortstack{50.1\\[-0.2ex]\textcolor{vfdInk!58}{2.1}} & \shortstack{47.5\\[-0.2ex]\textcolor{vfdInk!58}{1.9}} & \shortstack{46.5\\[-0.2ex]\textcolor{vfdInk!58}{2.2}} \\
NPR & \shortstack{52.2\\[-0.2ex]\textcolor{vfdInk!58}{3.8}} & \shortstack{45.6\\[-0.2ex]\textcolor{vfdInk!58}{3.4}} & \shortstack{44.9\\[-0.2ex]\textcolor{vfdInk!58}{3.0}} & \shortstack{45.9\\[-0.2ex]\textcolor{vfdInk!58}{2.7}} & \shortstack{45.5\\[-0.2ex]\textcolor{vfdInk!58}{3.5}} & \shortstack{43.5\\[-0.2ex]\textcolor{vfdInk!58}{3.0}} \\
UnivFD & \shortstack{37.1\\[-0.2ex]\textcolor{vfdInk!58}{2.1}} & \shortstack{36.6\\[-0.2ex]\textcolor{vfdInk!58}{1.8}} & \shortstack{36.0\\[-0.2ex]\textcolor{vfdInk!58}{0.3}} & \shortstack{38.1\\[-0.2ex]\textcolor{vfdInk!58}{1.0}} & \shortstack{37.2\\[-0.2ex]\textcolor{vfdInk!58}{1.9}} & \shortstack{37.4\\[-0.2ex]\textcolor{vfdInk!58}{0.2}} \\
ForgeLens & \shortstack{61.6\\[-0.2ex]\textcolor{vfdInk!58}{17.3}} & \shortstack{44.7\\[-0.2ex]\textcolor{vfdInk!58}{6.1}} & \shortstack{32.6\\[-0.2ex]\textcolor{vfdInk!58}{2.2}} & \shortstack{46.3\\[-0.2ex]\textcolor{vfdInk!58}{6.5}} & \shortstack{44.9\\[-0.2ex]\textcolor{vfdInk!58}{4.8}} & \shortstack{35.6\\[-0.2ex]\textcolor{vfdInk!58}{3.0}} \\
DeCoF & \shortstack{59.9\\[-0.2ex]\textcolor{vfdInk!58}{3.2}} & \shortstack{63.0\\[-0.2ex]\textcolor{vfdInk!58}{3.9}} & \shortstack{60.1\\[-0.2ex]\textcolor{vfdInk!58}{0.4}} & \shortstack{65.1\\[-0.2ex]\textcolor{vfdInk!58}{5.0}} & \shortstack{63.0\\[-0.2ex]\textcolor{vfdInk!58}{3.3}} & \shortstack{62.3\\[-0.2ex]\textcolor{vfdInk!58}{1.2}} \\
D3 & \shortstack{49.3\\[-0.2ex]\textcolor{vfdInk!58}{4.4}} & \shortstack{49.5\\[-0.2ex]\textcolor{vfdInk!58}{5.5}} & \shortstack{52.2\\[-0.2ex]\textcolor{vfdInk!58}{4.7}} & \shortstack{50.4\\[-0.2ex]\textcolor{vfdInk!58}{5.9}} & \shortstack{53.3\\[-0.2ex]\textcolor{vfdInk!58}{4.4}} & \shortstack{54.1\\[-0.2ex]\textcolor{vfdInk!58}{5.4}} \\
ReStraV & \shortstack{55.4\\[-0.2ex]\textcolor{vfdInk!58}{6.8}} & \shortstack{52.2\\[-0.2ex]\textcolor{vfdInk!58}{8.2}} & \shortstack{51.1\\[-0.2ex]\textcolor{vfdInk!58}{4.5}} & \shortstack{52.0\\[-0.2ex]\textcolor{vfdInk!58}{4.8}} & \shortstack{52.0\\[-0.2ex]\textcolor{vfdInk!58}{6.5}} & \shortstack{51.6\\[-0.2ex]\textcolor{vfdInk!58}{4.7}} \\
\cmidrule(lr){1-7}
\textbf{7-detector mean} & \shortstack{\textbf{51.4}\\[-0.2ex]\textcolor{vfdInk!58}{\textbf{5.7}}} & \shortstack{\textbf{48.5}\\[-0.2ex]\textcolor{vfdInk!58}{\textbf{4.5}}} & \shortstack{\textbf{46.4}\\[-0.2ex]\textcolor{vfdInk!58}{\textbf{2.5}}} & \shortstack{\textbf{49.7}\\[-0.2ex]\textcolor{vfdInk!58}{\textbf{4.0}}} & \shortstack{\textbf{49.1}\\[-0.2ex]\textcolor{vfdInk!58}{\textbf{3.8}}} & \shortstack{\textbf{47.3}\\[-0.2ex]\textcolor{vfdInk!58}{\textbf{2.8}}} \\
\midrule
\rowcolor{vfdStageB!9}
\textbf{Fine-tuned MLLM detectors} & \multicolumn{6}{c}{\textbf{BAcc} \textcolor{vfdInk!58}{/ FakeR}} \\
Skyra-SFT, official timestamp & \shortstack{67.6\\[-0.2ex]\textcolor{vfdInk!58}{50.4}} & \shortstack{61.8\\[-0.2ex]\textcolor{vfdInk!58}{31.6}} & \shortstack{52.9\\[-0.2ex]\textcolor{vfdInk!58}{9.2}} & \shortstack{50.1\\[-0.2ex]\textcolor{vfdInk!58}{9.6}} & \shortstack{58.7\\[-0.2ex]\textcolor{vfdInk!58}{18.7}} & \shortstack{49.3\\[-0.2ex]\textcolor{vfdInk!58}{1.2}} \\
Skyra-SFT, frame index & \shortstack{55.9\\[-0.2ex]\textcolor{vfdInk!58}{55.2}} & \shortstack{48.9\\[-0.2ex]\textcolor{vfdInk!58}{33.7}} & \shortstack{39.6\\[-0.2ex]\textcolor{vfdInk!58}{5.1}} & \shortstack{45.5\\[-0.2ex]\textcolor{vfdInk!58}{31.6}} & \shortstack{42.7\\[-0.2ex]\textcolor{vfdInk!58}{10.8}} & \shortstack{44.4\\[-0.2ex]\textcolor{vfdInk!58}{1.4}} \\
Skyra-RL, official timestamp & \shortstack{68.6\\[-0.2ex]\textcolor{vfdInk!58}{55.8}} & \shortstack{62.7\\[-0.2ex]\textcolor{vfdInk!58}{36.7}} & \shortstack{53.5\\[-0.2ex]\textcolor{vfdInk!58}{11.6}} & \shortstack{49.8\\[-0.2ex]\textcolor{vfdInk!58}{12.9}} & \shortstack{60.3\\[-0.2ex]\textcolor{vfdInk!58}{22.7}} & \shortstack{48.5\\[-0.2ex]\textcolor{vfdInk!58}{1.7}} \\
Skyra-RL, frame index & \shortstack{56.5\\[-0.2ex]\textcolor{vfdInk!58}{61.7}} & \shortstack{49.0\\[-0.2ex]\textcolor{vfdInk!58}{41.3}} & \shortstack{36.2\\[-0.2ex]\textcolor{vfdInk!58}{8.4}} & \shortstack{46.6\\[-0.2ex]\textcolor{vfdInk!58}{39.8}} & \shortstack{44.4\\[-0.2ex]\textcolor{vfdInk!58}{17.6}} & \shortstack{43.2\\[-0.2ex]\textcolor{vfdInk!58}{2.4}} \\
\BusterXpp\ & \shortstack{49.1\\[-0.2ex]\textcolor{vfdInk!58}{6.9}} & \shortstack{50.0\\[-0.2ex]\textcolor{vfdInk!58}{5.9}} & \shortstack{50.5\\[-0.2ex]\textcolor{vfdInk!58}{5.0}} & \shortstack{48.9\\[-0.2ex]\textcolor{vfdInk!58}{5.2}} & \shortstack{50.1\\[-0.2ex]\textcolor{vfdInk!58}{0.2}} & \shortstack{50.1\\[-0.2ex]\textcolor{vfdInk!58}{0.2}} \\
\bottomrule
\end{tabular}
\end{table}

\paragraph{Traditional-detector sensitivity.}
The seven-detector mean AUC decreases by 4.2 points under Full, but this average conceals sharply different responses. ForgeLens loses 26.0 points, whereas DeCoF and D3 gain 2.4 and 4.8 points. The detector ordering is consequently unstable: its Spearman correlation with Original is 0.36 after T1, 0.07 after T1+T2, 0.46 after T1+T3, 0.29 after T1+T4, and 0.07 under Full. These gains for individual detectors do not indicate reliable performance under the social dissemination simulation because T@5\% remains low throughout, reaching only 2.8\% for the seven-detector mean under Full.

\paragraph{Fine-tuned MLLM sensitivity.}
The isolated additions to T1 reveal different failure patterns. Spatial downsampling produces the largest FakeR loss, reducing the five-configuration mean from 29.9\% to 7.9\%. Conversion to 8 fps lowers BAcc by 11.7 and 12.9 points for the two official-timestamp Skyra configurations, compared with 3.4 and 2.4 points for their frame-index controls. This gap is consistent with sensitivity to the temporal representation rather than only to visual degradation. The news badge lowers mean FakeR from 29.9\% to 14.0\% while increasing RealR from 79.1\% to 88.5\%, shifting predictions toward \emph{Real} even though the scene content is unchanged. Under Full, every fine-tuned configuration has at most 2.4\% FakeR.

\begin{table}[H]
\caption{\textbf{Uncertainty of Full-condition changes relative to Original.} Values are percentage-point changes with 95\% anchor-cluster bootstrap confidence intervals.}
\label{tab:app_propagation_ci}
\centering
\begin{minipage}[t]{0.64\textwidth}
\vspace{0pt}
\centering
\footnotesize
\setlength{\tabcolsep}{4.3pt}
\begin{tabular}{@{}lcc@{}}
\toprule
\textbf{Method} & \textbf{Metric} & \textbf{Change [95\% CI]} \\
\midrule
Seven-detector mean & AUC & $-4.2\;[-5.6,-2.7]$ \\
Skyra-SFT, timestamp & BAcc & $-18.3\;[-21.1,-15.4]$ \\
Skyra-SFT, frame index & BAcc & $-11.5\;[-15.0,-8.3]$ \\
Skyra-RL, timestamp & BAcc & $-20.0\;[-23.1,-16.9]$ \\
Skyra-RL, frame index & BAcc & $-13.3\;[-16.7,-10.1]$ \\
\bottomrule
\end{tabular}
\end{minipage}\hfill
\begin{minipage}[t]{0.32\textwidth}
\vspace{0pt}
\footnotesize
\textbf{Interpretation.}
All reported intervals remain below zero under L1-stratified, anchor-cluster resampling. \BusterXpp\ is omitted from this change table because its BAcc remains near 50\% in both conditions; its FakeR nevertheless falls from 6.9\% to 0.2\%, showing that a stable BAcc can conceal a stronger shift toward \emph{Real}.
\end{minipage}
\end{table}

\FloatBarrier

\end{document}

%% file: math_commands.tex
\usepackage{amsmath,amsfonts,bm}

\def\eqref#1{equation~\ref{#1}}

\def\1{\bm{1}}

\DeclareMathAlphabet{\mathsfit}{\encodingdefault}{\sfdefault}{m}{sl}
\SetMathAlphabet{\mathsfit}{bold}{\encodingdefault}{\sfdefault}{bx}{n}

